\documentclass[twoside,11pt]{article}

\usepackage[abbrvbib,  preprint]{jmlr2e}

\usepackage{blindtext}

\usepackage{bm}
\usepackage{epstopdf}
\usepackage{amsmath}
\usepackage{multirow}
\usepackage{threeparttable}

\usepackage{hyperref}

\usepackage{float}
\usepackage{rotating}
\usepackage{amssymb}
\usepackage{amsfonts}
\usepackage{mathrsfs}
\usepackage{algorithmic}
\usepackage{graphicx}
\usepackage{textcomp}
\usepackage{dsfont}
\usepackage{color}

\usepackage{caption}

\usepackage{subcaption}
\usepackage{multicol}
\usepackage{footmisc}
\usepackage[T1]{fontenc}
\usepackage{bbding}
\usepackage{epsfig}
\usepackage{ulem}
\usepackage[thicklines]{cancel}
\usepackage[english]{babel}
\usepackage{balance}

\usepackage{lastpage}
\jmlrheading{x}{2025}{1-\pageref{LastPage}}{x/x; Revised x/x}{x/x}{x-x}{Weizhi Lu, Mingrui Chen and Weiyu Li}

\ShortHeadings{Quantization Can Enhance Feature Discrimination}{Lu, Chen and Li}
\firstpageno{1}

\newtheorem{property}[theorem]{Property}  % added by LU

\begin{document}

\title{Binary and Ternary Quantization Can Enhance \\ Feature Discrimination}

\author{\name Weizhi Lu  \email wzlu@sdu.edu.cn\\
       \addr School of Control Science and Engineering, Shandong University\\
       Key Laboratory of Machine Intelligence and System Control, Ministry of Education
      \AND
       \name Mingrui Chen  \email mrchen@mail.sdu.edu.cn\\
       \addr School of Control Science and Engineering, Shandong University
       \AND
       \name  Weiyu Li\thanks{Corresponding author.}  \email liweiyu@sdu.edu.cn\\
       \addr Zhongtai Securities Institute for Financial Studies, Shandong University\\
       National Center for Applied Mathematics in Shandong
}

\editor{}

\maketitle

\begin{abstract}%   <- trailing '%' for backward compatibility of .sty file

Quantization is widely applied in machine learning to reduce computational and storage
costs for both data and models. Considering that classification tasks are fundamental to the field,
it is crucial to investigate how quantization impacts classification performance. Traditional
research has focused on quantization errors, assuming that larger errors generally lead to
lower classification accuracy. However, this assumption lacks a solid theoretical foundation
and often contradicts empirical observations.  For example, despite introducing significant
errors,   $\{0,1\}$-binary  and   $\{0, \pm1\}$-ternary quantized data have sometimes achieved classification
accuracy comparable or even superior to full-precision data.  To reasonably explain this phenomenon, a more accurate evaluation of classification performance is required. To achieve this, we
propose a direct analysis of the feature discrimination of quantized data, instead of focusing
on quantization errors. Our analysis reveals that both binary and ternary quantization can potentially enhance, rather than degrade, the feature discrimination of the original
data. This finding is supported by classification experiments conducted on both synthetic
and real data.

\end{abstract}


\section{Introduction}

Quantization has been widely applied in machine learning to simplify data storage and computation complexities, while also catering to the requirements of algorithm deployment on digital hardware. In general, this operation will lead to a decrease in classification accuracy \citep{baras1999combined,hoefler2021sparsity}, due to reducing the precision of data or model parameters. To achieve a balance between  complexity and accuracy, it is crucial  to delve into the impact of quantization on classification.  Currently, the impact is mainly evaluated through  quantization errors, under the premise that  larger quantization errors generally lead to decreased classification accuracy \citep{lin2016fixed}. However, this premise lacks a solid theoretical basis \citep{lin2016fixed}, as it merely adopts the quantization principle from signal processing \citep{gray1998quantization}, which primarily focuses on data reconstruction fidelity rather than classification accuracy.  In practice, it appears challenging to accurately evaluate the classification performance using quantization errors.

	For instance, it has been observed that some extremely low bit-width quantization methods, such as 1-bit $\{0,1\}$-binary  and 2-bit $\{0,\pm1\}$-ternary  quantization, which have been successively applied in large-scale retrieval \citep{charikar2002similarity} and  deep network quantization \citep{qin2020binary,gholami2022survey, wang2023bitnet},  can achieve comparable  or even superior classification performance than their full-precision counterparts \citep{courbariaux2015binaryconnect, Zhouhan2016Neural,lu2023quantization},  despite suffering from high quantization errors.  Apparently, the remarkable classification  improvement resulting from quantization should not be attributed to the significant quantization errors. This reveals the inadequacy of quantization errors in evaluating the actual classification performance.  Due to the absence of a theoretical explanation, the classification improvement induced by quantization has often been regarded as incidental and received little attention. Instead of quantization errors, in the paper we demonstrate that this intriguing phenomenon can be reasonably  explained by feature discrimination. Following the Fisher's linear discriminant analysis \citep{fisher1936use}, we here refer to feature discrimination as the ratio between inter-class and intra-class scatters, and evaluate the classification performance based on the rule that the higher the feature discrimination, the easier the classification. To the best of our knowledge, this is the first study that exploits feature discrimination to investigate the impact of quantization on classification,  although it is more direct and reasonable than quantization errors in evaluating   classification performance. The scarcity of relevant research may be attributed to the  nonlinearity of the quantization operation, which substantially increases the analytical complexity of feature discrimination functions.

 In the paper, it is demonstrated that the impact of the threshold-based binary and ternary quantization on  feature discrimination can be analyzed, when the data are appropriately modeled using a Gaussian mixture model, with each Gaussian element representing one class of data. The Gaussian mixture model is chosen here based on two  considerations.  Firstly, the model has been well-established for approximating the distributions of real-world data \citep{torralba2003statistics,weiss2007makes} and their  feature transformations \citep{Wainwright99Gaussian, lam2000mathematical}. Secondly, the closure property of Gaussian distributions under linear operations can simplify the analysis of the feature discrimination function. By analyzing the discrimination across varying quantization thresholds, it is found that there exist certain quantization thresholds that can enhance the discrimination of  original data,  thereby yielding improved classification performance. This finding is validated through extensive classification experiments on both synthetic and real data, covering diverse modalities such as images, speech, and text.

 The related works are discussed as follows.  As mentioned earlier, our work should be the first to take advantage of  feature discrimination to investigate the impact of quantization on classification. In the filed of signal processing, there have been  a few works proposed to reduce the negative impact of quantization on signal detection or classification \citep{poor1977applications,oehler1995combining}. However, these studies did not employ feature discrimination analysis, distinguishing them  from our research  in both methodology and outcomes.  Specifically, in these studies the model design accounts for both   reconstruction loss and classification loss. The classification loss is commonly modeled in various ways,  such as directly minimizing the classification error on quantized data \citep{srinivasamurthy2002reduced},  enlarging the inter-class distance between quantized data \citep{jana2000optimal,jana2003optimal}, reducing the difference between the distributions of quantized data and original data \citep{baras1999combined}, as well as minimizing the discrepancy in classification  before and after quantization \citep{dogahe2011quantization}. Based on these loss models,  the classification performance of quantized data  can  only approach, but not exceed, the performance of original data \citep{baras1999combined}.

 %and in other words, it is hard to derive the quantization thresholds for the binary or ternary quantization to improve the classification performance.

\section{Problem Formulation}
In this section, we  specify the   feature  discrimination functions for the original (non-quantized) and  quantized data. Prior to this, we introduce the binary and ternary quantization functions,  as well as the data modeling.

%In this section,  we provide some prerequisite knowledge for investigating the impact of quantization on feature discrimination, including the forms of quantization functions, the statistical modeling of real data, and the definition of discrimination.

	\subsection{Quantization functions}
The binary and ternary quantization functions are formulated as
\begin{equation}\label{b_q}
	f_b(x;\tau) =
	\begin{cases}
		1, & \text{if } x > \tau \\
		0, & \text{otherwise}
	\end{cases}
\end{equation}
and
\begin{equation}\label{t_q}
	f_t(x;\tau) =
	\begin{cases}
		1, & \text{if } x > \tau \\
		0, & \text{if } -\tau \le x \le \tau \\
		-1, & \text{if } x < -\tau
	\end{cases}
\end{equation}
where the threshold parameter $\tau\in (-\infty, +\infty)$  for $f_b(x;\tau)$, and  $\tau\in [0,+\infty)$ for $f_t(x;\tau)$. The two functions  operate  element-wise  on a vector  $\mathbf{x}=(x_1,x_2,\cdots,x_n)^\top\in\mathbb{R}^{n}$, namely $f_b(\mathbf{x};\tau)=(f_b(x_1;\tau),f_b(x_2;\tau),\cdots, f_b(x_n;\tau))^\top$ and the same applies to $f_t(\mathbf{x};\tau)$.

\subsection{Data distributions}
Throughout the work, we denote each data sample using a vector.  For the sake of generality, as discussed before, we assume that the data vector randomly drawn from a class is a random vector $\mathbf{X}=\{X_1,X_2,\cdots,X_n\}^\top$, with its each element $X_i$ following a Gaussian distribution $N(\mu_{1,i}, \sigma^2)$; and similarly, for the random vector $\mathbf{Y}=\{Y_1,Y_2,\cdots,Y_n\}^\top$ drawn from another class, we suppose its each element $Y_i\sim N(\mu_{2,i}, \sigma^2)$, where $\mu_{2,i}\neq\mu_{1,i}$. Considering that the discrimination between the two random vectors $\mathbf{X}$ and $\mathbf{Y}$ positively correlates with the discrimination between their each pair of corresponding elements  $X_i$ and $Y_i$, we propose to analyze  the discrimination at the element level, specifically between  $X_i$ and $Y_i$,  rather than  between the entire vectors, $\mathbf{X}$ and $\mathbf{Y}$.   For notational convenience, without causing confusion, in the sequel we will omit the subscript "$i$" of $X_i$ and $Y_i$, and  write their distributions as $X \sim N(\mu_1, \sigma^2)$ and $Y \sim N(\mu_2, \sigma^2)$, where $\mu_1\neq\mu_2$. Note that we assume here that the two variables $X$ and $Y$ share the same variance $\sigma^2$. This assumption is common in statistical research, as the data we intend to investigate are often drawn from the same or similar scenarios and thus exhibit similar noise levels.

%For the sake of generality,  as discussed before, we typically assume that each element of a data vector drawn from a class follows a Gaussian distribution.  their elements drawn from a Gaussian disctribution  each element holds a Gaussian distribution, different from the
%we assume two classes of data are drawn from two different Gaussian distributions: $X \sim N(\mu_1, \sigma^2)$ and $Y \sim N(\mu_2, \sigma^2)$, where $\mu_1\neq\mu_2$. Note that we here simply suppose the data is one dimensional for easing the following discrimination analysis, and the analysis can be extended to  multidimensional data.

%For the sake of generality,  we assume two classes of data are drawn from two different Gaussian distributions: $X \sim N(\mu_1, \sigma^2)$ and $Y \sim N(\mu_2, \sigma^2)$, where $\mu_1\neq\mu_2$. Note that we here simply suppose the data is one dimensional for easing the following discrimination analysis, and the analysis can be extended to  multidimensional data.

%Provided the two different variables $X$ and $Y$, we then can study the classification on the mixture  $Z$ of them.  Typically, we assume the mixture $Z$ has a balanced class distribution, namely having samples drawn from  $X$ and $Y$ with equal probability. To unify data scales,  it is common  in machine learning to standardize  each sample of $Z$ by  subtracting the mean and dividing by the standard deviation. This standardization will transform the distributions of $X$ and $Y$ (contained in $Z$) to be

To facilitate data processing, we  commonly standardize variables $X$ and $Y$ using Z-score standardization, which induces specific relationships between their distributions.  More precisely, in a  binary classification problem,  the dataset we handle is a  mixture, denoted as  $Z$, comprising two classes of samples drawn respectively from  $X$ and $Y$. Usually, the mixture $Z$ is assumed to  possess  a balanced class distribution, meaning that samples are drawn from  $X$ and $Y$ with equal probabilities. Under this assumption, when we perform standardization by subtracting the mean and dividing by the standard deviation for each sample in $Z$,  the distributions of $X$ and $Y$ (in $Z$) will become

%\begin{equation}\label{dis-1}
%\begin{align}\label{dis-1}
%\tilde{X} &= \frac{X - E[Z]}{\sqrt{D[Z]}} \\
%&\sim N\left(\frac{(\mu_1 - \mu_2)/2}{\sqrt{\sigma^2 + \frac{1}{4}(\mu_1 - \mu_2)^2}}, \frac{\sigma^2}{\sigma^2 + \frac{1}{4}(\mu_1 - \mu_2)^2}\right)
%\end{align}
\begin{equation}\label{dis-1}
\tilde{X} = \frac{X - E[Z]}{\sqrt{D[Z]}}\sim N(\tilde{\mu}, \tilde{\sigma}^2),
\end{equation}
%\end{equation}
and
\begin{equation}\label{dis-2}
\tilde{Y} = \frac{Y - E[Z]}{\sqrt{D[Z]}} \sim N(-\tilde{\mu}, \tilde{\sigma}^2),
\end{equation}
with
$$
\tilde{\mu}=\frac{(\mu_1 - \mu_2)/2}{\sqrt{\sigma^2 + \frac{1}{4}(\mu_1 - \mu_2)^2}}, \quad
\tilde{\sigma}^2=\frac{\sigma^2}{\sigma^2 + \frac{1}{4}(\mu_1 - \mu_2)^2},
$$
%\begin{equation}\label{dis-2}
%	\tilde{Y} = \frac{Y - E[Z]}{\sqrt{D[Z]}} \sim N\left(\frac{-(\mu_1 - \mu_2)/2}{\sqrt{\sigma^2 + \frac{1}{4}(\mu_1 - \mu_2)^2}}, \frac{\sigma^2}{\sigma^2 + \frac{1}{4}(\mu_1 - \mu_2)^2}\right)
%\end{equation}
where $E[Z]$ and $D[Z]$ denote the expectation and variance of $Z$, which have expressions $E[Z]=\frac{1}{2}(\mu_1 + \mu_2)$ and  $D[Z]=\sigma^2 + \frac{1}{4}(\mu_1 - \mu_2)^2$.

From Equations \eqref{dis-1} and \eqref{dis-2}, it can be seen that after standardization, the two classes of variables $\tilde{X}$ and $\tilde{Y}$ still exhibit Gaussian distributions, but showcase two interesting properties: 1) their means are symmetric about zero; and 2) they have the sum of the square of the mean and the variance equal to one.   In the paper, we  will  focus our study on such standardized data, which can be characterized as follows.

\begin{property}[Distributions of two classes of standardized data]\label{data-distribution}  Consider a dataset comprising two classes of data, each sampled equally from two Gaussian distributions with distinct means but identical variances. Upon standardization, the two classes of data will follow the distributions  $X \sim N(\mu, \sigma^2)$ and $Y \sim N(-\mu, \sigma^2)$, respectively, where the relationship $\mu^2 + \sigma^2 = 1$ holds, with $\mu \in(0,1)$.
\end{property}

%Then $Z$  has the probability density function $p(z)= \frac{1}{2}N(z;\mu, \sigma^2)+\frac{1}{2}N(z;-\mu, \sigma^2)$, where $\mu^2 + \sigma^2 = 1$ and $0 < \mu < 1$.

\subsection{Feature discrimination}

Following the Fisher's linear discriminant rule,  we define the discrimination between two classes of data as the ratio of the expected inter-class distance to the expected intra-class distance, as specified below.
\begin{definition}[Discrimination between two classes of data]\label{def-discrimination} For two classes of data with  samples respectively drawn from the variables $X$ and $Y$, the discrimination between them is defined as
	\begin{equation}\label{d-1}
		D = \frac{E[(X_1 - Y_1)^2]}{E[(X_1 - X_2)^2] + E[(Y_1 - Y_2)^2]} \quad
	\end{equation}
	where \(X_1\) and \(X_2\) are i.i.d. samples of \(X\), and \(Y_1\) and \(Y_2\) are i.i.d samples of \(Y\).
	
\end{definition}
In the sequel, we will utilize the above definition $D$  to  denote the discrimination between original (non-quantized) data; and for the binary and ternary quantized data, as detailed below, we adopt $D_b$ and $D_t$ to represent their discrimination.

\begin{definition}[Discrimination between two classes of quantized data]\label{def-discrimination-Q} Following the discrimination  specified in Definition \ref{def-discrimination},   the discrimination between two binary quantized data $X_b=f_b(X;\tau)$ and  $Y_b=f_b(Y;\tau)$, is formulated as
	\begin{equation}\label{d-B}
		D_b = \frac{E[(X_{1,b} - Y_{1,b})^2]}{E[(X_{1,b} - X_{2,b})^2] + E[(Y_{1,b} - Y_{2,b})^2]} \quad
	\end{equation}
	where $X_{1,b}$ and $X_{2,b}$ are i.i.d. samples of $X_b$, and $Y_{1,b}$ and $Y_{2,b}$  are i.i.d. samples of $Y_b$. Similarly, the discrimination between two ternary quantized data $X_t=f_t(X;\tau)$ and  $Y_t=f_t(Y;\tau)$  is expressed as	
	\begin{equation}\label{d-T}
		D_t = \frac{E[(X_{1,t} - Y_{1,t})^2]}{E[(X_{1,t} - X_{2,t})^2] + E[(Y_{1,t} - Y_{2,t})^2]} \quad
	\end{equation}
	where $X_{1,t}$ and $X_{2,t}$ are i.i.d. samples of $X_t$, and $Y_{1,t}$ and $Y_{2,t}$  are i.i.d. samples of $Y_t$.
	
\end{definition}

\subsection{Goal}

The major goal of the paper  is to investigate whether there exist  threshold values $\tau$ in the binary quantization $f_b(x;\tau)$ and the ternary quantization $f_t(x;\tau)$, such that the quantization can enhance  the feature discrimination of original data, namely having $D_b>D$ and $D_t>D$.

%As stated before,  the major goal of the paper is to explore whether the binary or ternary quantization can enhance the discrimination between two classes of data. Formally, this requires us to investigate if there exist  threshold values $\tau$ in the quantization function $f_b(x;\tau)$ or $f_t(x;\tau)$, such that the discrimination between quantized data is higher than the discrimination between original data, namely having $D_b>D$ or $D_t>D$.

\section{Discrimination Analysis} \label{sec-DiscriminationAnalysis}
\subsection{Theoretical results}

\begin{theorem}[Binary quantization]\label{prop:bq} Consider the discrimination $D$ between two classes of data \(X \sim N(\mu, \sigma^2)\) and \(Y \sim N(-\mu, \sigma^2)\) as specified in Property \ref{data-distribution}, as well as the discrimination $D_b$
 between their binary quantization $X_b=f_b(X;\tau)$ and  $Y_b=f_b(Y;\tau)$. We have   \( D_b > D \), if there exists a quantization threshold  $\tau \in (-\infty, +\infty)$ such that
	\begin{equation} \label{eq-bq}
		\beta - \alpha + \frac{\mu^2(1-2\beta) - \mu\sqrt{\mu^2 + 4\beta(1-\beta)}}{1 + \mu^2} > 0,
	\end{equation}
	where \(\alpha = \Phi\left(\frac{\tau - \mu}{\sigma}\right)\) and \(\beta = \Phi\left(\frac{\tau + \mu}{\sigma}\right)\), with  \(\Phi(\cdot)\) denoting the cumulative distribution function of the standard normal distribution.
\end{theorem}

\begin{theorem}[Ternary quantization]\label{prop:tq} Consider the discrimination $D$ between two classes of data  \( X \sim N(\mu, \sigma^2) \) and \( Y \sim N(-\mu, \sigma^2) \) as specified in Property \ref{data-distribution}, as well as the discrimination $D_t$ between their ternary quantization  $X_t=f_t(X;\tau)$ and  $Y_t=f_t(Y;\tau)$. We have  \( D_t > D \), if there exists a quantization threshold  $\tau \in [0, +\infty)$ such that
	\begin{equation} \label{eq-tq}
		\beta - \alpha + \frac{\mu^2 -\sqrt{\mu^4 + 8\mu^2\beta}}{2} > 0,
	\end{equation}
	where \( \alpha = \Phi\left(\frac{-\tau - \mu}{\sigma}\right) \) and \( \beta = \Phi\left(\frac{-\tau + \mu}{\sigma}\right) \), with  \( \Phi(\cdot) \) denoting the cumulative distribution function of the standard normal distribution.
\end{theorem}

\noindent\textbf{Remarks:} Regarding the  two theorems, there are three issues worth discussing. 1) The two theorems suggest that both binary and ternary quantization methods can indeed enhance the classification performance of original data, if there exist quantization thresholds $\tau$ that can satisfy the constraints shown in Equations \eqref{eq-bq} and \eqref{eq-tq}.  The following numerical analysis demonstrates that the desired threshold $\tau$ does exist, when the two classes of data $X \sim N(\mu, \sigma^2)$ and $Y \sim N(-\mu, \sigma^2)$ are assigned appropriate values for $\mu$ and $\sigma$. This threshold $\tau$ can be typically determined using gradient descent methods, as detailed in Appendix \ref{sec-algorithm}. 2) Our theoretical analysis is based on the premise that  the data vectors within the same class follow Gaussian distributions in each dimension of the vectors. This condition should hold true when  two classes of data are readily separable, as in this case the data points within each class should cluster tightly, allowing for Gaussian approximation. This explains the recent research findings, that the binary or ternary quantization tends to achieve comparable or superior classification performance,  when handling relatively simple datasets \citep{courbariaux2015binaryconnect, Zhouhan2016Neural}, or distinguishable features \citep{lu2023quantization}. 3) The conclusion we derive in Theorem \ref{prop:bq} for $\{0, 1\}$-binary quantization also applies to another popular $\{-1,1\}$-binary quantization \citep{qin2020binary}, since the Euclidean distance of the former is equivalent to the cosine distance of the latter.

%This result explains why quantized data sometimes exhibit better classification performance than the original data in classification tasks.

\subsection{Numerical validation}

\begin{figure*}[ht]
	\centering
\begin{minipage}{\textwidth}
\centering
		\begin{subfigure}[b]{0.35\textwidth}
			\includegraphics[width=\textwidth]{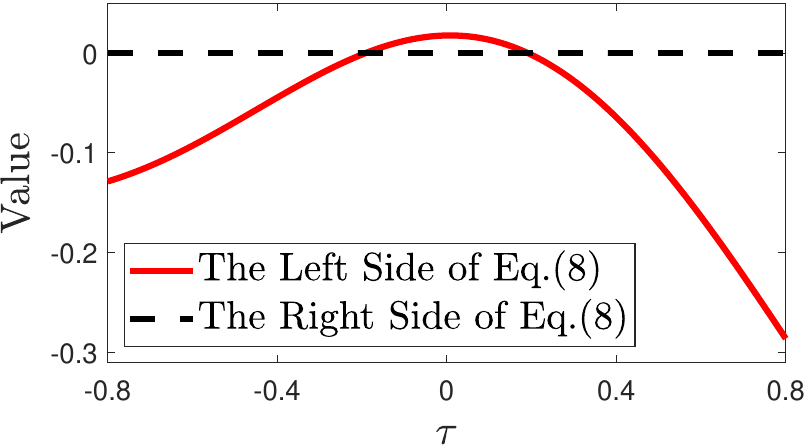}
			\caption{\centering \footnotesize Theoretical results for binary quantization}
		\end{subfigure}
		\begin{subfigure}[b]{0.35\textwidth}
			\includegraphics[width=\textwidth]{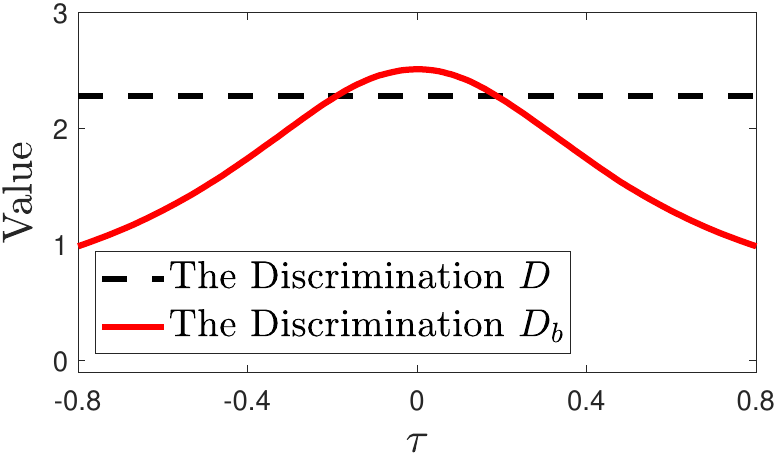}
			\caption{\centering \footnotesize Numerical results for binary quantization}
		\end{subfigure}
\end{minipage}
\begin{minipage}{\textwidth}
\centering
		\begin{subfigure}[b]{0.35\textwidth}
			\includegraphics[width=\textwidth]{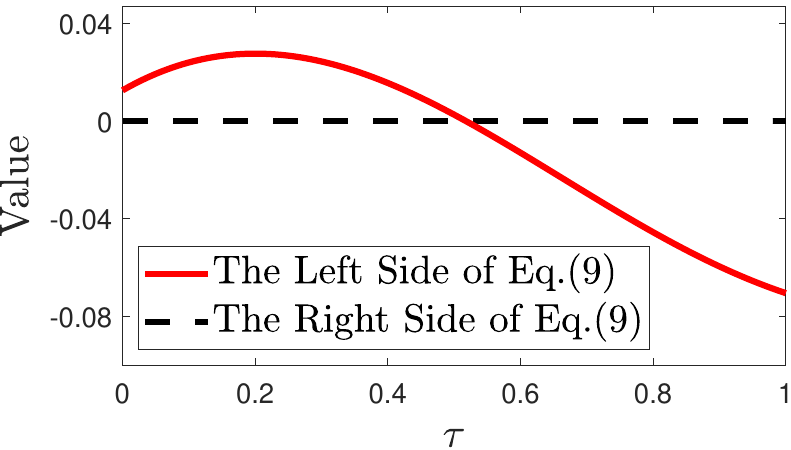}
			\caption{\centering \footnotesize Theoretical results for ternary quantization}
		\end{subfigure}
		\begin{subfigure}[b]{0.35\textwidth}
			\includegraphics[width=\textwidth]{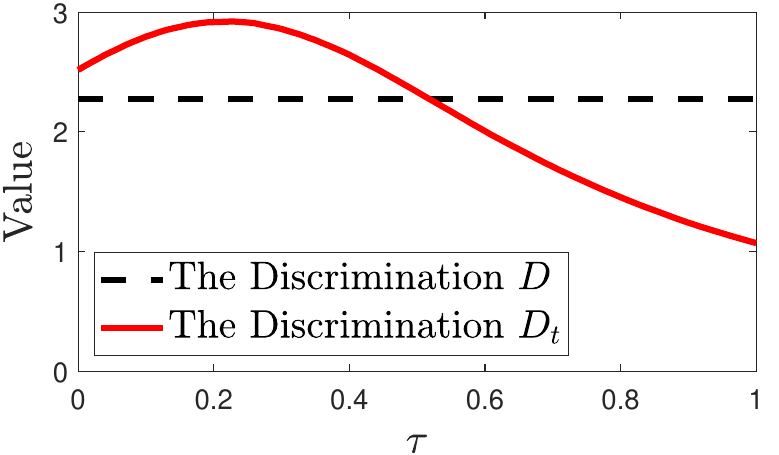}
			\caption{\centering \footnotesize Numerical results for ternary quantization}
		\end{subfigure}
\end{minipage}
	\captionsetup{font=normalsize}
	\caption{Consider two classes of data $X\sim N(\mu, \sigma^2)$ and $Y\sim N(-\mu, \sigma^2)$, with  $\mu=0.8$ and $\sigma^2=0.36$, as specified in Property \ref{data-distribution}. The values for the left and right sides of Equations \eqref{eq-bq} and \eqref{eq-tq} are provided in (a) and (c) for binary and ternary quantization, respectively; and the discrimination $D$, $D_b$ and $D_t$ statistically estimated with Equations \eqref{d-1}, \eqref{d-B} and \eqref{d-T} are illustrated in (b) and (d) for  binary and ternary quantization, respectively.}
\label{fig:Numerical simu-1}
\end{figure*}

In this part, we conduct numerical analyses for two primary objectives. Firstly, we aim to prove the existence of the desired quantization threshold $\tau$ that holds  Equations \eqref{eq-bq} and \eqref{eq-tq}, namely making  the left sides of the two inequalities larger than their right sides (with values equal to zero). For this purpose, we compute the values of the left sides of Equations \eqref{eq-bq} and \eqref{eq-tq}, through assigning specific values to $\tau$, as well as to the two variables $X$ and $Y$'s distribution parameters $\mu$ and $\sigma^2$. Note that we here set $\sigma^2 = 1-\mu^2$, $\mu \in(0,1)$, in accordance with Property \ref{data-distribution}. In Figure \ref{fig:Numerical simu-1}, we examine the case that fixes   $\mu=0.8$ and $\sigma^2=0.36$, while varying the value of $\tau$ with a step width 0.01. The results for binary quantization and ternary quantization are provided in Figures \ref{fig:Numerical simu-1} (a) and (c), respectively. It can be seen that for the two quantization methods,  their conditions shown in Equations \eqref{eq-bq} and \eqref{eq-tq}   will hold when respectively having  $\tau\in[-0.2, 0.2]$ and $\tau\in[0, 0.5]$. This proves the existence of the desired quantization threshold $\tau$ that can enhance feature discrimination. For limited space, we here only discuss the case  of $\mu=0.8$ (and $\sigma^2=1-\mu^2$)  in Figure  \ref{fig:Numerical simu-1}. By examining  different $\mu\in(0,1)$ in the same way, we can find that the quantization threshold $\tau$ that holds Equations \eqref{eq-bq} and \eqref{eq-tq}, is present  when $\mu\in(0.76,1)$ and $\mu\in(0.66,1)$, respectively; see Figures \ref{fig:Numerical simu-2} and \ref{fig:Numerical simu-3} for more evidences. This result implies two consequences. On one hand, ternary quantization has more chances to enhance feature discrimination compared to binary quantization, as the former has a broader range of $\mu$.  On the other hand, the enhanced discrimination tends to be achieved when  $\mu$ is sufficiently large, coupled with a correspondingly small $\sigma$, or when the  discrimination between two classes of data is sufficiently high.

%\textcolor{blue}{Empirically, as depicted in Figure \ref{fig:real data-density}, the two specific ranges of $\mu$ values are attainable for the commonly-used features of real data.}

%and at  the two lower boundary points $u=0.76$ and $0.66$, as shown in Figures \ref{fig:Numerical simu-2} and \ref{fig:Numerical simu-3},   quantization can achieve similar discrimination  with the original data.

The second goal is to verify that the quantization thresholds $\tau$ we estimate with Equations \eqref{eq-bq} and \eqref{eq-tq}  in Theorems \ref{prop:bq} and  \ref{prop:tq}, can indeed  enhance feature discrimination.  To this end, it needs to prove that the ranges of $\tau$  derived by Equations \eqref{eq-bq} and \eqref{eq-tq}, such as the ones depicted in Figures \ref{fig:Numerical simu-1} (a) and (c),  are consistent with the ranges we can statistically estimate by the discrimination definitions $D$, $D_b$ and $D_t$,  as specified in Definitions  \ref{def-discrimination}  and \ref{def-discrimination-Q}.  To estimate the discrimination  $D$, $D_b$ and $D_t$, we randomly generate 10,000 samples from $X \sim N(0.8, 0.36)$ and $Y \sim N(-0.8, 0.36)$, respectively, and then statistically estimate them with Equations \eqref{d-1}, \eqref{d-B} and \eqref{d-T}, across varying $\tau$ (with a step width 0.01). The results are provided in Figures \ref{fig:Numerical simu-1} (b) and (d), respectively for  binary  and ternary quantization.  It can be seen that we have $\tau\in[-0.2, 0.2]$ for $D_b>D$, and have   $\tau\in[0, 0.5]$ for $D_t>D$. The results coincide with the theoretical results shown in Figures \ref{fig:Numerical simu-1} (a) and (c), validating  Theorems \ref{prop:bq} and  \ref{prop:tq}.

\section{Experiments}

%By the previous theoretical and numerical analysis, we have proved that the binary and ternary quantization can enhance the feature discrimination between two classes of data, when the data vectors of each class have Gaussian distributions at their feature vectors' each dimension point.  Considering the enhanced feature discrimination should yield improved classification performance, in this section we aim to prove the enhanced feature discrimination by evaluating the classification performance. For the sake of generality, we conduct  binary classification experiments both on synthetic data and real data. The synthetic data can  satisfy the distribution conditions required by our theoretical analysis, while the real data usually cannot.

Through previous theoretical and numerical analyses, we have demonstrated that binary and ternary quantization can enhance feature discrimination between two classes of data, when the data vectors within each class exhibit Gaussian distributions across each dimension  of their feature vectors. As enhanced feature discrimination is expected to improve classification performance, this section aims to confirm this improvement by evaluating classification performance. To see the generalizability of our theoretical findings, we evaluate  classification performance on both  synthetic  and real data. Synthetic data can accurately adhere to the distribution conditions specified in our theoretical analysis, whereas real-world data typically cannot.

\subsection{Synthetic data}
\label{sec:synthetic data}
\subsubsection{Setting}

\paragraph{Classification.} In this section,  we mainly examine the binary classification, as it directly reflects the feature discrimination between two classes, enabling the validation of our theoretical findings. We will employ two fundamental classifiers: the $k$-nearest neighbors (KNN) algorithm (with $k=5$) \citep{peterson2009k}, utilizing both Euclidean and cosine distances as similarity metrics, and the support vector machine (SVM) \citep{cortes1995support}, equipped with a linear kernel.

% In this part, we will primarily examine binary classification using two fundamental classifiers: the $k$-nearest neighbors (KNN) algorithm (with $k=5$) \citep{peterson2009k}, utilizing both Euclidean and cosine distances as similarity metrics, and the support vector machine (SVM) \citep{cortes1995support}, equipped with a linear kernel. We focus on   binary classification mainly because it  can directly reflect the feature discrimination between two classes and then validate our theoretical results.

\paragraph{Data generation.} We generate  two classes of data by randomly and independently   drawing the samples  from two different random vectors $\mathbf{X}=\{X_1,X_2,\cdots,X_n\}^\top$ and $\mathbf{Y}=\{Y_1,Y_2,\cdots,Y_n\}^\top$, for which we set  $X_i\sim N(\mu_{i}, \sigma_i^2)$ and $Y_i\sim N(-\mu_{i},  \sigma_i^2)$, with $\mu_i\in(-1,0)\cup(0,1)$ and $\sigma_i^2=1-\mu_i^2$, according to the data distributions specified in Property \ref{data-distribution}.  Considering the fact that the features of real-world data usually exhibit sparse structures \citep{weiss2007makes, kotz2012laplace}, we further suppose that the means $\mu_i$ decay exponentially in magnitude, i.e. $|\mu_{i+1}|/|\mu_{i}|=e^{-\lambda}$, $\lambda\geq 0$, and set $\mu_1=0.8$ in the following simulation. It can be seen that with the increasing of $\lambda$, the mean's magnitude $|\mu_i|$ (with $i>1$)  will become smaller, indicating a smaller data element $X_i$ (in magnitude) and a sparser data structure. However,  the data element  $X_i$ with smaller $\mu_i$, is not favorable for quantization to enhance feature discrimination, as indicated by  previous numerical analyses. The impact of data sparsity  on quantization can be investigated by increasing the value of the  parameter $\lambda$.

%By this fact, we can challenge the performance of quantization by increasing the value of the  parameter $\lambda$, namely increasing the data sparsity level.
% This part also should discuss dimension 数据维度的影响

With the data model described above, we randomly generate two classes of data, each class containing 1000 samples. The dataset is split into two parts for training and testing, in a ratio of 4:1. Then we evaluate the KNN and SVM classification  on them. The classification accuracy is determined by averaging the accuracy results obtained from repeating the data generation and classification process 100 times. The major results for KNN with Euclidean distance are presented in the main text, as shown in Figures \ref{fig:synthetic-lambda knn-Eucl} to \ref{fig:Quantization_Error}; and other results, such as  those for KNN with cosine distance and SVM, are given in Appendix \ref{sec-other_experiments}, specifically in Figures \ref{fig:synthetic-lambda knn-Cos} to \ref{fig:synthetic svm}. It can be seen that the three classifiers exhibit similar performance trends. For brevity, we will focus more on the results of KNN with Euclidean distance in the subsequent discussion.
\subsubsection{Results}

\begin{figure*}[t]
	\centering
	\begin{minipage}{\textwidth}
		\centering	
		\begin{subfigure}[b]{0.24\textwidth}
			\includegraphics[width=\textwidth]{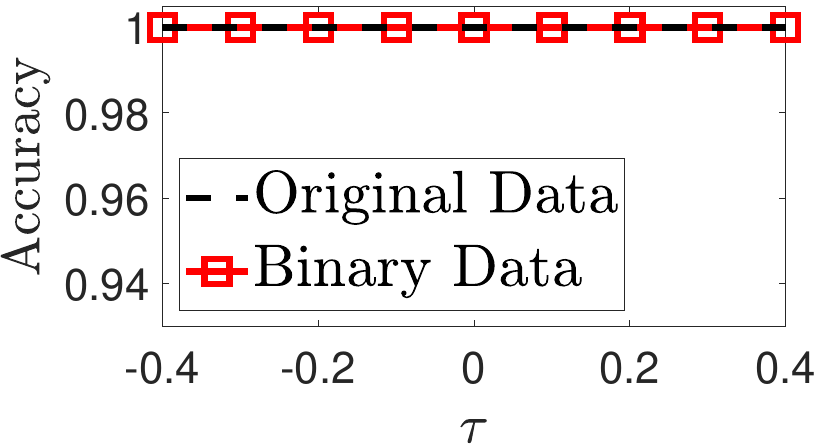}
			\caption{\scriptsize Binary data ($\lambda=0$)}
		\end{subfigure}
		\begin{subfigure}[b]{0.24\textwidth}
			\includegraphics[width=\textwidth]{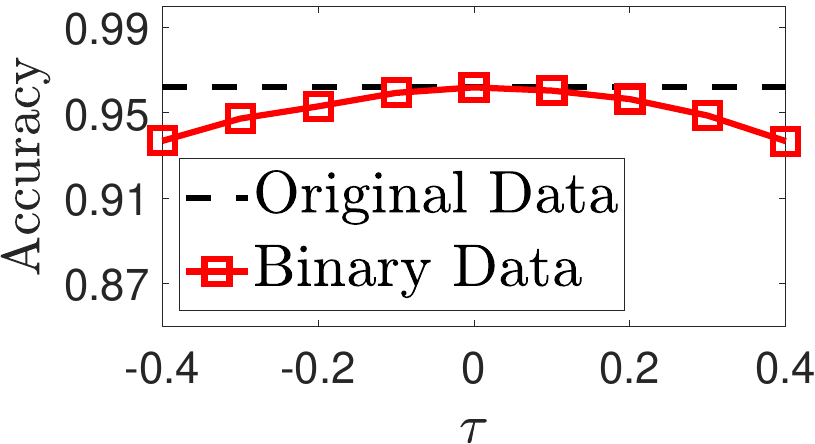}
			\caption{\scriptsize Binary data ($\lambda=0.01$)}
		\end{subfigure}
		\begin{subfigure}[b]{0.24\textwidth}
			\includegraphics[width=\textwidth]{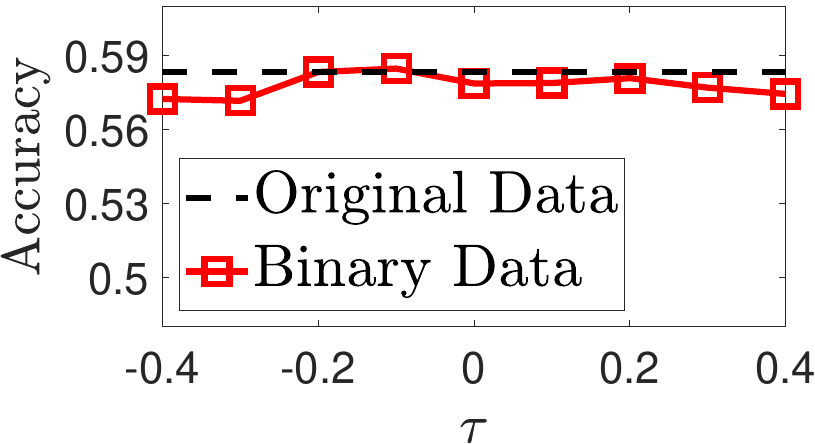}
			\caption{\scriptsize Binary data ($\lambda=0.1$)}
		\end{subfigure}
		\begin{subfigure}[b]{0.24\textwidth}
			\includegraphics[width=\textwidth]{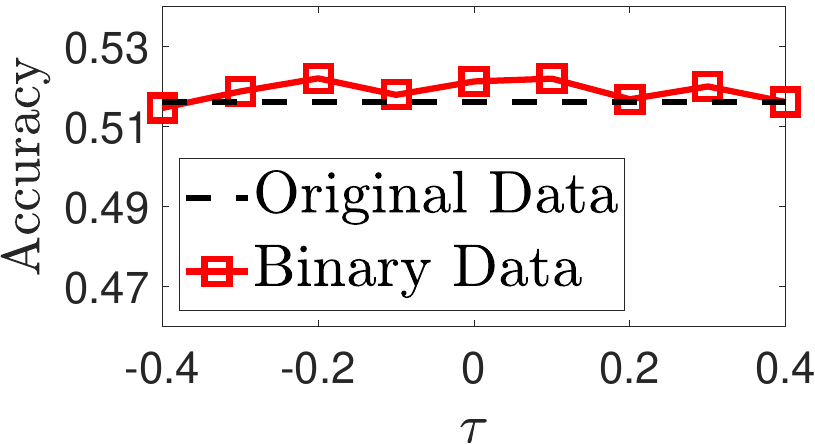}
			\caption{\scriptsize Binary data ($\lambda=1$)}
		\end{subfigure}
	\end{minipage}
	\begin{minipage}{\textwidth}
		\centering
		\begin{subfigure}[b]{0.24\textwidth}
			\includegraphics[width=\textwidth]{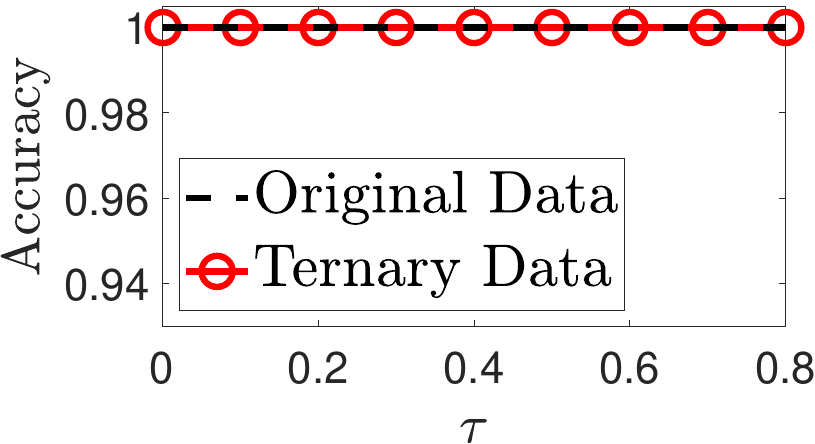}
			\caption{\scriptsize Ternary data ($\lambda=0$)}
		\end{subfigure}
		\begin{subfigure}[b]{0.24\textwidth}
			\includegraphics[width=\textwidth]{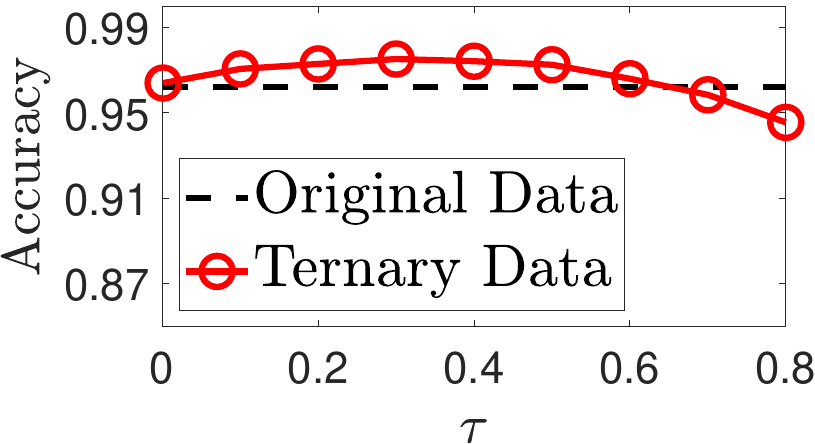}
			\caption{\scriptsize Ternary data ($\lambda=0.01$)}
		\end{subfigure}
		\begin{subfigure}[b]{0.24\textwidth}
			\includegraphics[width=\textwidth]{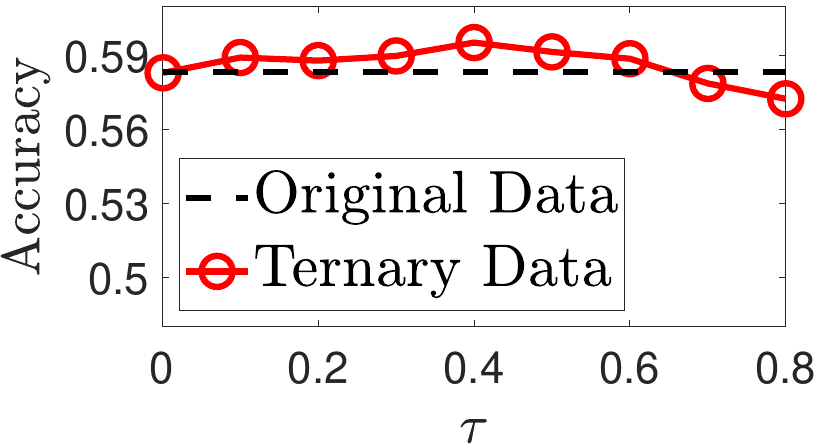}
			\caption{\scriptsize Ternary data ($\lambda=0.1$)}
		\end{subfigure}
		\begin{subfigure}[b]{0.24\textwidth}
			\includegraphics[width=\textwidth]{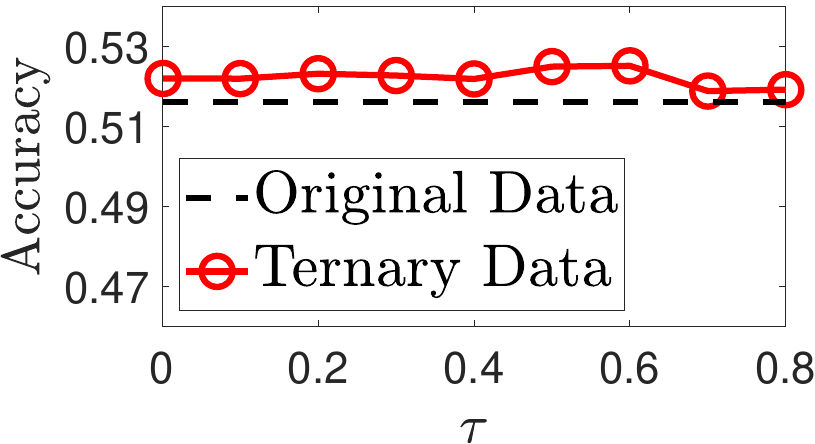}
			\caption{\scriptsize Ternary data ($\lambda=1$)}
		\end{subfigure}

	\end{minipage}
	\caption{KNN (Euclidean distance) classification accuracy for the 10,000-dimensional binary, ternary, and original data that are generated with  the varying  parameter $\lambda\in\{0,0.01,0.1,1\}$, which controls the data sparsity.}
	\captionsetup{font=normalsize}
	\label{fig:synthetic-lambda knn-Eucl}
\end{figure*}

\begin{figure*}[!ht]
	\centering
	\begin{minipage}{\textwidth}	
	\centering
		\begin{subfigure}[b]{0.32\textwidth}
			\includegraphics[width=\textwidth]{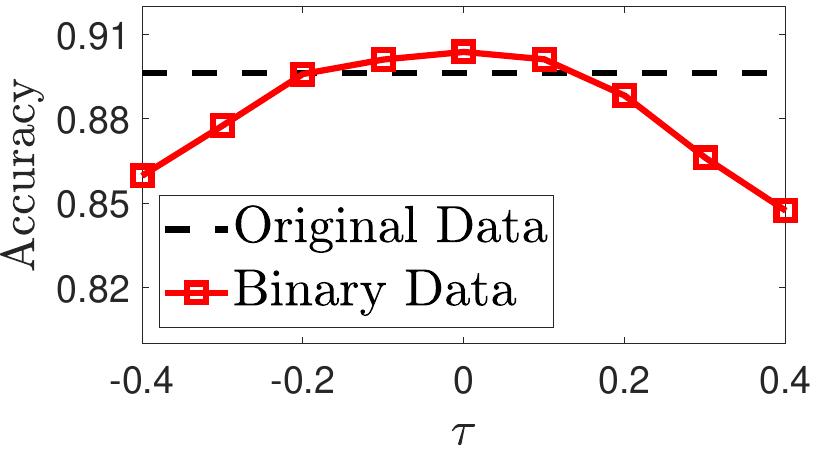}
			\caption{\footnotesize 1-dim binary data}
		\end{subfigure}
		\begin{subfigure}[b]{0.32\textwidth}
			\includegraphics[width=\textwidth]{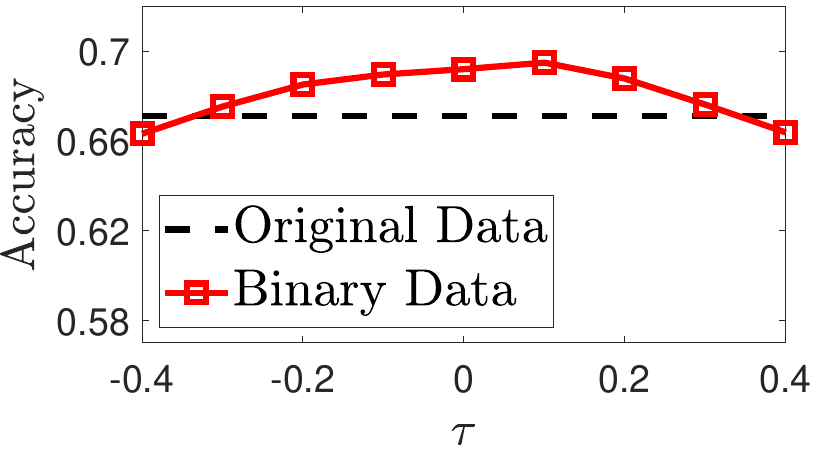}
			\caption{\footnotesize 100-dim binary data}
		\end{subfigure}
		\begin{subfigure}[b]{0.32\textwidth}
			\includegraphics[width=\textwidth]{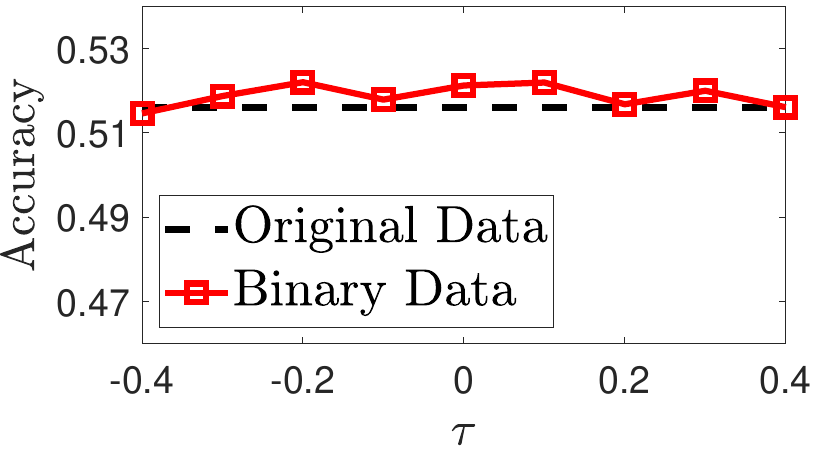}
			\caption{\footnotesize 10,000-dim binary data}
		\end{subfigure}
	\end{minipage}
	\begin{minipage}{\textwidth}
     \centering
		\begin{subfigure}[b]{0.32\textwidth}
			\includegraphics[width=\textwidth]{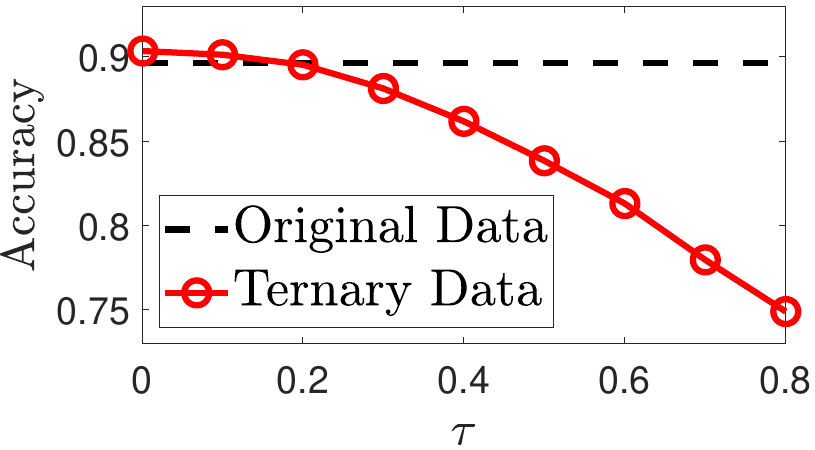}
			\caption{\footnotesize 1-dim ternary data}
		\end{subfigure}
		\begin{subfigure}[b]{0.32\textwidth}
			\includegraphics[width=\textwidth]{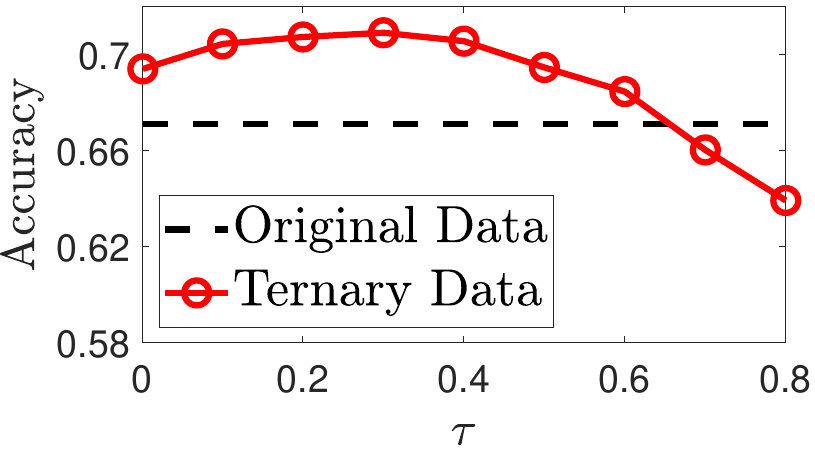}
			\caption{\footnotesize 100-dim ternary data}
		\end{subfigure}
		\begin{subfigure}[b]{0.32\textwidth}
			\includegraphics[width=\textwidth]{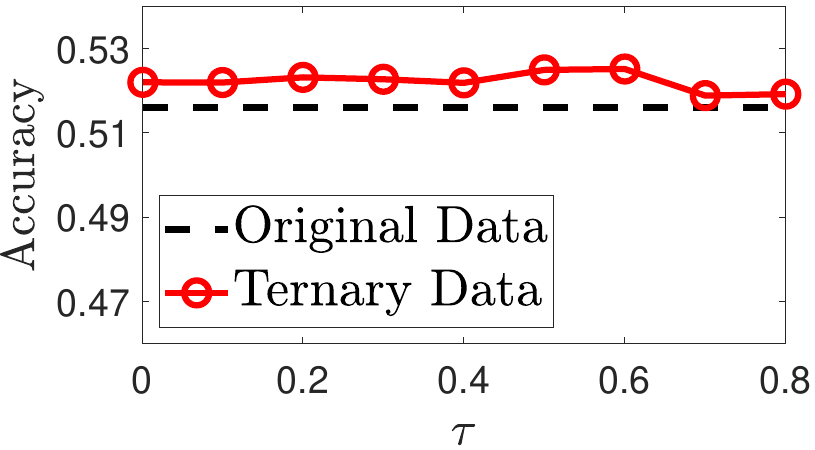}
			\caption{\footnotesize 10,000-dim ternary data}
		\end{subfigure}
	\end{minipage}
	\caption{KNN (Euclidean distance) classification accuracy for the binary, ternary, and original data generated with the parameter $\lambda=1$, and with varying dimensions $n\in\{1,100,10000\}$.}
	\captionsetup{font=normalsize}
	\label{fig:synthetic knn-Eucl}
	
\end{figure*}

%\textbf{Comparison between the data with different sparsity.} In Figure \ref{fig:synthetic-lambda knn-Eucl}, we  investigate   the classification performance for the data generated with different parameters $\lambda\in\{0, 0.1, 1, 10\}$, namely with different sparsity levels. Recall that the larger the $\lambda$, the smaller the $|\mu_i|$, or say the sparser the data vector. Based on the previous analysis, the data vector with smaller $|\mu_i|$ is not conducive to enhancing feature discrimination through quantization. Nevertheless, empirically, the negative effect does not appear to be significant. For instance, even when $\lambda=10$, which corresponds to a very sparse data vector with $|\mu_i|$ decaying at a rate of approximately $e^{-10}\approx 1/22000$, there still exist quantization thresholds $\tau$ that can yield better classification performance than the original data.  This suggests that  quantization can be effectively applied to sparse data to enhance feature discrimination. Given the prevalence of sparse data \citep{weiss2007makes, kotz2012laplace},  this implies a broad application of quantization.

\paragraph{Comparison between the data with different sparsity.} In Figure \ref{fig:synthetic-lambda knn-Eucl}, we  investigate   the classification performance for the data generated with different parameters $\lambda\in\{0, 0.01, 0.1, 1\}$, namely with different sparsity levels. Recall that the larger the $\lambda$, the smaller the $|\mu_i|$, or say the smaller the data element $X_i$ (in magnitude). By previous analyses, the data element $X_i$ with smaller $|\mu_i|$ is not conducive to enhancing feature discrimination through quantization. Nevertheless, empirically, the negative effect does not appear to be significant. From Figure \ref{fig:synthetic-lambda knn-Eucl}, it can be seen that when increasing $\lambda$ from $0.1$ to $1$, there have been quantization thresholds $\tau$ that can yield better classification performance than the original, full-precision data. In addition, it noteworthy  that as $\lambda$ increases,  the overall classification accuracy of original data will decrease. This decreasing trend also impacts the absolute performance of the quantized data, even though it may outperform original data.

%\textbf{Comparison between the data with different sparsity.} In Figure \ref{fig:synthetic-lambda knn-Eucl}, we  investigate   the classification performance for the data generated with different parameters $\lambda\in\{0, 0.1, 1, 10\}$, namely with different sparsity levels. Recall that the larger the $\lambda$, the smaller the $|\mu_i|$, or say the sparser the data vector. Based on the previous analysis, the data vector with smaller $|\mu_i|$ is not conducive to enhancing feature discrimination through quantization. Nevertheless, empirically, the negative effect does not appear to be significant. For instance, even when $\lambda=10$, which corresponds to a very sparse data vector with $|\mu_i|$ decaying at a rate of approximately $e^{-10}\approx 1/22000$, there still exist quantization thresholds $\tau$ that can yield better classification performance than the original data.  This suggests that  quantization can be effectively applied to sparse data to enhance feature discrimination. Given the prevalence of sparse data \citep{weiss2007makes, kotz2012laplace},  this implies a broad application of quantization.

\paragraph{Comparison between the data with different dimensions.} The impact of data dimensions $n\in\{1,100,10000\}$ on classification is investigated in Figure \ref{fig:synthetic knn-Eucl}, where the data are generated with the exponentially decaying parameter $\lambda=1$. It can be seen that with the increasing of data dimension, the range of the quantization thresholds $\tau$ that outperform  original data tends to expand, but the performance advantage declines. As previously discussed, the decline should be attributed to the data element $X_i$ with small means $|\mu_i|$, whose quantity will rise with the data dimension $n$, particularly when the decay parameter $\lambda$ of $|\mu_i|$ is large. To alleviate this adverse effect, it is recommended to choose a relatively smaller $\lambda$  for high-dimensional data, indicating a structure that is not overly sparse. Conversely, when the high-dimensional data is highly sparse, we should reduce its dimension to enhance the classification performance under quantization.

\paragraph{Comparison between binary quantization and ternary quantization.} From Figures \ref{fig:synthetic-lambda knn-Eucl} and \ref{fig:synthetic knn-Eucl}, it can be seen that ternary quantization surpasses binary quantization by offering  broader ranges of quantization thresholds $\tau$ that can yield higher  classification accuracy than original data. This observation is consistent with our previous theoretical and numerical analyses.

\paragraph{Comparison between KNN and SVM.} Combining the results  in Figures \ref{fig:synthetic-lambda knn-Eucl}, \ref{fig:synthetic knn-Eucl}, and \ref{fig:synthetic-lambda knn-Cos}--\ref{fig:synthetic svm}, we can say that both KNN and SVM enable quantization to improve the classification accuracy of original data, within specific ranges of quantization thresholds $\tau$. If closely examining these ranges, it can be observed that  KNN often performs  better when using Euclidean distance than using cosine distance. This  can be attributed to the advantage of Euclidean distance over cosine distance in measuring the distance between $0$ and $\pm 1$. Also, KNN often outperforms SVM, such as the case of $\lambda=0.1$  shown in Figures \ref{fig:synthetic-lambda knn-Eucl} and \ref{fig:synthetic-lambda svm}. This is because the support vector of SVM relies on a few data points located on the boundary between two classes, which may deteriorate during quantization. In contrast, KNN depends on the high-quality data points within each class, making it  resilient to quantization noise.

\begin{figure*}[t!]
	\centering
	\begin{minipage}{\textwidth}	
	\centering
		\begin{subfigure}[b]{0.32\textwidth}
			\includegraphics[width=\textwidth]{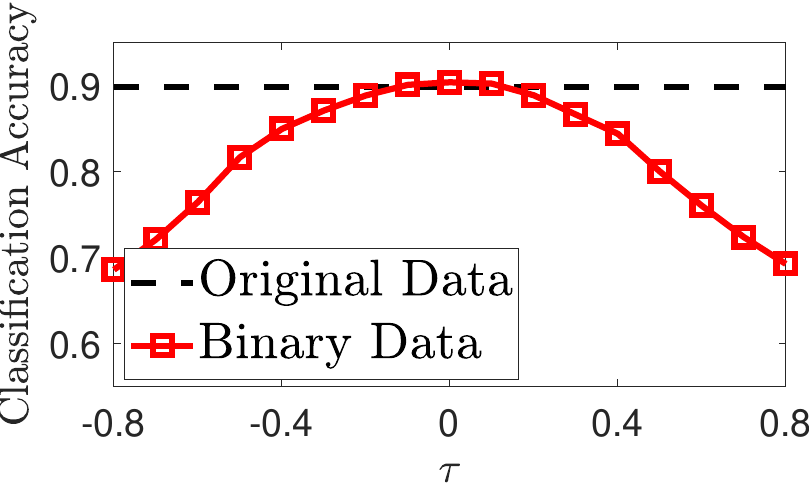}
			\caption{\footnotesize Classification accuracy (B)}
		\end{subfigure}
		\begin{subfigure}[b]{0.32\textwidth}
			\includegraphics[width=\textwidth]{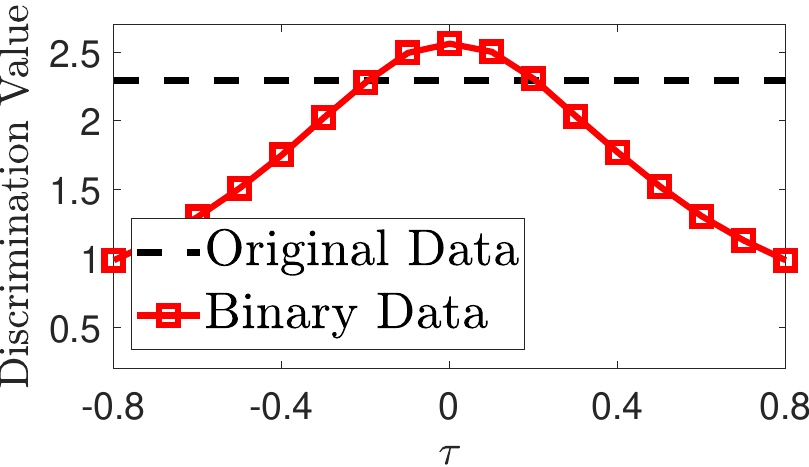}
			\caption{\footnotesize Discrimination value (B)}
		\end{subfigure}
		\begin{subfigure}[b]{0.32\textwidth}
			\includegraphics[width=\textwidth]{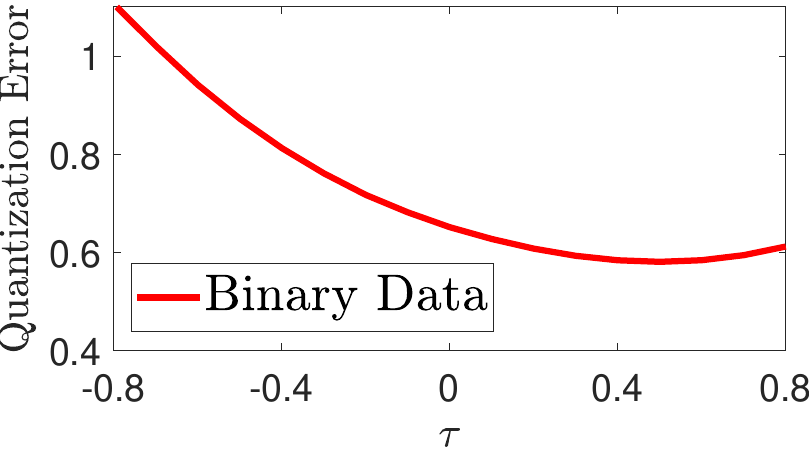}
			\caption{\footnotesize Quantization error (B)}
		\end{subfigure}
	\end{minipage}
	\begin{minipage}{\textwidth}
     \centering
		\begin{subfigure}[b]{0.32\textwidth}
			\includegraphics[width=\textwidth]{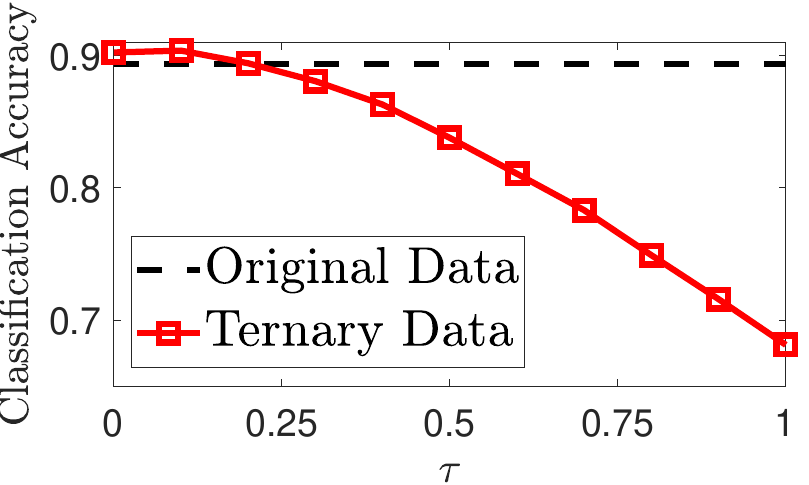}
			\caption{\footnotesize Classification accuracy (T)}
		\end{subfigure}
		\begin{subfigure}[b]{0.32\textwidth}
			\includegraphics[width=\textwidth]{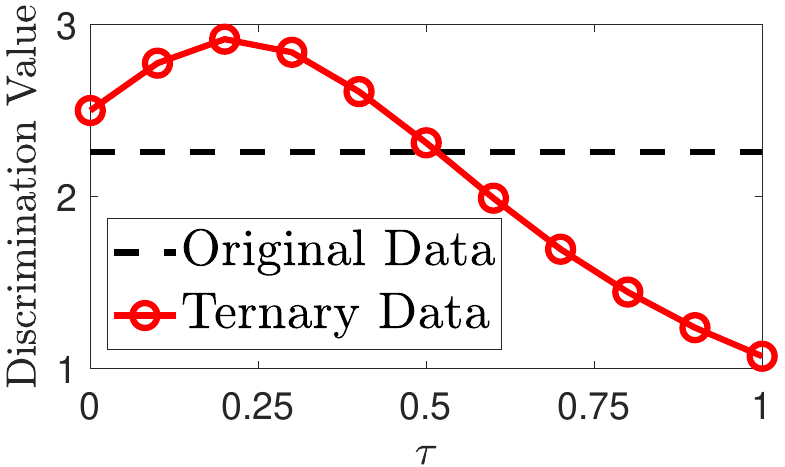}
			\caption{\footnotesize Discrimination value (T)}
		\end{subfigure}
		\begin{subfigure}[b]{0.32\textwidth}
			\includegraphics[width=\textwidth]{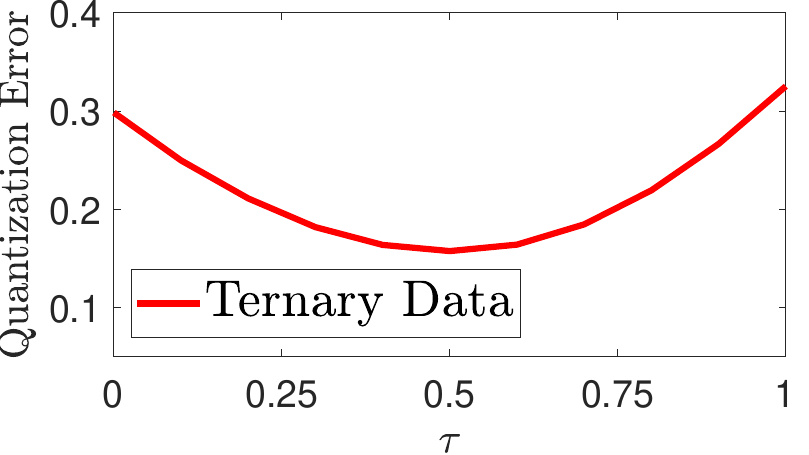}
			\caption{\footnotesize Quantization error (T)}
		\end{subfigure}
	\end{minipage}
	\caption{ KNN (Euclidean distance) classification accuracy, feature discrimination values and quantization errors are provided for the synthetic data   generated with the decay parameter $\lambda=1$ and data dimension $n=1$, as well as for their binary and ternary quantization derived  across different  thresholds $\tau$.}
	\captionsetup{font=normalsize}
	\label{fig:Quantization_Error}
	
\end{figure*}

\paragraph{Comparison between classification accuracy, feature discrimination and quantization error.}
 As theoretically predicted, Figure \ref{fig:Quantization_Error} illustrates that  higher classification accuracy tends to be achieved at quantization thresholds $\tau$ that result in greater feature discrimination values, rather than smaller quantization errors. In particular, the variation in classification accuracy with respect to $\tau$ is more closely associated with feature discrimination values  than with quantization errors. This confirms the superiority of our feature discrimination analysis in assessing the classification performance of quantized data, compared to the traditional approach of using quantization errors.

\subsection{Real data}

\label{sec:real data}

\subsubsection{Setting}
\paragraph{Datasets.}The classification is conducted on three different types of datasets, including the image datasets YaleB \citep{Lee05}, CIFAR10 \citep{Krizhevsky09Cifar10} and ImageNet1000 \citep{ImageNet09}, the speech dataset TIMIT \citep{fisher1986darpa}, and the text dataset Newsgroup \citep{lang1995newsweeder}. The datasets are briefly introduced as follows. YaleB  contains  face images of 38 persons, with about 64 faces per person. CIFAR10 consists of 60,000 color images from 10 different classes, with each class having 6,000 images. ImageNet1000 consists of 1000 object classes, with 1,281,167 training images, 50,000 validation images, and 100,000 test images. For the above three image datasets, we separately extract their features using Discrete Wavelet Transform (DWT), ResNet18 \citep{he2016deep} and VGG16 \citep{simonyan2014very}. For ease of simulation, the resulting feature vectors are  dimensionally reduced by integer multiples, leading to the sizes of 1200, 5018, and 5018 respectively.  From TIMIT, as in \citep{mohamed2011deep, hutchinson2012deep}, we extract 39 classes of 429-dimensional phoneme features for classification, totally with 1,134,138 training samples and 58,399 test samples.  Newsgroup comprises 20 categories of texts, with 11,269 samples for training, and  7,505 samples for testing. The feature dimension is reduced to 5000 by selecting the top 5000 most frequent words in the bag of words, as done in \citep{larochelle2012learning}.

\paragraph{Classification.}  To validate the broad applicability of our feature discrimination analysis results, we here examine not only binary classification but also multiclass classification. Regarding the classifiers, we employ not only the fundamental  KNN and SVM which mainly involve linear operations, but also  more complicated classifiers involving nonlinear operations, such as the  multilayer perceptron (MLP) \citep{rumelhart1986learning} and decision trees \citep{quinlan1986induction}.   For brevity, in the main text we present the classification results of YaleB, Newsgroup, and TIMIT using KNN (with Euclidean distance) and SVM, as illustrated in Figures \ref{fig:yaleb-knn-svm} to \ref{fig:newsgroup-knn-svm}. The results for other datasets, such as CIFAR10 and ImageNet1000, and other classifiers, including KNN with cosine distance, MLP and decision trees, are provided in  Appendix \ref{sec-other_experiments},  specifically in Figures \ref{fig:cifar-knn-svm} to \ref{fig:ImageNet-KNN-MC}.

For each dataset, we iterate through randomly-selected class pairs to perform binary classification. The samples for training and testing are selected according to the default settings of the datasets. For YaleB without prior settings, we randomly assign half of the samples for training and the remaining half for testing. In the simulation, we need to test the classification performance of quantized data across varying quantization threshold $\tau$ values.
 For ease of analysis, we apply a uniform threshold $\tau$ value to all dimensions of feature vectors in each trial, though individual thresholds per dimension could potentially yield higher feature discrimination. To tackle the scale varying of the threshold $\tau$ values across different data, we here suppose $\tau=\gamma\cdot\eta$,  where  $\eta$ denotes  the average magnitude of the feature elements  of all feature vectors participating in classification, and  $\gamma$ is a scaling parameter. By adjusting the value of $\gamma$ within a narrow range, as illustrated later, we can derive the desired threshold $\tau$ values for various types of data.

%For conciseness, we assign a uniform threshold $\tau$ value across all dimensions of feature vectors in each trial, although higher feature discrimination could potentially be achieved by assigning a specific threshold value to each dimension.

%The value of $\tau$ should correlate with the element scale of the feature vectors, in the pursuit of improving classification over original data.

%\textcolor{blue}{To verify the generalizability of our feature discrimination analysis between two classes, we not only evaluate binary classification  using KNN and SVM, but also conduct multiclass classification, as well as nonlinear  classification using  multilayer perceptron (MLP) \citep{rumelhart1986learning} and decision trees \citep{quinlan1986induction}.   Due to space limitations, in the main body, we present the classification results of YaleB, Newsgroup, and TIMIT using KNN with Euclidean distance and SVM, as illustrated in Figures \ref{fig:yaleb-knn-svm} to \ref{fig:newsgroup-knn-svm}. The results for other datasets, such as CIFAR10 and ImageNet1000, and other classifiers, including KNN with cosine distance, MLP and decision trees, are provided in the appendix, but briefly discussed within the main text.}

\subsubsection{Results}

\begin{figure*}[t]
    \centering

		\begin{subfigure}[b]{0.24\textwidth}
			\includegraphics[width=\textwidth]{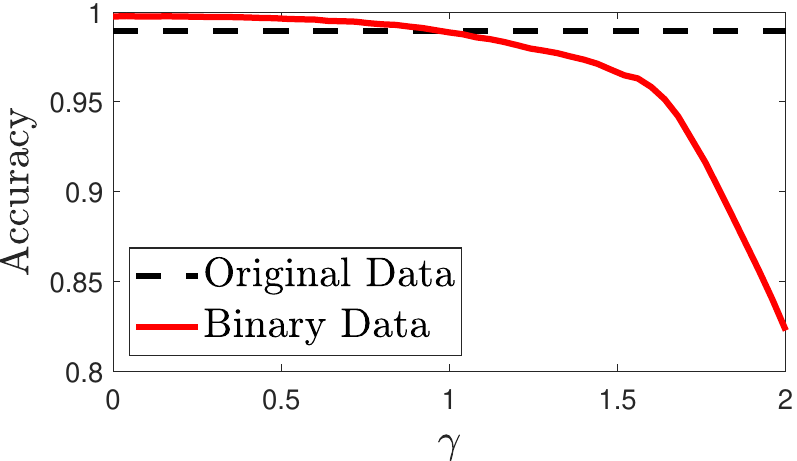}
			\caption{\centering \footnotesize Binary data (KNN)}
		\end{subfigure}
		\begin{subfigure}[b]{0.24\textwidth}
			\includegraphics[width=\textwidth]{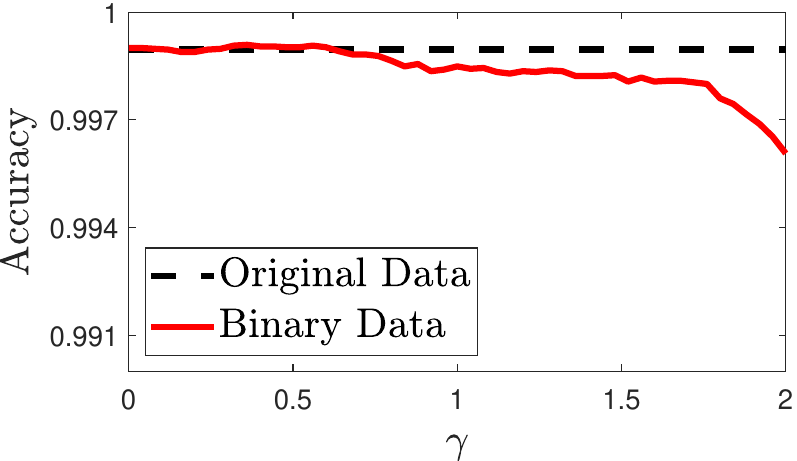}
			\caption{\centering \footnotesize Binary data (SVM)}
		\end{subfigure}
         \begin{subfigure}[b]{0.24\textwidth}
			\includegraphics[width=\textwidth]{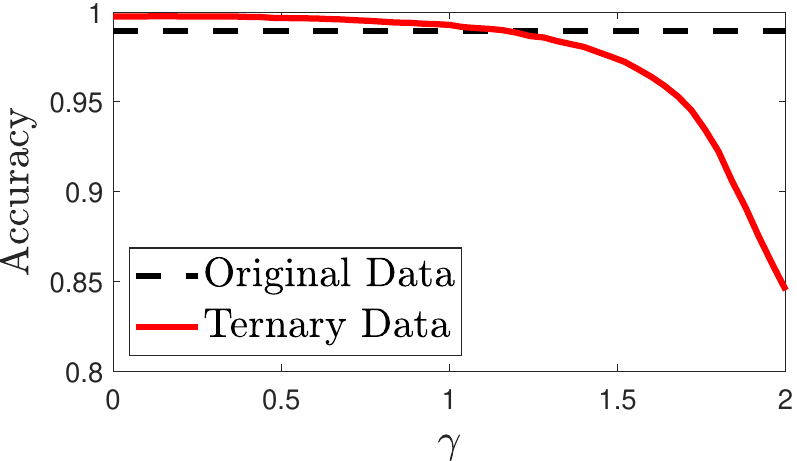}
			\caption{\centering \footnotesize Ternary data (KNN)}
		\end{subfigure}
		\begin{subfigure}[b]{0.24\textwidth}
			\includegraphics[width=\textwidth]{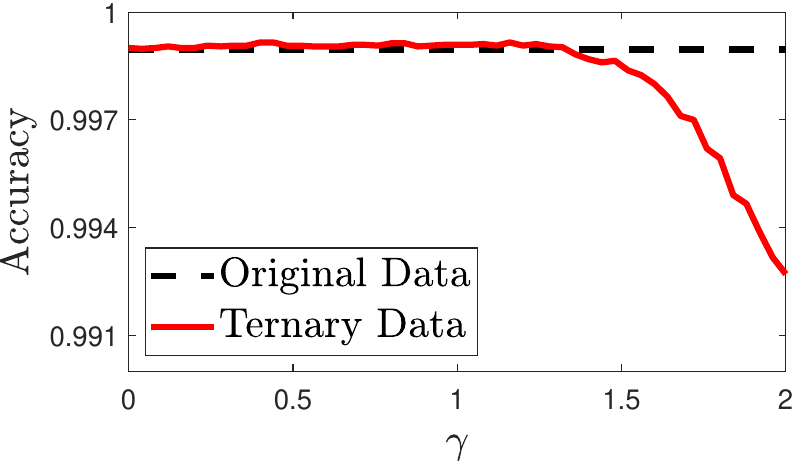}
			\caption{\centering \footnotesize Ternary data (SVM)}
		\end{subfigure}
	
	\caption{Classification accuracy for the binary, ternary, and original data by KNN (Euclidean distance) and SVM on YaleB. The parameter $\gamma$ corresponds to a threshold  $\tau=\gamma\cdot\eta$,  where  $\eta$ denotes  the average magnitude of the feature elements  in all  feature vectors.}
	\captionsetup{font=normalsize}
	\label{fig:yaleb-knn-svm}
\end{figure*}
\begin{figure*}[t!]
    \centering

		\begin{subfigure}[b]{0.24\textwidth}
			\includegraphics[width=\textwidth]{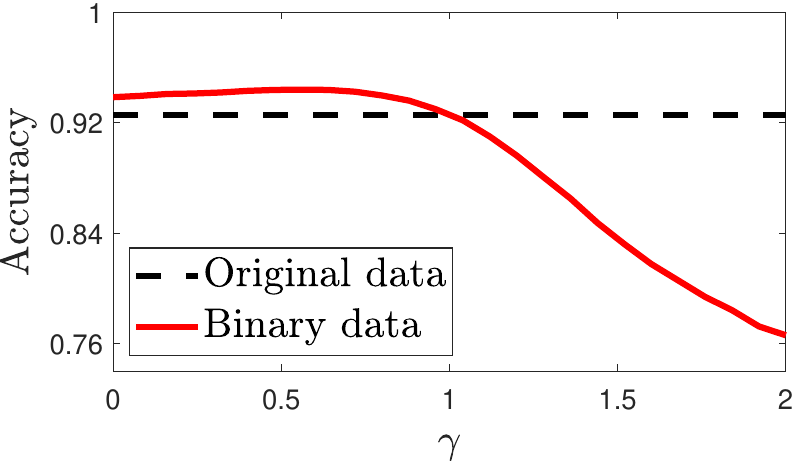}
			\caption{\centering \footnotesize Binary data (KNN) }
		\end{subfigure}
		\begin{subfigure}[b]{0.24\textwidth}
			\includegraphics[width=\textwidth]{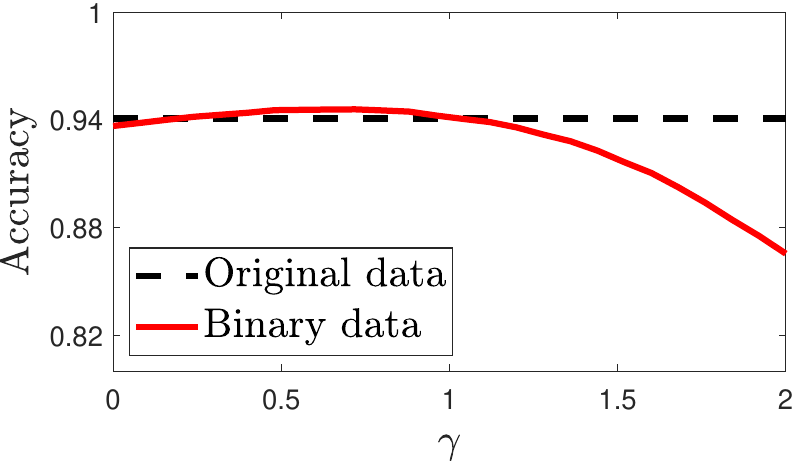}
			\caption{\centering \footnotesize Binary data (SVM)}
		\end{subfigure}
        \begin{subfigure}[b]{0.24\textwidth}
			\includegraphics[width=\textwidth]{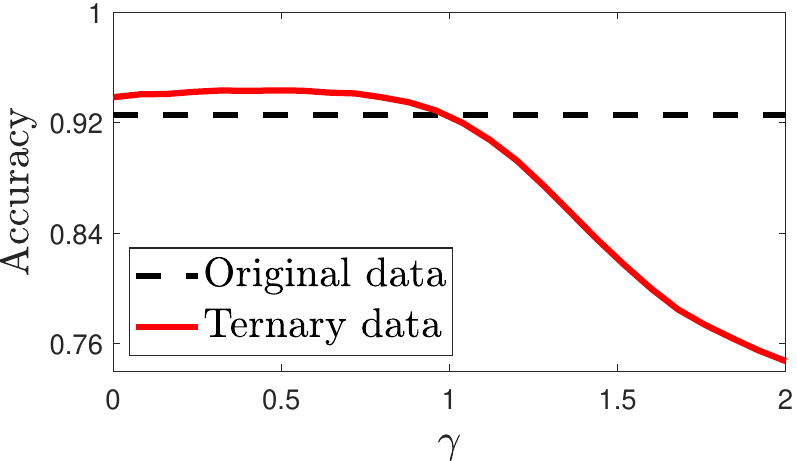}
			\caption{\centering \footnotesize Ternary data (KNN) }
		\end{subfigure}
		\begin{subfigure}[b]{0.24\textwidth}
			\includegraphics[width=\textwidth]{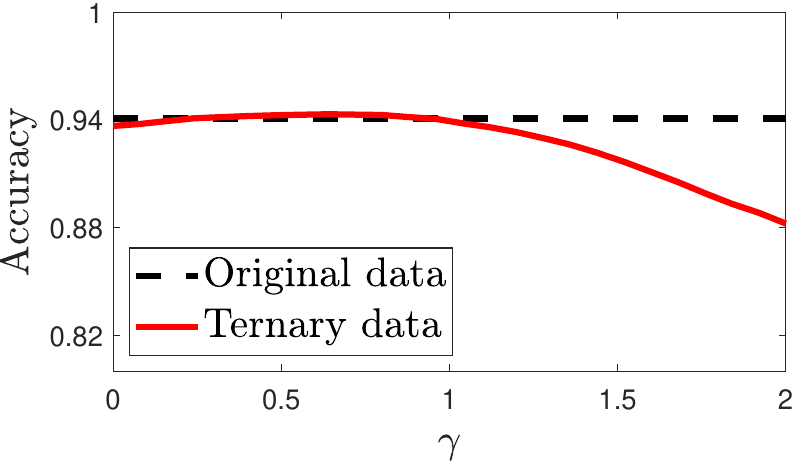}
			\caption{\centering \footnotesize Ternary data (SVM)}
		\end{subfigure}
	
	\caption{Classification accuracy for the binary, ternary, and original data by KNN (Euclidean distance) and SVM on TIMIT. The parameter $\gamma$ corresponds to a threshold  $\tau=\gamma\cdot\eta$,  where  $\eta$ denotes  the average magnitude of the feature elements  in all  feature vectors.}
	\captionsetup{font=normalsize}
	\label{fig:timit-knn-svm}
	
\end{figure*}
\begin{figure*}[t!]
    \centering
		\begin{subfigure}[b]{0.24\textwidth}
			\includegraphics[width=\textwidth]{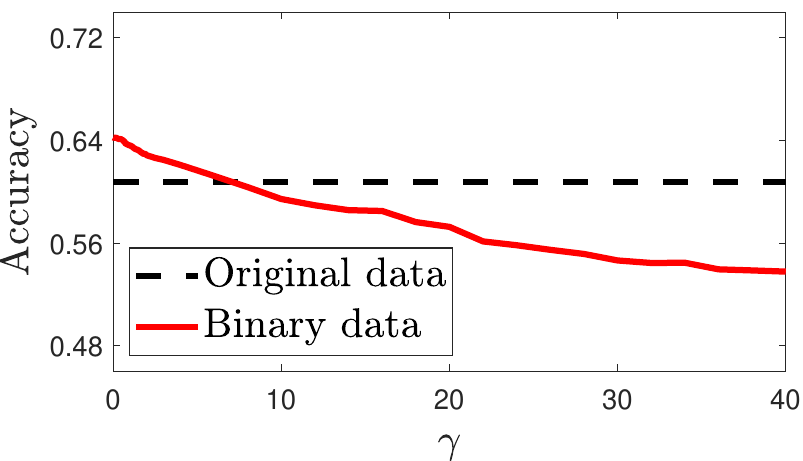}
			\caption{\footnotesize Binary data (KNN)}
		\end{subfigure}
		\begin{subfigure}[b]{0.24\textwidth}
			\includegraphics[width=\textwidth]{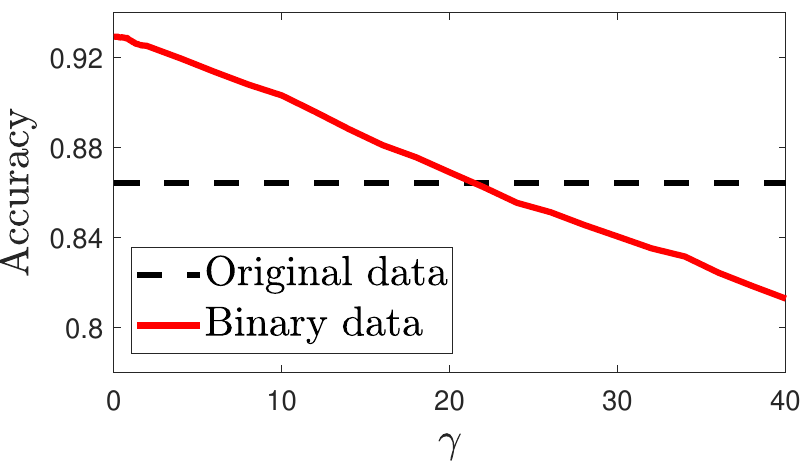}
			\caption{\footnotesize Binary data (SVM)}
		\end{subfigure}
		\begin{subfigure}[b]{0.24\textwidth}
			\includegraphics[width=\textwidth]{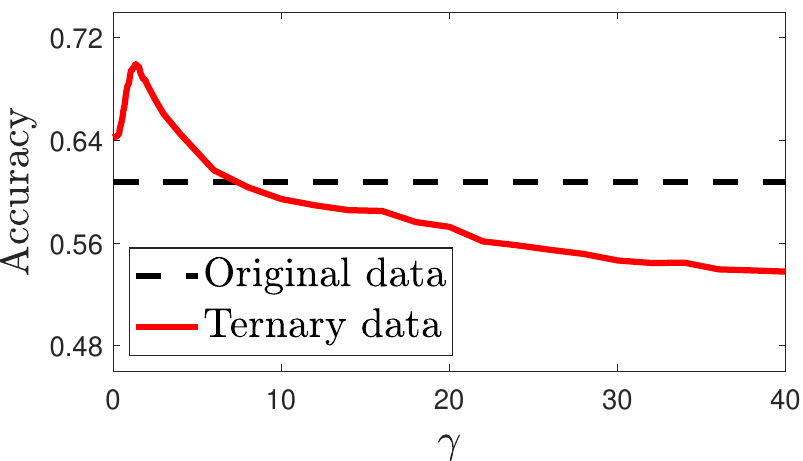}
			\caption{\centering \footnotesize Ternary data (KNN)}
		\end{subfigure}
		\begin{subfigure}[b]{0.24\textwidth}
			\includegraphics[width=\textwidth]{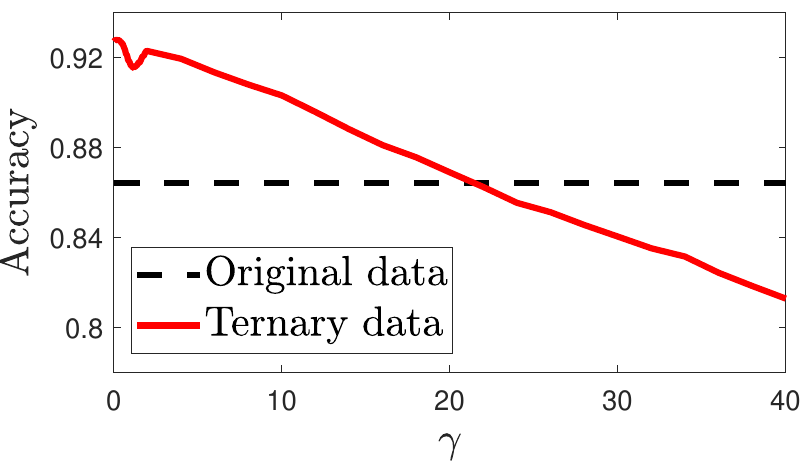}
			\caption{\centering \footnotesize Ternary data (SVM)}
		\end{subfigure}
	\caption{Classification accuracy for the binary, ternary, and original data by KNN (Euclidean distance) and SVM on Newsgroup. The parameter $\gamma$ corresponds to a threshold  $\tau=\gamma\cdot\eta$,  where  $\eta$ denotes  the average magnitude of the feature elements  in all  feature vectors.}
	\captionsetup{font=normalsize}
	\label{fig:newsgroup-knn-svm}
	
\end{figure*}

%\paragraph{Binary classification using KNN and SVM.}

\paragraph{Binary classification.} In Figures \ref{fig:yaleb-knn-svm}-\ref{fig:newsgroup-knn-svm},\ref{fig:cifar-knn-svm}, \ref{fig:Real data knn-Cos} and \ref{fig:ImageNet-KNN-BC}, we present the  binary classification results of KNN and SVM on five distinct datasets.  It can be seen that  there indeed exist  specific ranges of quantization threshold $\tau$ values that allow both binary and ternary quantization  to achieve superior, or  comparable classification performance compared to the original data. Note that as illustrated in Figure \ref{fig:real data-density}, each data category approximately follows, rather than perfectly  conforms to the  Gaussian distribution assumption made in our theoretical analysis. This underscores the robustness of our theoretical results. Similar to the classification of synthetic data, the classification of real data exhibits the following performance trends. 1) When using Euclidean distance, KNN  consistently obtains  quantization threshold $\tau$ values that raise the classification accuracy of original data across all datasets. 2) In contrast  to cosine distance, Euclidean distance enables KNN to cover a wider range of $\tau$ values that facilitate  classification improvement. 3) With SVM, quantization occasionally achieves comparable performance to the original data, rather than surpassing it, as exemplified in Figure \ref{fig:cifar-knn-svm}. 4) Ternary quantization often outperforms binary quantization by offering a broader range of threshold $\tau$ values conductive to classification improvement. The reasoning behind these results has been elaborated in our previous classification analysis of synthetic data. Overall, the consistency in classification performance between real data and synthetic data  substantiates the rationality and generalizability of our theoretical findings.

KNN and SVM are  well-suited for our linear discrimination analysis since they use linear operations to assess feature similarity. Other, more complex nonlinear classifiers, like MLP and decision trees,  can also obtain the  threshold $\tau$ values that enhance classification performance, as demonstrated in Figures \ref{fig:MLP-binary} and \ref{fig:Decision tree-binary}. This capability  stems from the fact that these nonlinear classifiers are grounded  in linear operations,  which serve to  measure the linear similarity between features or model parameters, thereby aligning with our analytical framework.

\paragraph{Multiclass classification.} By enhancing feature discrimination between two classes, the performance of multiclass classification is expected to improve as well. This improvement has been observed in recent experiments \citep{lu2023quantization}, particularly in the challenging  1,000-class classification on deep features of ImageNet1000, as illustrated in Figure \ref{fig:ImageNet-KNN-MC}. In multiclass classification, we simply apply a uniform threshold to each dimension of the feature vectors. This approach conforms to our analytical framework rooted in binary classification, as classification in each  dimension can be approximately considered a binary classification problem, where the values of feature elements across different classes indicate the intensity of the feature attribute   at that dimension, i.e., whether it is strong or weak. The successful adaption to multiclass classification highlights the broad applicability of our theoretical findings.

\section{Conclusion}

The paper proposes a novel methodology that leverages feature discrimination to evaluate the influence of quantization on classification accuracy. Unlike the conventional approach, which focuses primarily on quantization errors, this feature discrimination-based analysis  provides a more direct and logically sound assessment of classification performance.

The theoretical analysis reveals that common binary and ternary quantization techniques can enhance the feature discrimination of the original data, particularly when data vectors within the same class exhibit Gaussian distributions along each dimension. These theoretical insights are supported by numerical simulations. Moreover, the enhanced feature discrimination is validated through classification tasks conducted on both synthetic data adhering to Gaussian distributions and a variety of real-world datasets that do not strictly conform to Gaussian distributions, such as convolutional features extracted from images, spectral features derived from speech, and term frequency features obtained from texts. This underscores the robustness and generalizability of the theoretical findings.

A significant contribution of this research lies in its challenge to the widely accepted notion that larger quantization errors generally result in poorer classification performance. By challenging  this conventional view, the study opens avenues for the exploration of more advanced quantization techniques that could potentially enhance, rather than impair, model performance. Consequently, this research is expected to drive advancements in the fields of model quantization and compression, particularly within the realms of deep neural networks \citep{qin2020binary,gholami2022survey} and biological neural networks \citep{haufe2010modeling,dasgupta2017neural}.

\bibliography{icml2025_conference}

\begin{thebibliography}{41}
\providecommand{\natexlab}[1]{#1}
\providecommand{\url}[1]{\texttt{#1}}
\expandafter\ifx\csname urlstyle\endcsname\relax
  \providecommand{\doi}[1]{doi: #1}\else
  \providecommand{\doi}{doi: \begingroup \urlstyle{rm}\Url}\fi

\bibitem[Baras and Dey(1999)]{baras1999combined}
J.~S. Baras and S.~Dey.
\newblock Combined compression and classification with learning vector
  quantization.
\newblock \emph{IEEE Transactions on Information Theory}, 45\penalty0
  (6):\penalty0 1911--1920, 1999.

\bibitem[Bertsekas(1997)]{bertsekas1997nonlinear}
D.~P. Bertsekas.
\newblock Nonlinear programming.
\newblock \emph{Journal of the Operational Research Society}, 48\penalty0
  (3):\penalty0 334--334, 1997.

\bibitem[Charikar(2002)]{charikar2002similarity}
M.~S. Charikar.
\newblock Similarity estimation techniques from rounding algorithms.
\newblock In \emph{Proceedings of the thiry-fourth annual ACM symposium on
  Theory of computing}, pages 380--388, 2002.

\bibitem[Cortes and Vapnik(1995)]{cortes1995support}
C.~Cortes and V.~Vapnik.
\newblock Support-vector networks.
\newblock \emph{Machine learning}, 20\penalty0 (3):\penalty0 273--297, 1995.

\bibitem[Courbariaux et~al.(2015)Courbariaux, Bengio, and
  David]{courbariaux2015binaryconnect}
M.~Courbariaux, Y.~Bengio, and J.-P. David.
\newblock Binaryconnect: Training deep neural networks with binary weights
  during propagations.
\newblock \emph{Advances in neural information processing systems}, 28, 2015.

\bibitem[Dasgupta et~al.(2017)Dasgupta, Stevens, and
  Navlakha]{dasgupta2017neural}
S.~Dasgupta, C.~F. Stevens, and S.~Navlakha.
\newblock A neural algorithm for a fundamental computing problem.
\newblock \emph{Science}, 358\penalty0 (6364):\penalty0 793--796, 2017.

\bibitem[Deng et~al.(2009)Deng, Dong, Socher, Li, Li, and Fei-Fei]{ImageNet09}
J.~Deng, W.~Dong, R.~Socher, L.-J. Li, K.~Li, and L.~Fei-Fei.
\newblock {ImageNet: A Large-Scale Hierarchical Image Database}.
\newblock In \emph{IEEE Conference on Computer Vision and Pattern Recognition},
  2009.

\bibitem[Dogahe and Murthi(2011)]{dogahe2011quantization}
B.~M. Dogahe and M.~N. Murthi.
\newblock Quantization for classification accuracy in high-rate quantizers.
\newblock In \emph{Digital Signal Processing and Signal Processing Education
  Meeting}, pages 277--282. IEEE, 2011.

\bibitem[Fisher(1936)]{fisher1936use}
R.~A. Fisher.
\newblock The use of multiple measurements in taxonomic problems.
\newblock \emph{Annals of eugenics}, 7\penalty0 (2):\penalty0 179--188, 1936.

\bibitem[Fisher et~al.(1986)Fisher, Doddington, and
  Goudie-Marshall]{fisher1986darpa}
W.~M. Fisher, G.~R. Doddington, and K.~M. Goudie-Marshall.
\newblock The darpa speech recognition research database: Specifications and
  status.
\newblock In \emph{Proceedings of DARPA Workshop on Speech Recognition}, pages
  93--99, 1986.

\bibitem[Gholami et~al.(2022)Gholami, Kim, Dong, Yao, Mahoney, and
  Keutzer]{gholami2022survey}
A.~Gholami, S.~Kim, Z.~Dong, Z.~Yao, M.~W. Mahoney, and K.~Keutzer.
\newblock A survey of quantization methods for efficient neural network
  inference.
\newblock In \emph{Low-Power Computer Vision}, pages 291--326. Chapman and
  Hall/CRC, 2022.

\bibitem[Gray and Neuhoff(1998)]{gray1998quantization}
R.~M. Gray and D.~L. Neuhoff.
\newblock Quantization.
\newblock \emph{IEEE transactions on information theory}, 44\penalty0
  (6):\penalty0 2325--2383, 1998.

\bibitem[Haufe et~al.(2010)Haufe, Tomioka, Nolte, M{\"u}ller, and
  Kawanabe]{haufe2010modeling}
S.~Haufe, R.~Tomioka, G.~Nolte, K.-R. M{\"u}ller, and M.~Kawanabe.
\newblock Modeling sparse connectivity between underlying brain sources for
  {EEG/MEG}.
\newblock \emph{IEEE transactions on biomedical engineering}, 57\penalty0
  (8):\penalty0 1954--1963, 2010.

\bibitem[He et~al.(2016)He, Zhang, Ren, and Sun]{he2016deep}
K.~He, X.~Zhang, S.~Ren, and J.~Sun.
\newblock Deep residual learning for image recognition.
\newblock In \emph{Proceedings of the IEEE conference on computer vision and
  pattern recognition}, pages 770--778, 2016.

\bibitem[Hoefler et~al.(2021)Hoefler, Alistarh, Ben-Nun, Dryden, and
  Peste]{hoefler2021sparsity}
T.~Hoefler, D.~Alistarh, T.~Ben-Nun, N.~Dryden, and A.~Peste.
\newblock Sparsity in deep learning: Pruning and growth for efficient inference
  and training in neural networks.
\newblock \emph{Journal of Machine Learning Research}, 22\penalty0
  (241):\penalty0 1--124, 2021.

\bibitem[Hutchinson et~al.(2012)Hutchinson, Deng, and Yu]{hutchinson2012deep}
B.~Hutchinson, L.~Deng, and D.~Yu.
\newblock A deep architecture with bilinear modeling of hidden representations:
  Applications to phonetic recognition.
\newblock In \emph{IEEE international conference on acoustics, speech and
  signal processing}, pages 4805--4808. IEEE, 2012.

\bibitem[Jana and Moulin(2000)]{jana2000optimal}
S.~Jana and P.~Moulin.
\newblock Optimal design of transform coders and quantizers for image
  classification.
\newblock In \emph{International Conference on Image Processing}, volume~3,
  pages 841--844. IEEE, 2000.

\bibitem[Jana and Moulin(2003)]{jana2003optimal}
S.~Jana and P.~Moulin.
\newblock Optimal transform coding of gaussian mixtures for joint
  classification/reconstruction.
\newblock In \emph{Data Compression Conference}, pages 313--322. IEEE, 2003.

\bibitem[Kotz et~al.(2012)Kotz, Kozubowski, and Podgorski]{kotz2012laplace}
S.~Kotz, T.~Kozubowski, and K.~Podgorski.
\newblock \emph{The Laplace distribution and generalizations: a revisit with
  applications to communications, economics, engineering, and finance}.
\newblock Springer Science \& Business Media, 2012.

\bibitem[Krizhevsky and Hinton(2009)]{Krizhevsky09Cifar10}
A.~Krizhevsky and G.~Hinton.
\newblock Learning multiple layers of features from tiny images.
\newblock \emph{Master's thesis, Department of Computer Science, University of
  Toronto}, 2009.

\bibitem[Lam and Goodman(2000)]{lam2000mathematical}
E.~Y. Lam and J.~W. Goodman.
\newblock A mathematical analysis of the dct coefficient distributions for
  images.
\newblock \emph{IEEE transactions on image processing}, 9\penalty0
  (10):\penalty0 1661--1666, 2000.

\bibitem[Lang(1995)]{lang1995newsweeder}
K.~Lang.
\newblock Newsweeder: Learning to filter netnews.
\newblock In \emph{Machine learning proceedings 1995}, pages 331--339.
  Elsevier, 1995.

\bibitem[Larochelle et~al.(2012)Larochelle, Mandel, Pascanu, and
  Bengio]{larochelle2012learning}
H.~Larochelle, M.~Mandel, R.~Pascanu, and Y.~Bengio.
\newblock Learning algorithms for the classification restricted boltzmann
  machine.
\newblock \emph{The Journal of Machine Learning Research}, 13\penalty0
  (1):\penalty0 643--669, 2012.

\bibitem[Lee et~al.(2005)Lee, Ho, and Kriegman]{Lee05}
K.~Lee, J.~Ho, and D.~Kriegman.
\newblock Acquiring linear subspaces for face recognition under variable
  lighting.
\newblock \emph{IEEE Transactions on Pattern Analysis and Machine
  Intelligence}, 27\penalty0 (5):\penalty0 684--698, 2005.

\bibitem[Li et~al.(2016)Li, Zhang, and Liu]{li2016ternary}
F.~Li, B.~Zhang, and B.~Liu.
\newblock Ternary weight networks.
\newblock \emph{arXiv}, 1605.04711, 2016.

\bibitem[Lin et~al.(2016{\natexlab{a}})Lin, Talathi, and
  Annapureddy]{lin2016fixed}
D.~Lin, S.~Talathi, and S.~Annapureddy.
\newblock Fixed point quantization of deep convolutional networks.
\newblock In \emph{International conference on machine learning}, pages
  2849--2858. PMLR, 2016{\natexlab{a}}.

\bibitem[Lin et~al.(2016{\natexlab{b}})Lin, Courbariaux, Memisevic, and
  Bengio]{Zhouhan2016Neural}
Z.~Lin, M.~Courbariaux, R.~Memisevic, and Y.~Bengio.
\newblock Neural networks with few multiplications.
\newblock In \emph{International Conference on Learning Representations},
  2016{\natexlab{b}}.

\bibitem[Lu et~al.(2023)Lu, Chen, Guo, and Li]{lu2023quantization}
W.~Lu, M.~Chen, K.~Guo, and W.~Li.
\newblock Quantization: Is it possible to improve classification?
\newblock In \emph{Data Compression Conference}, pages 318--327. IEEE, 2023.

\bibitem[Mohamed et~al.(2011)Mohamed, Sainath, Dahl, Ramabhadran, Hinton, and
  Picheny]{mohamed2011deep}
A.-r. Mohamed, T.~N. Sainath, G.~Dahl, B.~Ramabhadran, G.~E. Hinton, and M.~A.
  Picheny.
\newblock Deep belief networks using discriminative features for phone
  recognition.
\newblock In \emph{IEEE international conference on acoustics, speech and
  signal processing}, pages 5060--5063. IEEE, 2011.

\bibitem[Oehler and Gray(1995)]{oehler1995combining}
K.~L. Oehler and R.~M. Gray.
\newblock Combining image compression and classification using vector
  quantization.
\newblock \emph{IEEE transactions on pattern analysis and machine
  intelligence}, 17\penalty0 (5):\penalty0 461--473, 1995.

\bibitem[Peterson(2009)]{peterson2009k}
L.~E. Peterson.
\newblock K-nearest neighbor.
\newblock \emph{Scholarpedia}, 4\penalty0 (2):\penalty0 1883, 2009.

\bibitem[Poor and Thomas(1977)]{poor1977applications}
H.~Poor and J.~Thomas.
\newblock Applications of ali-silvey distance measures in the design
  generalized quantizers for binary decision systems.
\newblock \emph{IEEE Transactions on Communications}, 25\penalty0 (9):\penalty0
  893--900, 1977.

\bibitem[Qin et~al.(2020)Qin, Gong, Liu, Bai, Song, and Sebe]{qin2020binary}
H.~Qin, R.~Gong, X.~Liu, X.~Bai, J.~Song, and N.~Sebe.
\newblock Binary neural networks: A survey.
\newblock \emph{Pattern Recognition}, 105:\penalty0 107281, 2020.

\bibitem[Quinlan(1986)]{quinlan1986induction}
J.~R. Quinlan.
\newblock Induction of decision trees.
\newblock \emph{Machine learning}, 1:\penalty0 81--106, 1986.

\bibitem[Rumelhart et~al.(1986)Rumelhart, Hinton, and
  Williams]{rumelhart1986learning}
D.~E. Rumelhart, G.~E. Hinton, and R.~J. Williams.
\newblock Learning representations by back-propagating errors.
\newblock \emph{nature}, 323\penalty0 (6088):\penalty0 533--536, 1986.

\bibitem[Simonyan and Zisserman(2014)]{simonyan2014very}
K.~Simonyan and A.~Zisserman.
\newblock Very deep convolutional networks for large-scale image recognition.
\newblock \emph{arXiv preprint arXiv:1409.1556}, 2014.

\bibitem[Srinivasamurthy and Ortega(2002)]{srinivasamurthy2002reduced}
N.~Srinivasamurthy and A.~Ortega.
\newblock Reduced complexity quantization under classification constraints.
\newblock In \emph{Data Compression Conference}, pages 402--411. IEEE, 2002.

\bibitem[Torralba and Oliva(2003)]{torralba2003statistics}
A.~Torralba and A.~Oliva.
\newblock Statistics of natural image categories.
\newblock \emph{Network: computation in neural systems}, 14\penalty0
  (3):\penalty0 391--412, 2003.

\bibitem[Wainwright and Simoncelli(1999)]{Wainwright99Gaussian}
M.~J. Wainwright and E.~P. Simoncelli.
\newblock Scale mixtures of gaussians and the statistics of natural images.
\newblock In \emph{Proceedings of the 12th International Conference on Neural
  Information Processing Systems}, 1999.

\bibitem[Wang et~al.(2023)Wang, Ma, Dong, Huang, Wang, Ma, Yang, Wang, Wu, and
  Wei]{wang2023bitnet}
H.~Wang, S.~Ma, L.~Dong, S.~Huang, H.~Wang, L.~Ma, F.~Yang, R.~Wang, Y.~Wu, and
  F.~Wei.
\newblock Bitnet: Scaling 1-bit transformers for large language models.
\newblock \emph{arXiv preprint arXiv:2310.11453}, 2023.

\bibitem[Weiss and Freeman(2007)]{weiss2007makes}
Y.~Weiss and W.~T. Freeman.
\newblock What makes a good model of natural images?
\newblock In \emph{IEEE Conference on Computer Vision and Pattern Recognition},
  pages 1--8. IEEE, 2007.

\end{thebibliography}

\newpage
\appendix
\onecolumn
\section{Proofs for Section \ref{sec-DiscriminationAnalysis}}
\label{sec-proof}
\subsection{Proof of Theorem \ref{prop:bq}}
Let $X_1$ and $X_2$ be i.i.d. samples of $X$, and $Y_1$ and $Y_2$ be i.i.d. samples of $Y$.  Denote $X_{i,b}$ and $Y_{i,b}$ as the binary quantization of $X_i$ and $Y_i$, i.e.
 $X_{i,b}=f_b(X_i;\tau)$ and $Y_{i,b}=f_b(Y_i;\tau)$, where $i=1,2$. By the distributions of $X$ and $Y$ specified in Property \ref{data-distribution} and the binary quantization function $f_b(x;\tau)$ defined in Equation\eqref{b_q}, the probability mass functions of $X_{i,b}$ and $Y_{i,b}$ can be derived as
\begin{equation}\label{b-3}
	P(X_{i,b}=k)
	=\left\{
	\begin{aligned}
		1&-\alpha, & \quad k=1 \\
		&\alpha, & \quad k=0
	\end{aligned}
	\right.
\end{equation}
and
\begin{equation}\label{b-4}
	P(Y_{i,b}=k)
	=\left\{
	\begin{aligned}
		1&-\beta, & \quad k=1 \\
		&\beta, & \quad k=0
	\end{aligned}
	\right.
\end{equation}
where $\alpha=\Phi\left(\frac{\tau-\mu}{\sigma}\right)$ and $\beta=\Phi\left(\frac{\tau+\mu}{\sigma}\right)$. By the  probability functions, it is easy to deduce that
\begin{align*}
	&E\left[(X_1-X_2)^2\right]=2\sigma^2,\qquad\quad~~	E\left[(X_{1,b}-X_{2,b})^2\right]= 2\alpha - 2\alpha^2,\\
	&E\left[(Y_1-Y_2)^2\right]=2\sigma^2,\qquad\qquad	E\left[(Y_{1,b}-Y_{2,b})^2\right]= 2\beta - 2\beta^2,\\
&E\left[(X_1-Y_2)^2\right]=2\sigma^2+4\mu^2,\quad   E\left[(X_{1,b}-Y_{1,b})^2\right]= \alpha + \beta - 2\alpha\beta.
\end{align*}
With these equations,  the discrimination $D$ of  original data, as specified in Definition \ref{def-discrimination}, can be further expressed as
\begin{equation}\label{b-p-2}
D=\frac{E[(X_1 - Y_1)^2]}{E[(X_1 - X_2)^2] + E[(Y_1 - Y_2)^2]} = \frac{\sigma^2 + 2\mu^2}{2\sigma^2},
\end{equation}
and similarly,  the discrimination $D_b$ of binary quantized data, as specified in Definition \ref{def-discrimination-Q}, can be written as
\begin{equation}\label{b-p-3}
		D_b= \frac{E[(X_{1,b} - Y_{1,b})^2]}{E[(X_{1,b} - X_{2,b})^2] + E[(Y_{1,b} - Y_{2,b})^2]} = \frac{\alpha - 2\alpha\beta + \beta}{(2\alpha - 2\alpha^2) + (2\beta - 2\beta^2)}.
\end{equation}
Next, we are ready to prove that $D_b>D$ under the condition \eqref{eq-bq}. By Equations \eqref{b-p-2} and \eqref{b-p-3},  it is easy to see that $D_b>D$ is equivalent to
\begin{equation}\label{b-p-1}
(\sigma^2 + 2\mu^2)\alpha^2 - 2(\sigma^2\beta + \mu^2)\alpha + (\sigma^2 + 2\mu^2)\beta^2 - 2\mu^2\beta > 0.
\end{equation}
This inequality can be viewed as a quadratic inequality  in $\alpha$, which has the discriminant:
$$
\Delta = 4\mu^4 + 16(1-\beta)\mu^2\beta> 0.
$$
By the above inequality, the inequality \eqref{b-p-1} holds when $\alpha\in(-\infty, \alpha_1)\cup(\alpha_2,+\infty)$, where
$$
\alpha_1= \beta + \frac{\mu^2(1-2\beta) - \mu\sqrt{\mu^2 + 4\beta(1-\beta)}}{1 + \mu^2},
$$
and
\begin{equation}\label{b-p-4}
\alpha_2= \beta + \frac{\mu^2(1-2\beta) + \mu\sqrt{\mu^2 + 4\beta(1-\beta)}}{1 + \mu^2}.
\end{equation}
Given \eqref{b-p-4}, we can further derive $\alpha_2>\beta$,  since $\mu^2(1-2\beta) + \mu\sqrt{\mu^2 + 4\beta(1-\beta)}>0$. However, this result contradicts the conclusion that $\alpha<\beta$ we can derive  with the probability mass functions shown in \eqref{b-3} and \eqref{b-4}, mainly by the increasing property of  $\Phi(\cdot)$.
So the solution to the inequality \eqref{b-p-1} should be $\alpha\in(-\infty, \alpha_1)$, implying $\beta-\alpha + \frac{\mu^2(1-2\beta) - \mu\sqrt{\mu^2 + 4\beta(1-\beta)}}{1 + \mu^2}>0$.

\subsection{Proof of Theorem \ref{prop:tq}}
Let $X_1$ and $X_2$ be i.i.d. samples of $X$, and $Y_1$ and $Y_2$ be i.i.d. samples of $Y$.  Denote $X_{i,t}=f_t(X_i;\tau)$ and $Y_{i,t}=f_t(Y_i;\tau)$, where $i=1,2$. By the distributions of $X$ and $Y$ specified in Property \ref{data-distribution} and the ternary quantization $f_t(x;\tau)$ defined in Equation \eqref{t_q}, the probability mass functions of $X_{i,t}$ and $Y_{i,t}$ can be derived as
\begin{equation}\label{t-3}
	P(X_{i,t}=k)
	=\left\{
	\begin{aligned}
		&\beta, &k&=1\\
		1 - &\alpha - \beta,  &k&=0\\
		&\alpha, &k&=-1
	\end{aligned}
	\right.
\end{equation}

\begin{equation}\label{t-4}
	P(Y_{i,t}=k)
	=\left\{
	\begin{aligned}
		&\alpha, &k&=1\\
		1 - &\alpha - \beta, &k&=0\\
		&\beta, &k&=-1
	\end{aligned}
	\right.
\end{equation}
where $\alpha = \Phi(\frac{-\tau-\mu}{\sigma})$ and $\beta = \Phi(\frac{-\tau+\mu}{\sigma})$.

Then, by Definition \ref{def-discrimination-Q}, the discrimination $D_t$ of ternary quantization can be derived as
\begin{equation}\label{t-13}
		D_t = \frac{E[(X_{1,t} - Y_{1,t})^2]}{E[(X_{1,t} - X_{2,t})^2] + E[(Y_{1,t} - Y_{2,t})^2]}  = \frac{(\alpha + \alpha^2 - 2a\beta + \beta + \beta^2)}{2(\alpha - \alpha^2 + 2\alpha\beta + \beta - \beta^2)}.
\end{equation}
By Equations \eqref{b-p-2} and \eqref{t-13},  it can be seen that  $D_t> D$ is equivalent to
$$
\frac{(\alpha + \beta) + (\alpha - \beta)^2}{2(\alpha + \beta) - 2(\alpha - \beta)^2} > \frac{\sigma^2 + 2\mu^2}{2\sigma^2},
$$
which can simplify to
\begin{equation}\label{t-p-1}
	\alpha^2 - (2\beta + \mu^2)\alpha + \beta^2 - \mu^2\beta > 0.
\end{equation}
Clearly, \eqref{t-p-1} can be regarded as a quadratic inequality in $\alpha$, with its discriminant:
$$
\Delta= \mu^4 + 8\mu^2\beta> 0.
$$
This inequality implies that  the inequality \eqref{t-p-1} holds when $\alpha\in (-\infty,\alpha_1)\cup (\alpha_2,+\infty)$, where
$$
\alpha_1=\beta + \frac{\mu^2 - \sqrt{\mu^4 + 8\mu^2\beta}}{2}$$
and
\begin{equation}\label{t-p-2}
\alpha_2=\beta + \frac{\mu^2 + \sqrt{\mu^4 + 8\mu^2\beta}}{2}.
\end{equation}
In \eqref{t-p-2}, the term $\mu^2 + \sqrt{\mu^4 + 8\mu^2\beta}>0$, implying $\alpha_2>\beta$. In contrast, we will derive $\alpha<\beta$
by the probability functions shown in Equations \eqref{t-3} and \eqref{t-4}, particularly by the increasing property of $\Phi(\cdot)$. By this contradiction, we can say that $D_t>D$ holds only under the case of $\alpha\in (-\infty,\alpha_1)$, namely
$$
\beta-\alpha+\frac{\mu^2 - \sqrt{\mu^4 + 8\mu^2\beta}}{2}>0.
$$

\newpage
\section{Solution algorithms for Equations \eqref{eq-bq} and \eqref{eq-tq}}
\label{sec-algorithm}
In this section, we present two algorithms for deriving the quantization thresholds $\beta$ that can enhance feature discrimination, provided they exist as specified in Equations \eqref{eq-bq} and \eqref{eq-tq} (in Theorems \ref{prop:bq} and \ref{prop:tq}) for binary and ternary quantization, respectively.

\subsection{Solution algorithm for Equation \eqref{eq-bq}}
To derive a threshold $\beta$ that satisfies the inequality \eqref{eq-bq}, we can  minimize the objective function:
$$g(\tau) = -\beta + \alpha - \frac{\mu^2 (1 - 2\beta) - \mu\sqrt{\mu^2 + 4\beta(1 - \beta)}}{1 + \mu^2}$$
by iteratively descending the gradient:
$$g'(\tau) = -\frac{1 - \mu^2}{1 + \mu^2}\beta' + \alpha' + \frac{\mu(2\beta' - 4\beta\beta')}{(1 + \mu^2)\sqrt{\mu^2 + 4\beta(1 - \beta)}}$$
where $\alpha = \varphi\left(\frac{\tau - \mu}{\sigma}\right)$, $\beta = \varphi\left(\frac{\tau + \mu}{\sigma}\right)$, and $\varphi(\cdot)$ is the probability density function of the standard normal distribution. The gradient descent  step size $\gamma$ in the $k$-th iteration can be selected with the Armijo rule \citep{bertsekas1997nonlinear}:
$$g(\tau^{(k)} - \gamma \cdot g'(\tau^{(k)}) \leq g(\tau^{(k)}) - c\cdot \gamma \cdot (g'(\tau^{(k)}))^2$$
where $c \in (0, 1)$ is a constant.

\subsection{Solution algorithm for Equation \eqref{eq-tq}}
In a similar way, we can derive a threshold $\beta$  that holds  the inequality \eqref{eq-tq}  by minimizing

$$g(\tau) =  -\beta + \alpha - \frac{\mu^2 - \sqrt{\mu^4 + 8\mu^2 \beta}}{2}$$
via iteratively descending the gradient:

$$g'(\tau) = -\beta' + \alpha' + \frac{2\mu\beta'}{\sqrt{\mu^2 + 8\beta}}$$
where $\alpha = \varphi\left(\frac{-\tau - \mu}{\sigma}\right)$, $\beta = \varphi\left(\frac{-\tau + \mu}{\sigma}\right)$, and $\varphi(\cdot)$ is the probability density function of the standard normal distribution.   The gradient descent  step size can be determined with the Armijo rule described above.

\subsection{Experimental verification}
%To validate the proposed solution algorithms, we evaluate the KNN (with  Euclidean distance and $K=5$) classification accuracy both for the binary and ternary  quantization data on the synthetic  and real data, as specified in \ref{sec:synthetic data} and \ref{sec:real data}. For the solution algorithms, we set  the parameter $c = 10^{-3}$) for the Armijo rule, and terminate the iteration  when $| g'(\tau^{(k)}) |_2 < 10^{-12}$ or the maximum iteration number is reached. For comparison, we also examine the performance of the conventional quantization method which  determines quantization thresholds by minimizing $\ell_2$-norm quantization errors (MQE) \citep{li2016ternary}.

Our solution algorithms are expected to produce quantization thresholds that are conductive to enhancing  feature discrimination. To verify this, we evaluate the KNN classification accuracy  for both binary and ternary  quantization data derived from synthetic and real datasets, as described in Sections \ref{sec:synthetic data} and \ref{sec:real data}. In our algorithms, we set the parameter $c = 10^{-3}$ for the Armijo rule and halt the iteration when $| g'(\tau^{(k)}) |_2 < 10^{-12}$ or the maximum iteration number is reached. For comparison, we also assess the performance of the conventional quantization method, which determines quantization thresholds by minimizing $\ell_2$-norm quantization errors (MQE) \citep{li2016ternary}. The experimental results are presented in Tables \ref{tab:solution-binary} and \ref{tab:solution-ternary}. As shown, our quantization data achieve superior classification accuracy than the conventional MQE  data and the original data.

\begin{table}[htbp]
\centering
\caption{Classification accuracy (\%) of  the original data, as well as the binary quantization data derived with the conventional MQE method and our  method. The best results are highlighted in bold. \label{tab:solution-binary}}
\begin{tabular}{c|cccc}
\hline
Dataset & Original Data &  Binary data (MQE) & Binary data (Ours) \\
\hline
YaleB & 98.93 & 99.52 & \textbf{99.74} \\
CIFAR10 & 94.08 & 92.58 & \textbf{93.54} \\
TIMIT & 92.58 & 94.35 & \textbf{94.46} \\
Newsgroup & 60.78 & 62.41 & \textbf{62.50} \\
ImageNet & 91.88 & 91.79 & \textbf{93.22} \\
Synthetic Data & 89.78 & 81.89 & \textbf{90.98} \\
\hline
\end{tabular}
\label{tab:binary_results}
\end{table}

\begin{table*}[htbp]
\centering
\caption{Classification accuracy (\%) of the original data, as well as the ternary quantization data derived with the conventional MQE  method and our method. The best results are highlighted in bold. \label{tab:solution-ternary}}
\begin{tabular}{c|cccc}
\hline
Dataset & Original data & Ternary data (MQE) & Ternary data (Ours) \\
\hline
YaleB & 98.93 & 99.64 & \textbf{99.70} \\
CIFAR10 & 94.08 & 96.56 & \textbf{96.91} \\
TIMIT & 92.58 & 94.10 & \textbf{94.15} \\
Newsgroup & 60.78 & 65.47 & \textbf{67.10} \\
ImageNet & 91.88 & 94.66 & \textbf{95.00} \\
Synthetic Data & 89.78 & 84.05 & \textbf{89.81} \\
\hline
\end{tabular}
\label{tab:ternary_results}
\end{table*}

\newpage
\section{Other experimental results}
\label{sec-other_experiments}
\subsection{Numerical validation}
\begin{figure}[H]
	\centering
	\begin{minipage}{\textwidth}
		\centering
		\begin{subfigure}[b]{0.45\textwidth}
			\includegraphics[width=\textwidth]{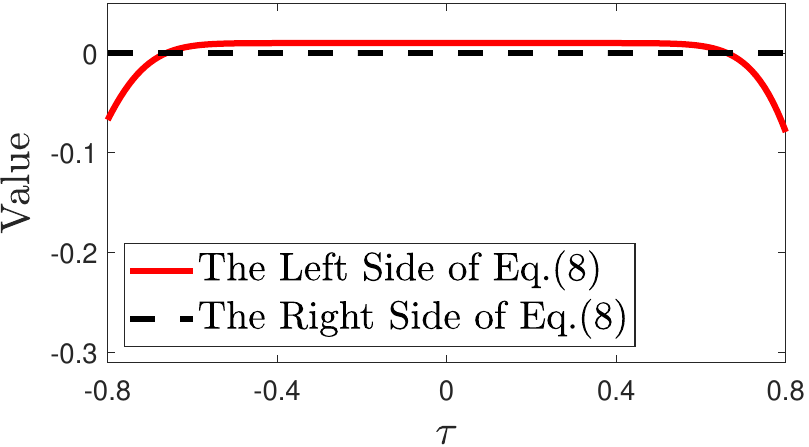}
			\caption{\footnotesize Theoretical results for the data with distribution parameters $\mu=0.99$ and $\sigma^2=0.02$}
		\end{subfigure}
		\begin{subfigure}[b]{0.45\textwidth}
			\includegraphics[width=\textwidth]{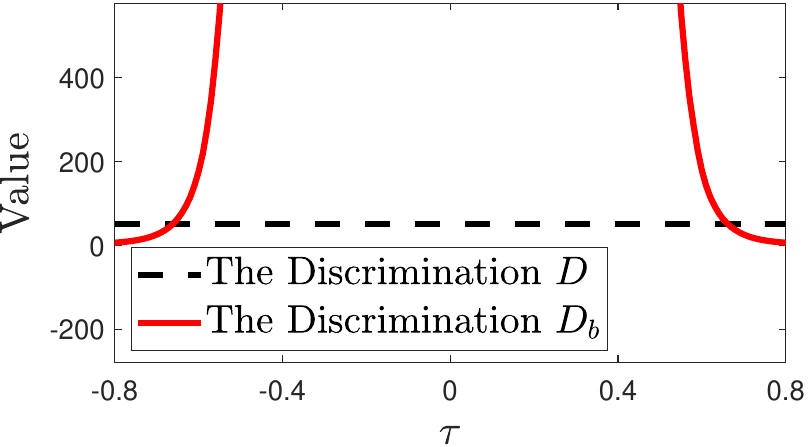}
			\caption{\footnotesize Numerical results for the data with distribution parameters $\mu=0.99$ and $\sigma^2=0.02$}
		\end{subfigure}
	\end{minipage}
	\begin{minipage}{\textwidth}
		\centering
		\begin{subfigure}[b]{0.45\textwidth}
			\includegraphics[width=\textwidth]{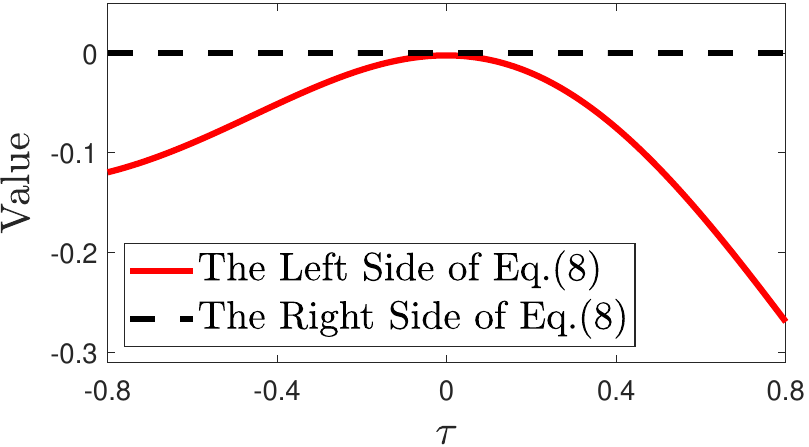}
			\caption{\footnotesize Theoretical results for the data with distribution parameters $\mu=0.76$ and $\sigma^2=0.42$}
		\end{subfigure}
		\begin{subfigure}[b]{0.45\textwidth}
			\includegraphics[width=\textwidth]{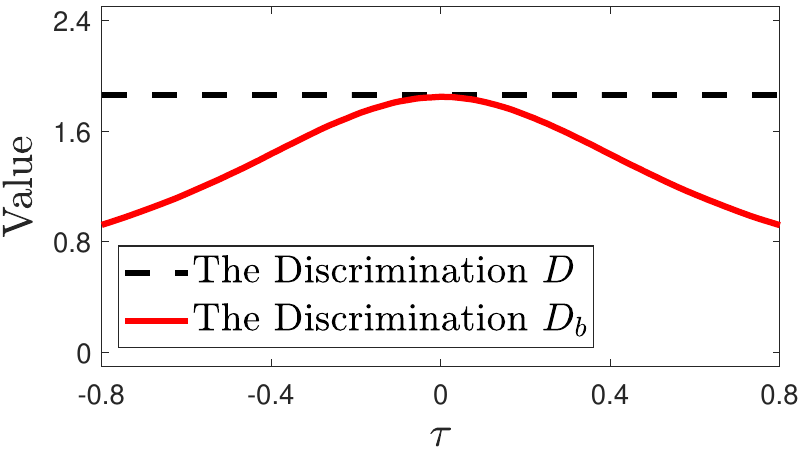}
			\caption{\footnotesize Numerical results for the data with distribution parameters $\mu=0.76$ and $\sigma^2=0.42$}
		\end{subfigure}
	\end{minipage}
	\captionsetup{font=normalsize}
	\caption{Consider the binary quantization on two classes of data $X\sim N(\mu, \sigma^2)$ and $Y\sim N(-\mu, \sigma^2)$  as specified in Property \ref{data-distribution}. For  two kinds of data with distribution parameters ($\mu=0.99$, $\sigma^2=0.02$) and ($\mu=0.76$, $\sigma^2=0.42$), the values for the left and right side of Equations \eqref{eq-bq}  are provided in (a) and (c)  respectively; and their discrimination $D$ and $D_b$ statistically estimated with Equations \eqref{d-1} and \eqref{d-B}  are illustrated in (b) and (d), respectively.}
	\label{fig:Numerical simu-2}
\end{figure}

\newpage
\begin{figure}[H]
	\centering
	\begin{minipage}{\textwidth}
		\centering
		\begin{subfigure}[b]{0.45\textwidth}
			\includegraphics[width=\textwidth]{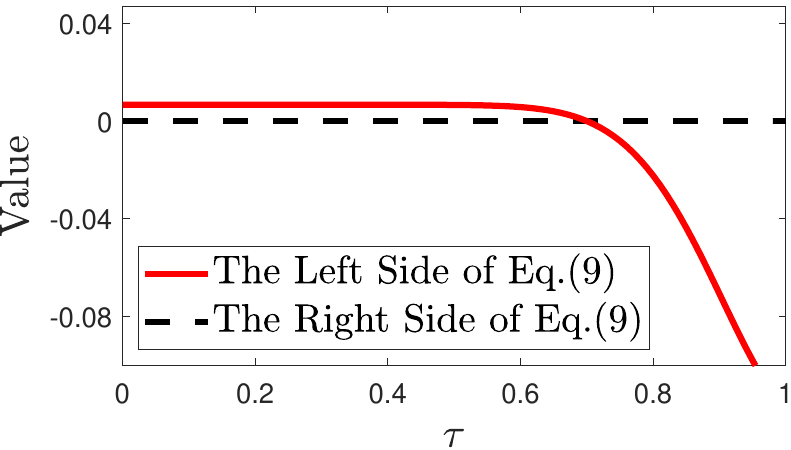}
			\caption{\footnotesize Theoretical results for the data with distribution parameters $\mu=0.99$ and $\sigma^2=0.02$}
		\end{subfigure}
		\begin{subfigure}[b]{0.45\textwidth}
			\includegraphics[width=\textwidth]{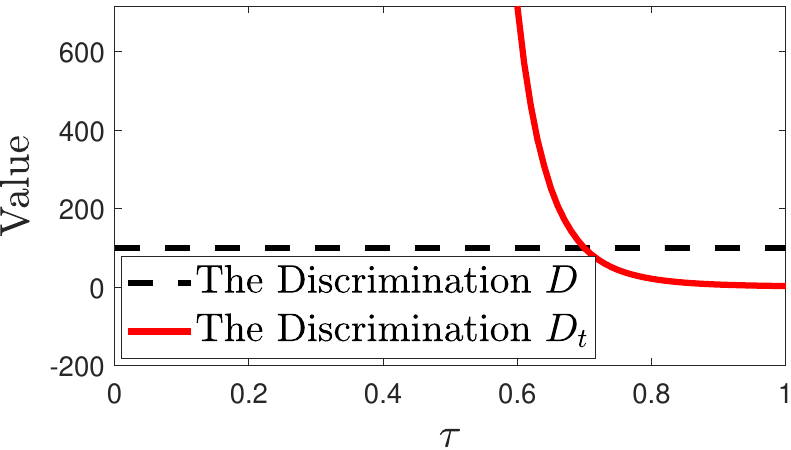}
			\caption{\footnotesize Numerical results for the data with distribution parameters $\mu=0.99$ and $\sigma^2=0.02$}
		\end{subfigure}
	\end{minipage}

	\begin{minipage}{\textwidth}
		\centering
		\begin{subfigure}[b]{0.45\textwidth}
			\includegraphics[width=\textwidth]{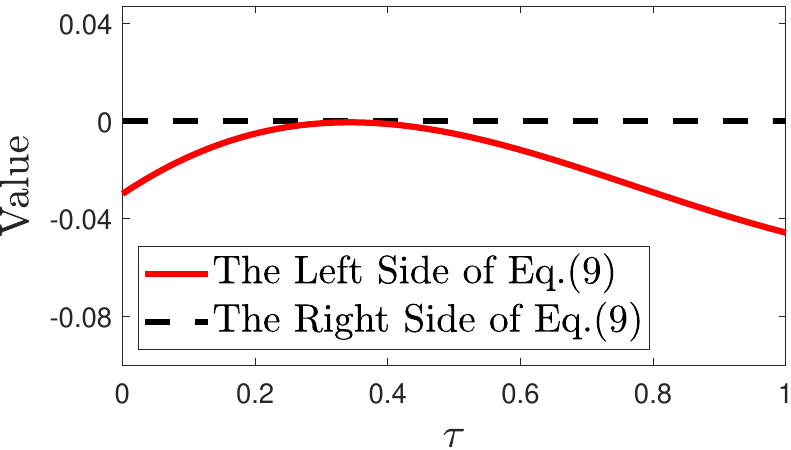}
			\caption{\footnotesize Theoretical results for the data with distribution parameters $\mu=0.66$ and $\sigma^2=0.56$}
		\end{subfigure}
		\begin{subfigure}[b]{0.45\textwidth}
			\includegraphics[width=\textwidth]{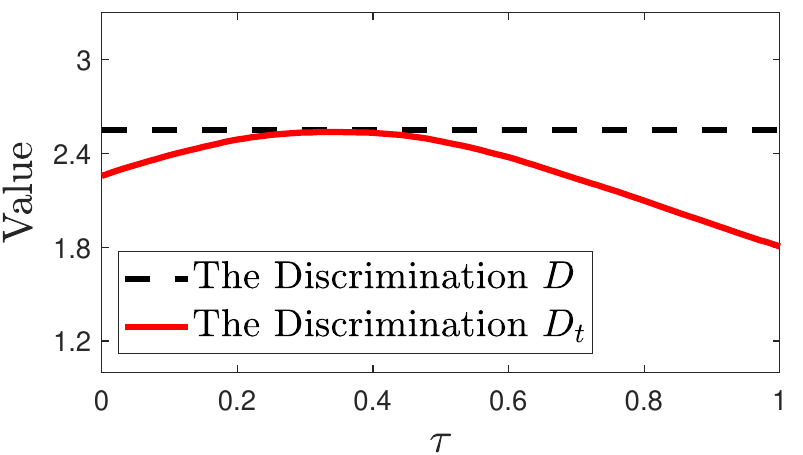}
			\caption{\footnotesize Numerical results for the data with distribution parameters $\mu=0.66$ and $\sigma^2=0.56$}
		\end{subfigure}
	\end{minipage}
	\captionsetup{font=normalsize}
	\caption{Consider the ternary quantization on two classes of data $X\sim N(\mu, \sigma^2)$ and $Y\sim N(-\mu, \sigma^2)$  as specified in Property \ref{data-distribution}.  For  two kinds of data with distribution parameters ($\mu=0.99$, $\sigma^2=0.02$) and ($\mu=0.66$, $\sigma^2=0.56$), the values for the left and right side of Equations \eqref{eq-tq}  are provided in (a) and (c)  respectively; and their discrimination $D$ and $D_t$ statistically estimated with Equations \eqref{d-1} and \eqref{d-T}  are illustrated in (b) and (d), respectively.}
	\label{fig:Numerical simu-3}
\end{figure}

\newpage
\subsection{Classification on synthetic data: KNN and SVM}

\begin{figure}[H]
	\centering
	\begin{minipage}{\textwidth}
		\centering	
		\begin{subfigure}[b]{0.24\textwidth}
			\includegraphics[width=\textwidth]{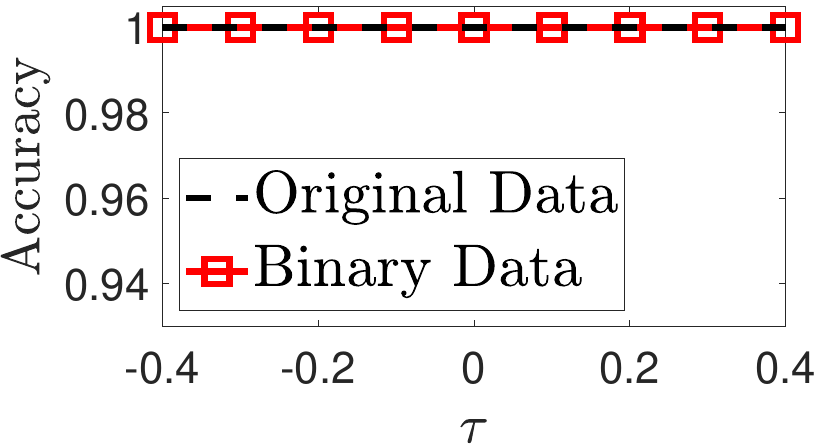}
			\caption{\scriptsize Binary data ($\lambda=0$)}
		\end{subfigure}
		\begin{subfigure}[b]{0.24\textwidth}
			\includegraphics[width=\textwidth]{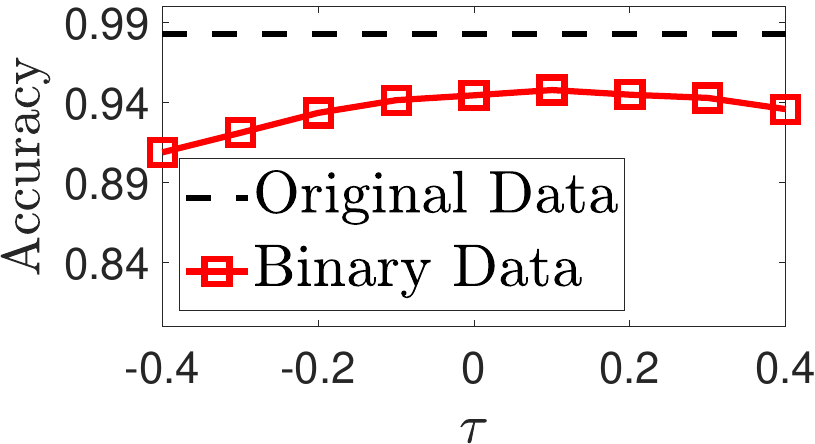}
			\caption{\scriptsize Binary data ($\lambda=0.01$)}
		\end{subfigure}
		\begin{subfigure}[b]{0.24\textwidth}
			\includegraphics[width=\textwidth]{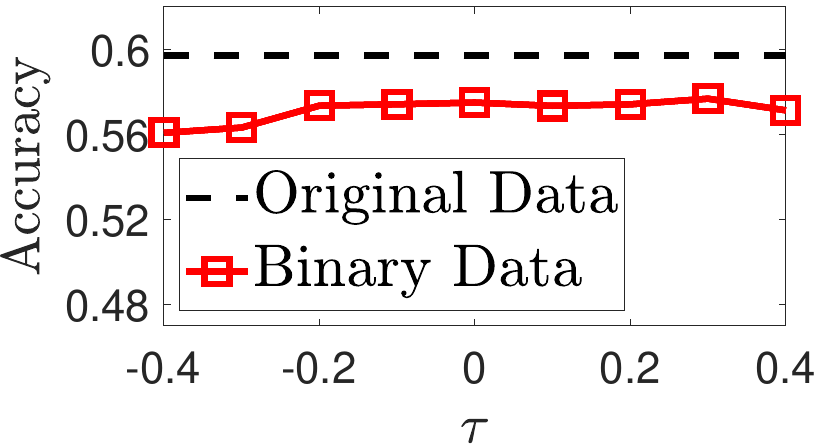}
			\caption{\scriptsize Binary data ($\lambda=0.1$)}
		\end{subfigure}
		\begin{subfigure}[b]{0.24\textwidth}
			\includegraphics[width=\textwidth]{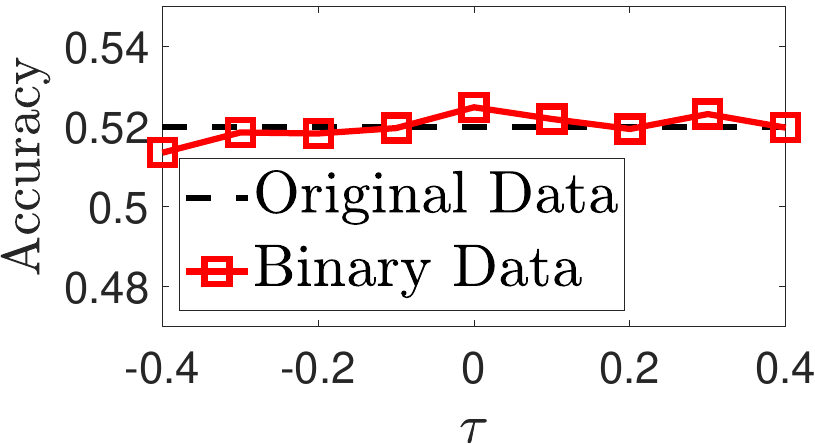}
			\caption{\scriptsize Binary data ($\lambda=1$)}
		\end{subfigure}
	\end{minipage}
	\begin{minipage}{\textwidth}
		\centering
		\begin{subfigure}[b]{0.24\textwidth}
			\includegraphics[width=\textwidth]{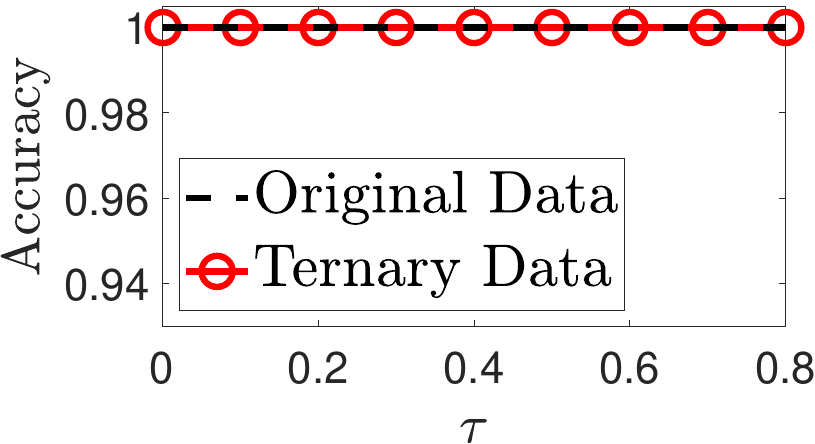}
			\caption{\scriptsize Ternary data ($\lambda=0$)}
		\end{subfigure}
		\begin{subfigure}[b]{0.24\textwidth}
			\includegraphics[width=\textwidth]{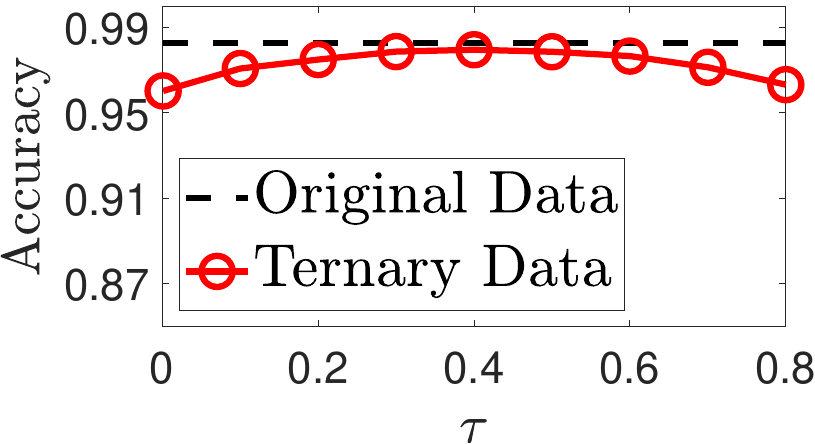}
			\caption{\scriptsize Ternary data ($\lambda=0.01$)}
		\end{subfigure}
		\begin{subfigure}[b]{0.24\textwidth}
			\includegraphics[width=\textwidth]{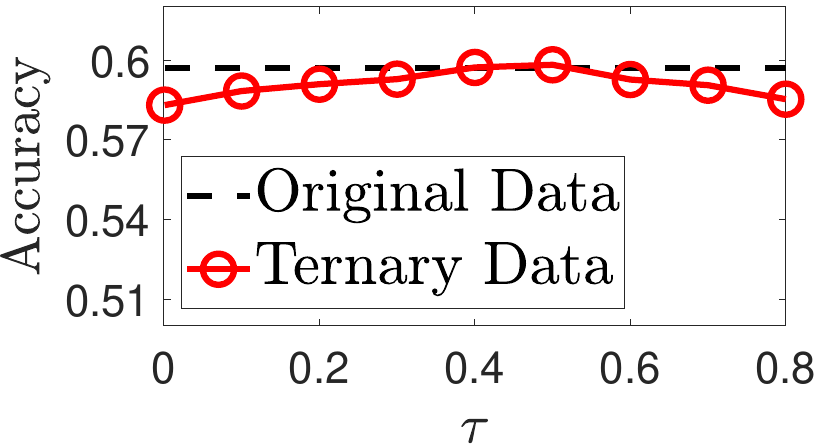}
			\caption{\scriptsize Ternary data ($\lambda=0.1$)}
		\end{subfigure}
		\begin{subfigure}[b]{0.24\textwidth}
			\includegraphics[width=\textwidth]{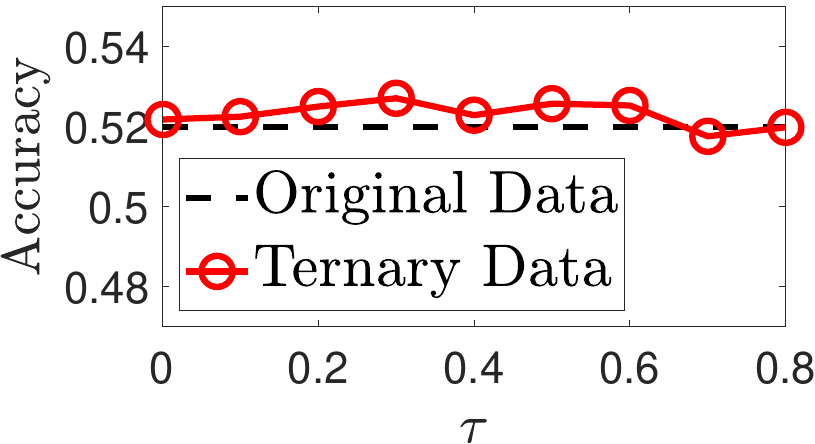}
			\caption{\scriptsize Ternary data ($\lambda=1$)}
		\end{subfigure}
	\end{minipage}
	\caption{ KNN (Cosine) classification accuracy for the 10,000-dimensional binary, ternary, and original data that are generated with  the varying  parameter $\lambda\in\{0,0.01,0.1,1\}$, which controls the data sparsity.}
	\captionsetup{font=normalsize}
	\label{fig:synthetic-lambda knn-Cos}
	
\end{figure}

\begin{figure}[H]
	\centering
	\begin{minipage}{\textwidth}
		\centering	
		\begin{subfigure}[b]{0.24\textwidth}
			\includegraphics[width=\textwidth]{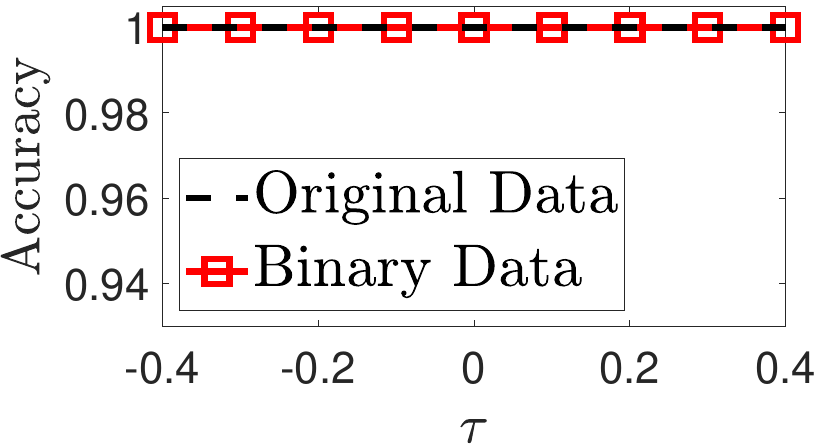}
			\caption{\scriptsize Binary data ($\lambda=0$)}
		\end{subfigure}
		\begin{subfigure}[b]{0.24\textwidth}
			\includegraphics[width=\textwidth]{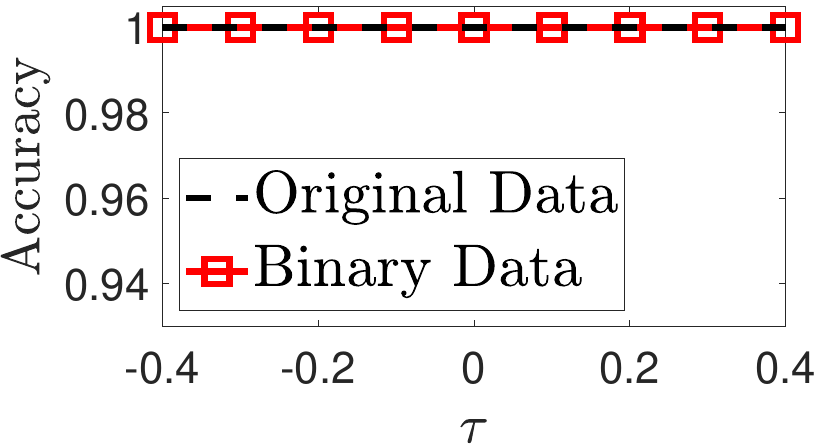}
			\caption{\centering \scriptsize Binary data ($\lambda=0.01$)}
		\end{subfigure}
		\begin{subfigure}[b]{0.24\textwidth}
			\includegraphics[width=\textwidth]{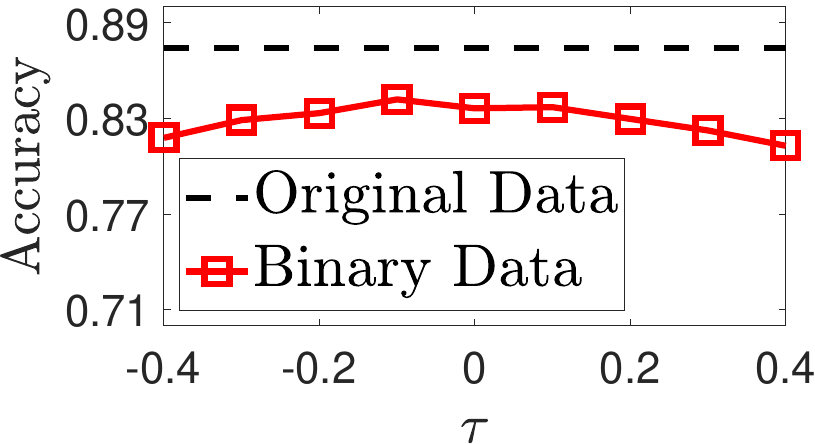}
			\caption{\scriptsize Binary data ($\lambda=0.1$)}
		\end{subfigure}
		\begin{subfigure}[b]{0.24\textwidth}
			\includegraphics[width=\textwidth]{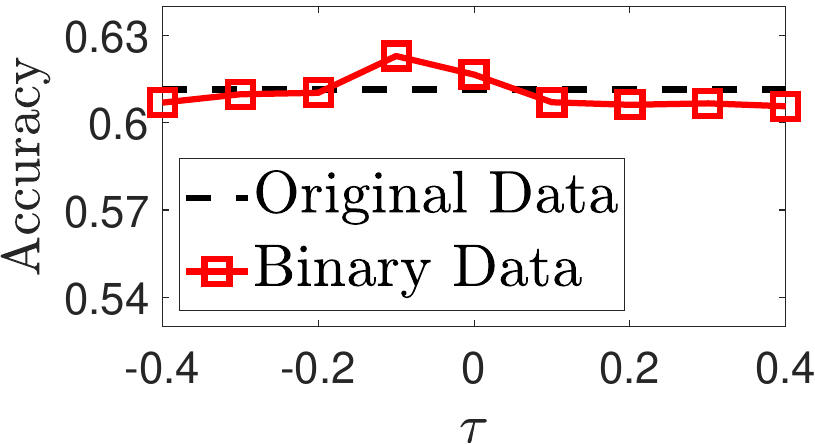}
			\caption{\centering \scriptsize Binary data ($\lambda=1$)}
		\end{subfigure}
	\end{minipage}
	\begin{minipage}{\textwidth}
		\centering
		\begin{subfigure}[b]{0.24\textwidth}
			\includegraphics[width=\textwidth]{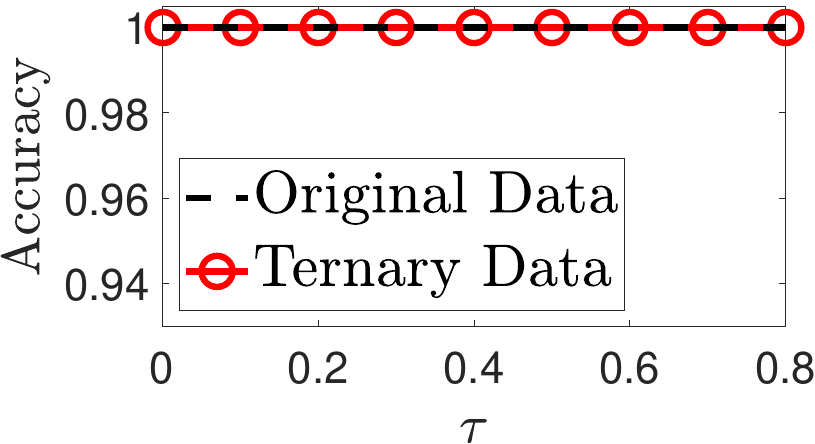}
			\caption{\scriptsize Ternary data ($\lambda=0$)}
		\end{subfigure}
		\begin{subfigure}[b]{0.24\textwidth}
			\includegraphics[width=\textwidth]{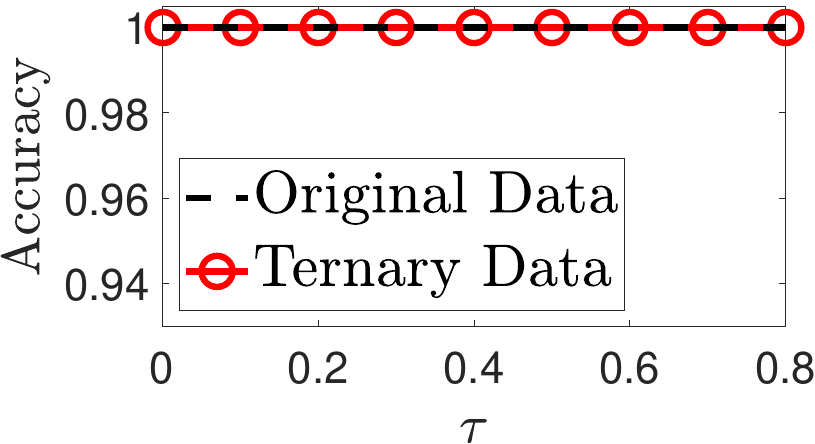}
			\caption{\scriptsize Ternary data ($\lambda=0.01$)}
		\end{subfigure}	
		\begin{subfigure}[b]{0.24\textwidth}
			\includegraphics[width=\textwidth]{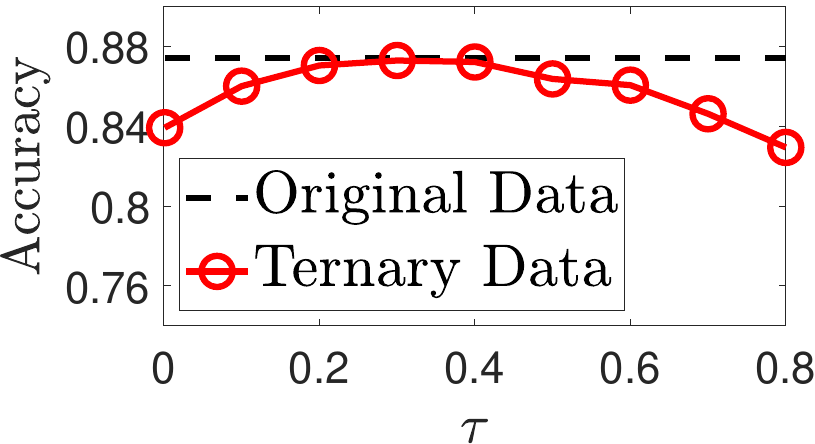}
			\caption{\scriptsize Ternary data ($\lambda=0.1$)}
		\end{subfigure}
		\begin{subfigure}[b]{0.24\textwidth}
			\includegraphics[width=\textwidth]{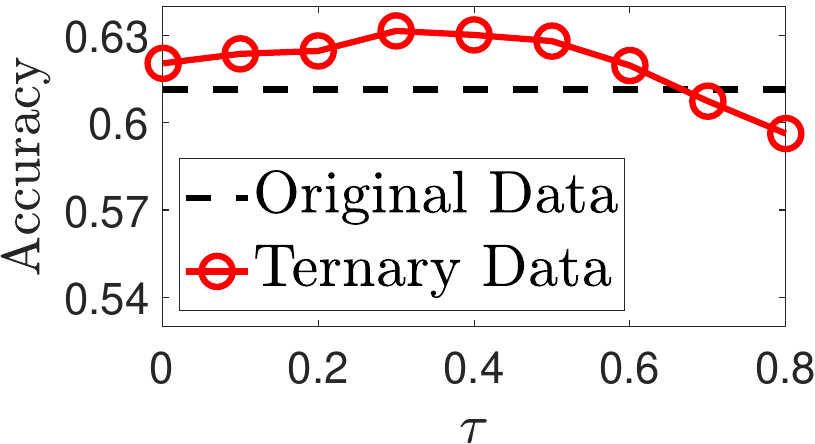}
			\caption{\scriptsize Ternary data ($\lambda=1$)}
		\end{subfigure}
	\end{minipage}
	\caption{SVM classification accuracy for the 10,000-dimensional binary, ternary, and original data that are generated with  the varying  parameter $\lambda\in\{0,0.01,0.1,1\}$, which controls the data sparsity. }
	\captionsetup{font=normalsize}
	\label{fig:synthetic-lambda svm}
	
\end{figure}

\newpage

\begin{figure}[H]
	\centering
	\begin{minipage}{\textwidth}		
		\centering
		\begin{subfigure}[b]{0.4\textwidth}
			\includegraphics[width=\textwidth]{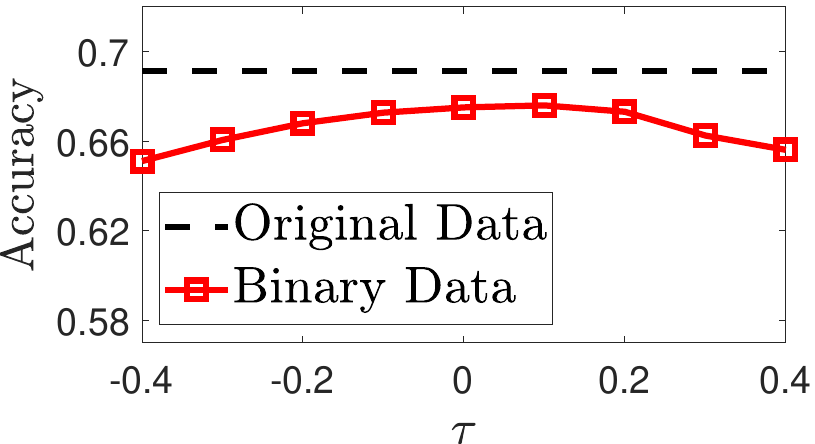}
			\caption{\footnotesize 100-dim binary data}
		\end{subfigure}
		\begin{subfigure}[b]{0.4\textwidth}
			\includegraphics[width=\textwidth]{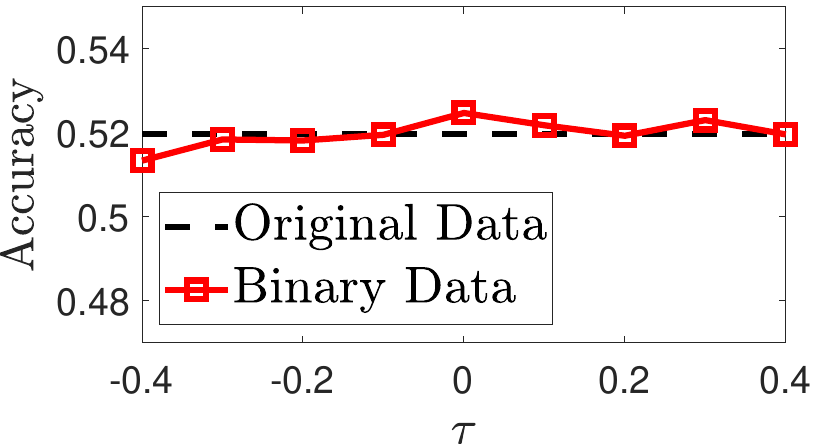}
			\caption{\footnotesize 10,000-dim binary data}
		\end{subfigure}
	\end{minipage}
	\begin{minipage}{\textwidth}
			\centering
		\begin{subfigure}[b]{0.4\textwidth}
			\includegraphics[width=\textwidth]{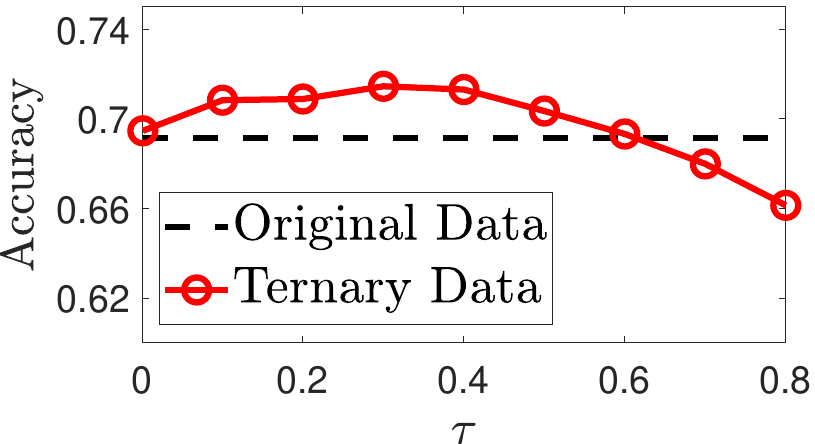}
			\caption{\footnotesize 100-dim ternary data}
		\end{subfigure}
		\begin{subfigure}[b]{0.4\textwidth}
			\includegraphics[width=\textwidth]{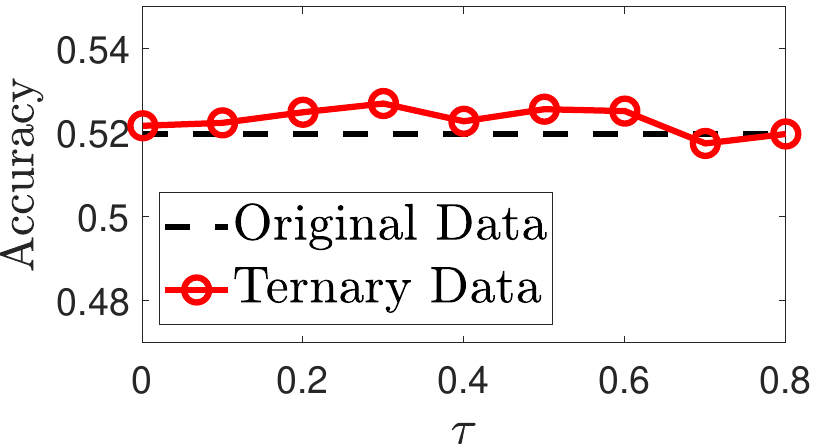}
			\caption{\footnotesize 10,000-dim ternary data}
		\end{subfigure}
	\end{minipage}
	\caption{KNN (Cosine) classification accuracy for the binary, ternary, and original data generated with the sparsity parameter $\lambda=1$, and with varying dimensions $n\in\{100,10000\}$.}
	\captionsetup{font=normalsize}
	\label{fig:synthetic knn-Cos}
	
\end{figure}

\begin{figure}[H]
	\centering
	\begin{minipage}{\textwidth}		
		\begin{subfigure}[b]{0.32\textwidth}
			\includegraphics[width=\textwidth]{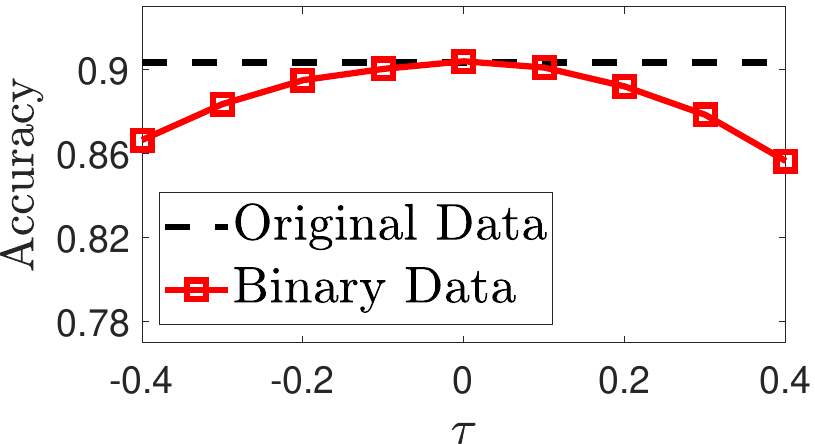}
			\caption{\footnotesize 1-dim binary data}
		\end{subfigure}
		\begin{subfigure}[b]{0.32\textwidth}
			\includegraphics[width=\textwidth]{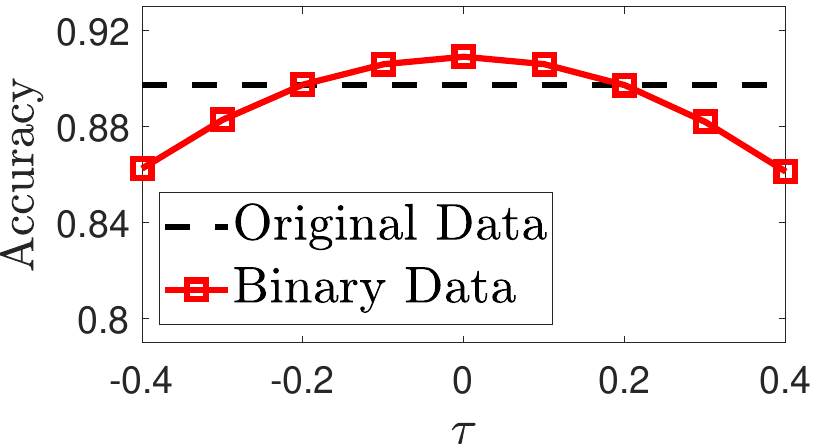}
			\caption{\footnotesize 100-dim binary data}
		\end{subfigure}
		\begin{subfigure}[b]{0.32\textwidth}
			\includegraphics[width=\textwidth]{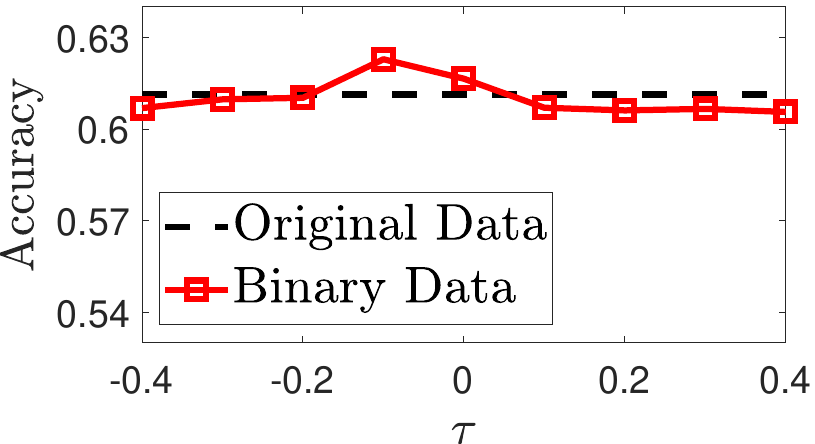}
			\caption{\footnotesize 10,000-dim binary data}
		\end{subfigure}
	\end{minipage}
	\begin{minipage}{\textwidth}
		\begin{subfigure}[b]{0.32\textwidth}
			\includegraphics[width=\textwidth]{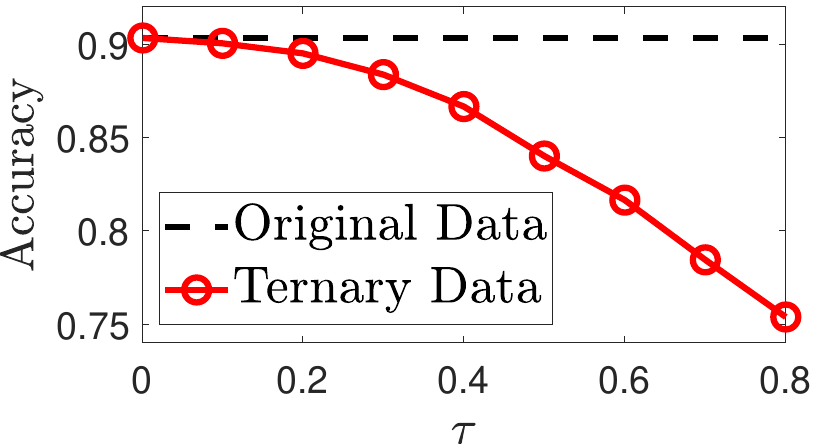}
			\caption{\footnotesize 1-dim ternary data}
		\end{subfigure}
		\begin{subfigure}[b]{0.32\textwidth}
			\includegraphics[width=\textwidth]{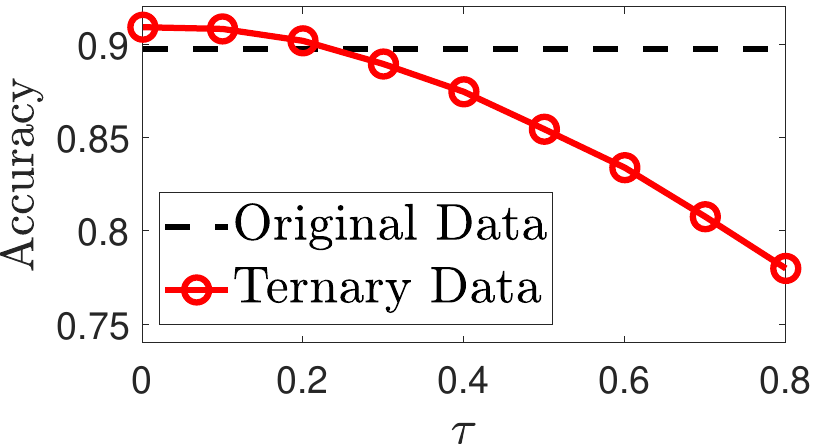}
			\caption{\footnotesize 100-dim ternary data}
		\end{subfigure}
		\begin{subfigure}[b]{0.32\textwidth}
			\includegraphics[width=\textwidth]{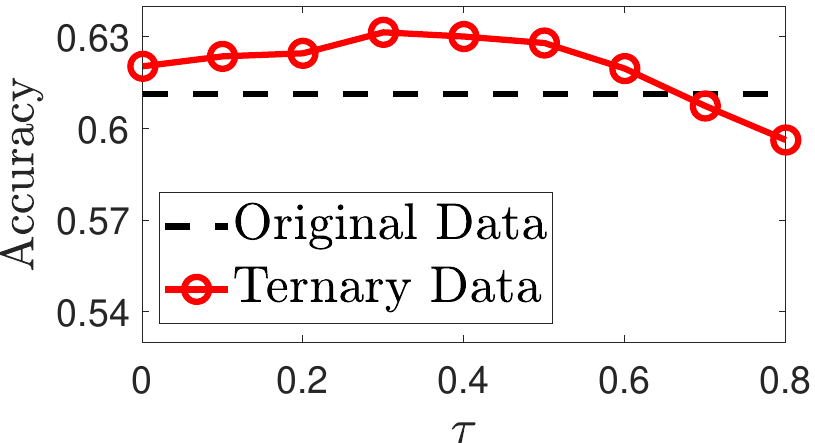}
			\caption{\footnotesize 10,000-dim ternary data}
		\end{subfigure}
	\end{minipage}
	\caption{SVM classification accuracy for the binary, ternary, and original data generated with the sparsity parameter $\lambda=1$, and with varying dimensions $n\in\{1,100,10000\}$.}
	\captionsetup{font=normalsize}
	\label{fig:synthetic svm}
	
\end{figure}

\newpage
\subsection{Classification on real data: KNN and SVM}

\begin{figure}[h]
    \centering
	\begin{minipage}{\textwidth}
		\centering
		\begin{subfigure}[b]{0.45\textwidth}
			\includegraphics[width=\textwidth]{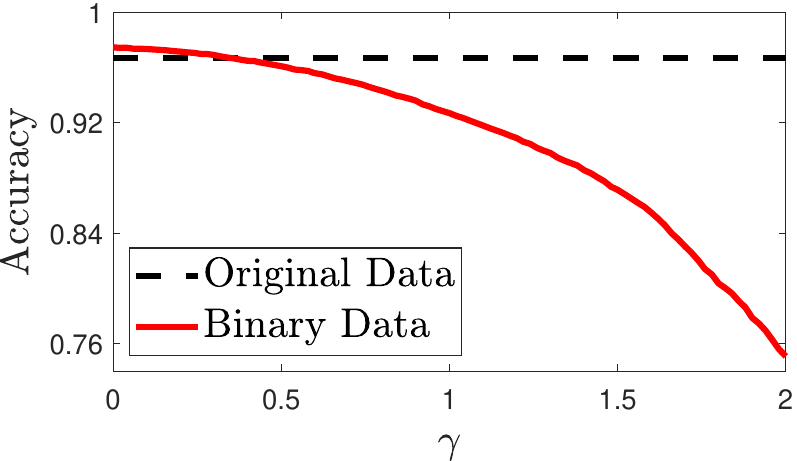}
			\caption{\centering \footnotesize Binary data with KNN (Euclidean)}
		\end{subfigure}
		\begin{subfigure}[b]{0.45\textwidth}
			\includegraphics[width=\textwidth]{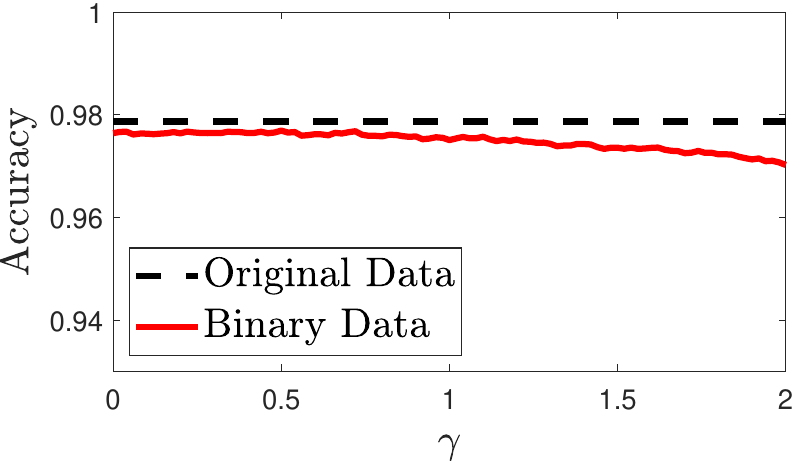}
			\caption{\centering \footnotesize Binary data with SVM}
		\end{subfigure}
	\end{minipage}

	\begin{minipage}{\textwidth}
		\centering
		\begin{subfigure}[b]{0.45\textwidth}
			\includegraphics[width=\textwidth]{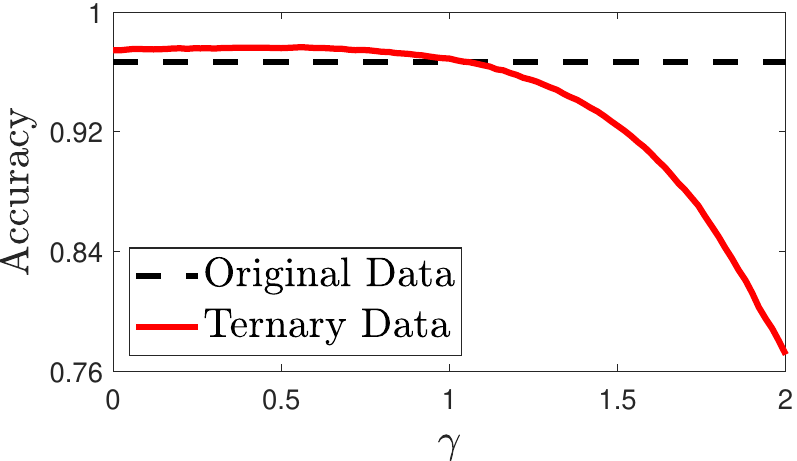}
			\caption{\centering \footnotesize Ternary data with KNN (Euclidean)}
		\end{subfigure}
		\begin{subfigure}[b]{0.45\textwidth}
			\includegraphics[width=\textwidth]{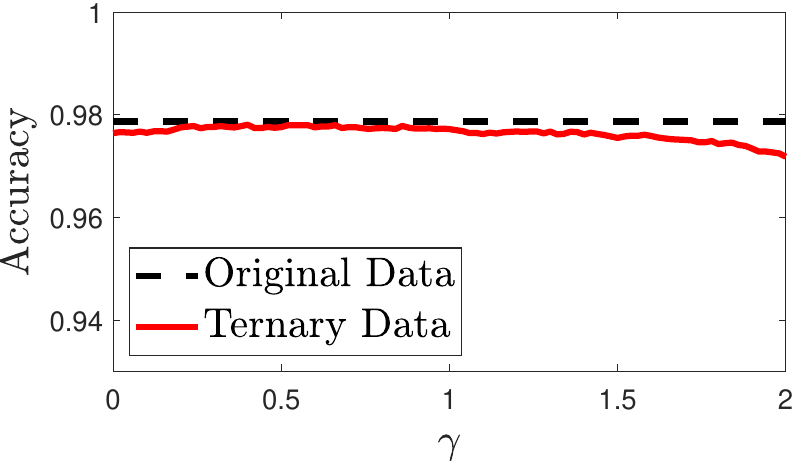}
			\caption{\centering \footnotesize Ternary data with SVM}
		\end{subfigure}
	\end{minipage}
	
	\caption{Classification accuracy for the binary, ternary, and original data by KNN (Euclidean distance) and SVM on CIFAR10. The parameter $\gamma$ corresponds to a threshold  $\tau=\gamma\cdot\eta$,  where  $\eta$ denotes  the average magnitude of the feature elements  in all  feature vectors. \textbf{Comment:} It can be seen that for both binary and ternary quantization, there exist quantization threshold $\tau$ values that can  achieve improved or at least comparable classification performance compared to the original data.}
	\captionsetup{font=normalsize}
	\label{fig:cifar-knn-svm}
	
\end{figure}

\newpage
\begin{figure}[H]
    \centering
	\begin{minipage}{\textwidth}
		\centering
		\begin{subfigure}[b]{0.45\textwidth}
			\includegraphics[width=\textwidth]{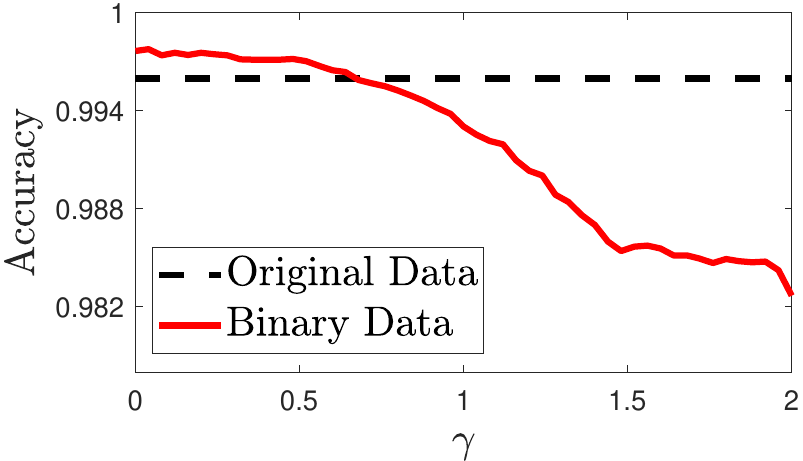}
			\caption{\centering Binary quantization on YaleB}
		\end{subfigure}
		\begin{subfigure}[b]{0.45\textwidth}
			\includegraphics[width=\textwidth]{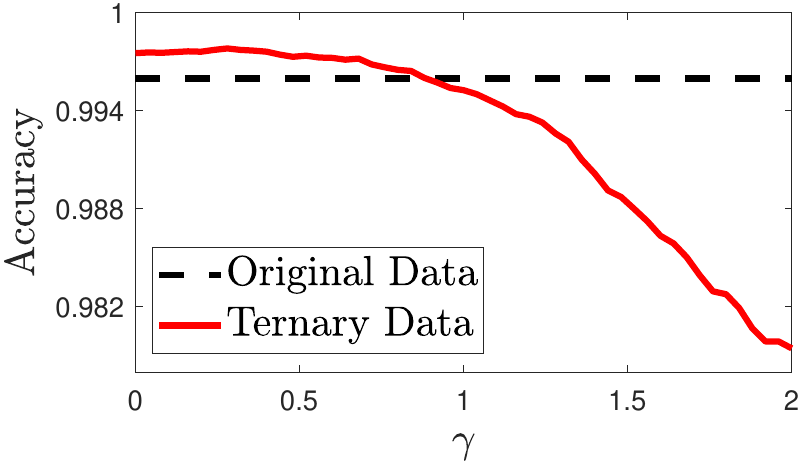}
			\caption{\centering Ternary quantization on YaleB}
		\end{subfigure}
	\end{minipage}
	
	\begin{minipage}{\textwidth}
	\centering
	\begin{subfigure}[b]{0.45\textwidth}
		\includegraphics[width=\textwidth]{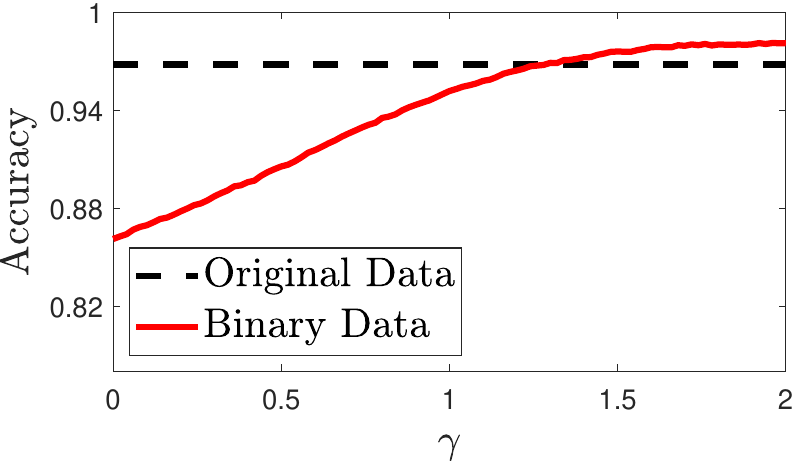}
		\caption{\centering Binary quantization on CIFAR10}
	\end{subfigure}
	\begin{subfigure}[b]{0.45\textwidth}
		\includegraphics[width=\textwidth]{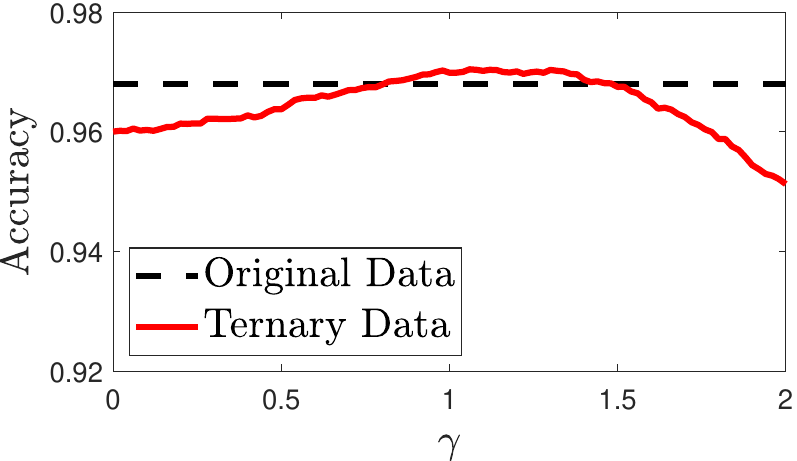}
		\caption{\centering Ternary quantization on CIFAR10}
	\end{subfigure}
	\end{minipage}
	
	\begin{minipage}{\textwidth}
	\centering
	\begin{subfigure}[b]{0.45\textwidth}
		\includegraphics[width=\textwidth]{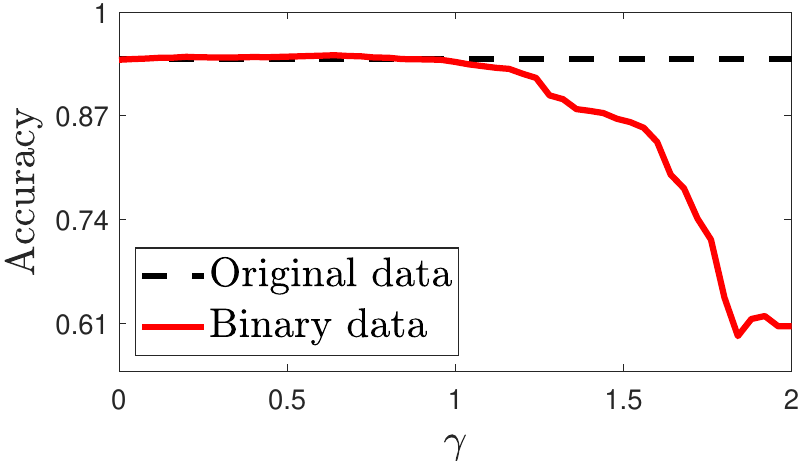}
		\caption{\centering Binary quantization on TIMIT}
	\end{subfigure}
	\begin{subfigure}[b]{0.45\textwidth}
		\includegraphics[width=\textwidth]{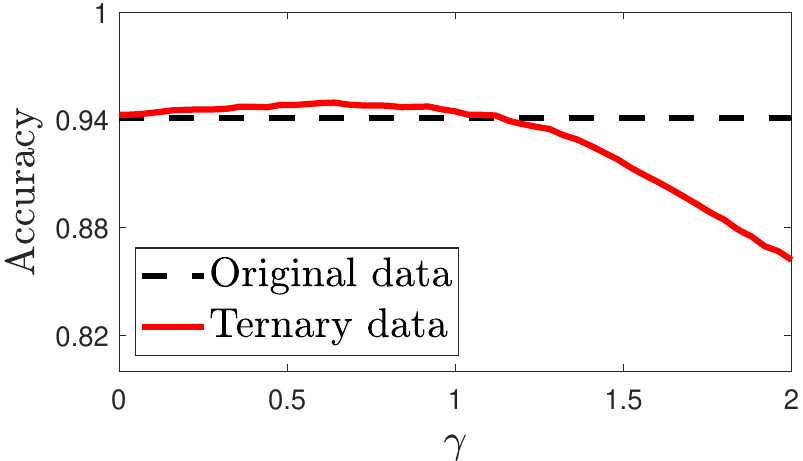}
		\caption{\centering Ternary quantization on TIMIT}
	\end{subfigure}
	\end{minipage}
	
	\begin{minipage}{\textwidth}
	\centering
	\begin{subfigure}[b]{0.45\textwidth}
		\includegraphics[width=\textwidth]{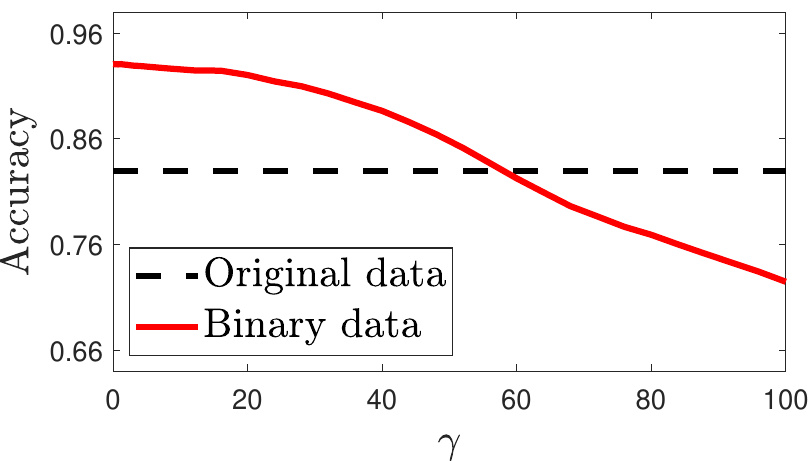}
		\caption{\centering Binary quantization on Newsgroup}
	\end{subfigure}
	\begin{subfigure}[b]{0.45\textwidth}
		\includegraphics[width=\textwidth]{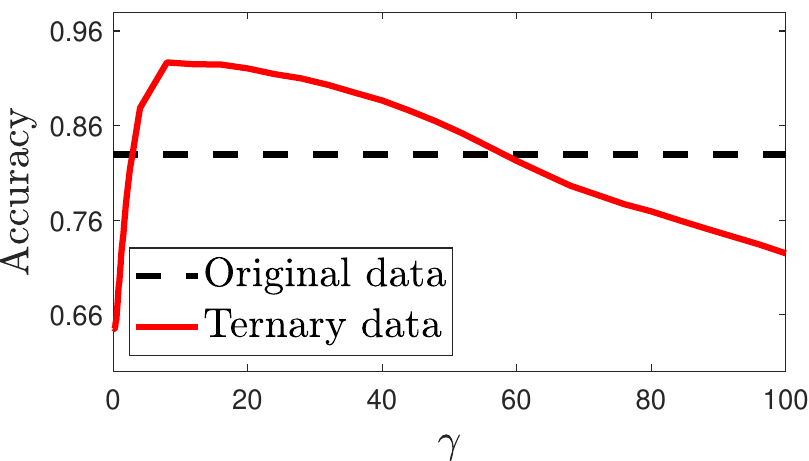}
		\caption{\centering Tenary quantization on Newsgroup}
	\end{subfigure}
	\end{minipage}
	\caption{Classification accuracy for the binary, ternary, and original data by KNN (Cosine distance) on four different datasets. The parameter $\gamma$ corresponds to a quantization threshold  $\tau=\gamma\cdot\eta$,  where  $\eta$ denotes  the average magnitude of the feature elements  in all  feature vectors. \textbf{Comment:} It can be seen that for both binary and ternary quantization, there exist quantization threshold $\tau$ values that can  achieve improved  classification performance compared to the original data.}
	\captionsetup{font=normalsize}
	\label{fig:Real data knn-Cos}
	
\end{figure}

\newpage
\begin{figure}[H]
	\centering
	\begin{minipage}{\textwidth}
		\centering
		\begin{subfigure}[b]{0.45\textwidth}
			\includegraphics[width=\textwidth]{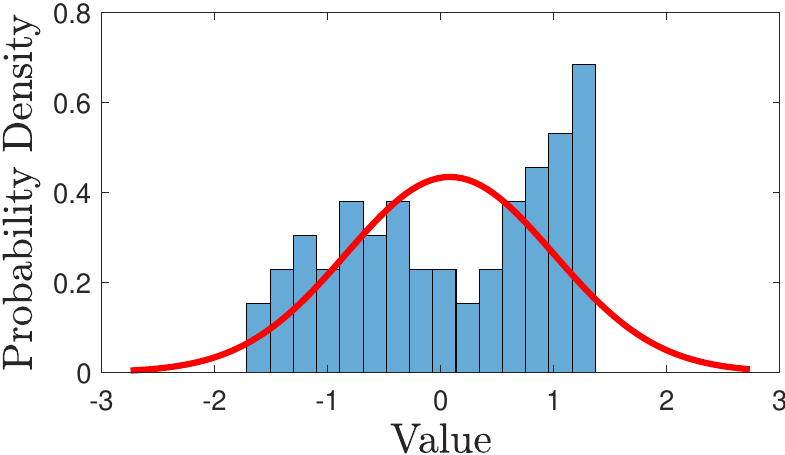}
			\caption{\centering \footnotesize YaleB}
		\end{subfigure}
		\begin{subfigure}[b]{0.45\textwidth}
			\includegraphics[width=\textwidth]{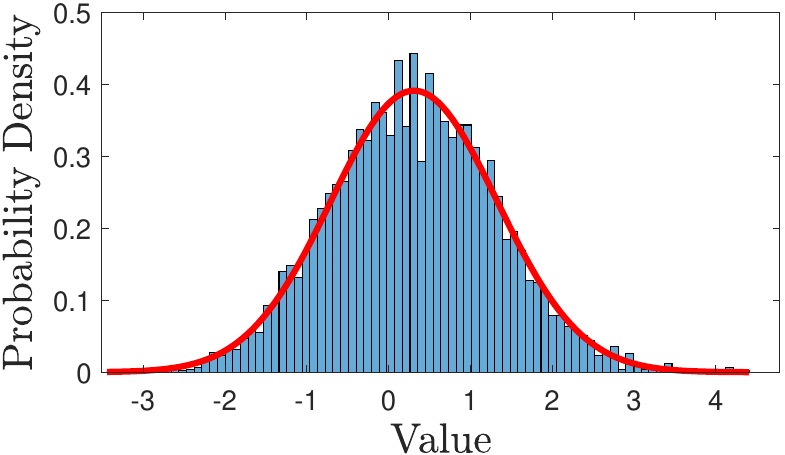}
			\caption{\centering \footnotesize CIFAR10}
		\end{subfigure}
	\end{minipage}
	\begin{minipage}{\textwidth}
		\centering	
		\begin{subfigure}[b]{0.45\textwidth}
			\includegraphics[width=\textwidth]{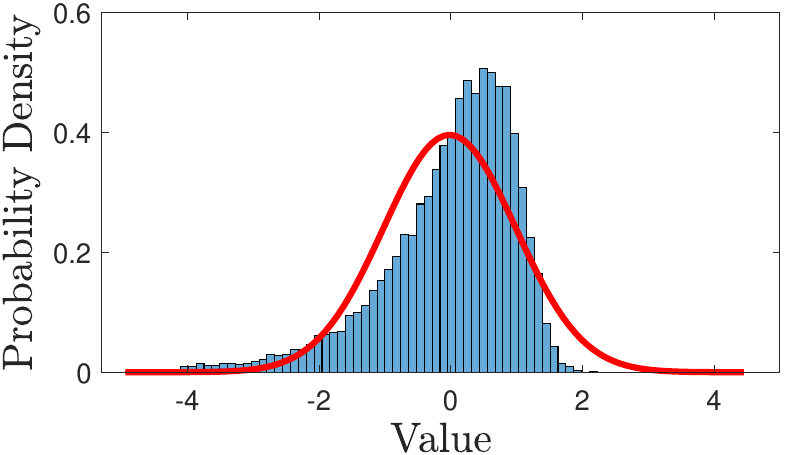}
			\caption{\centering \footnotesize TIMIT}
		\end{subfigure}
		\begin{subfigure}[b]{0.45\textwidth}
			\includegraphics[width=\textwidth]{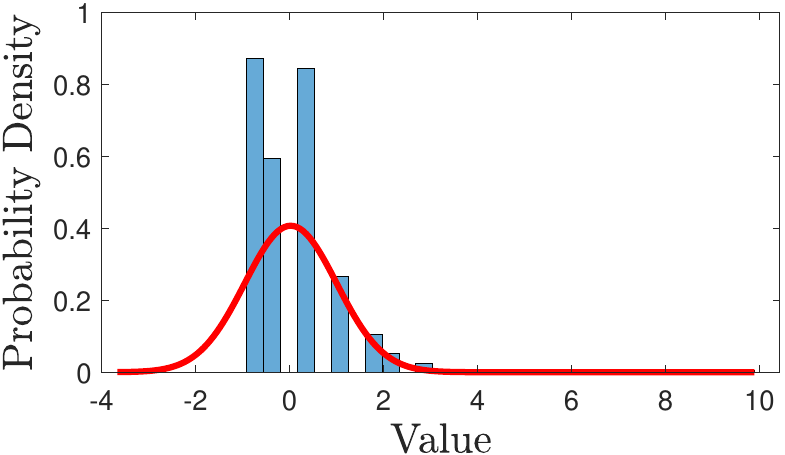}
			\caption{\centering \footnotesize Newsgroup}
		\end{subfigure}
	\end{minipage}
	\captionsetup{font=normalsize}
	\caption{The histogram (blue bar) of the element values across one dimension of the feature vectors within a single class of samples  selected from four different datasets, accompanied with a Gaussian fitting curve (red line).}

 %% Numerical
	\label{fig:real data-density}
\end{figure}

\newpage
\subsection{Classification on real data:  Multilayer Perceptrons (MLP) and  Decision Trees}

\begin{figure}[H]
    \centering
	\begin{minipage}{0.9\textwidth}
		\centering
		\begin{subfigure}[b]{0.45\textwidth}
			\includegraphics[width=\textwidth]{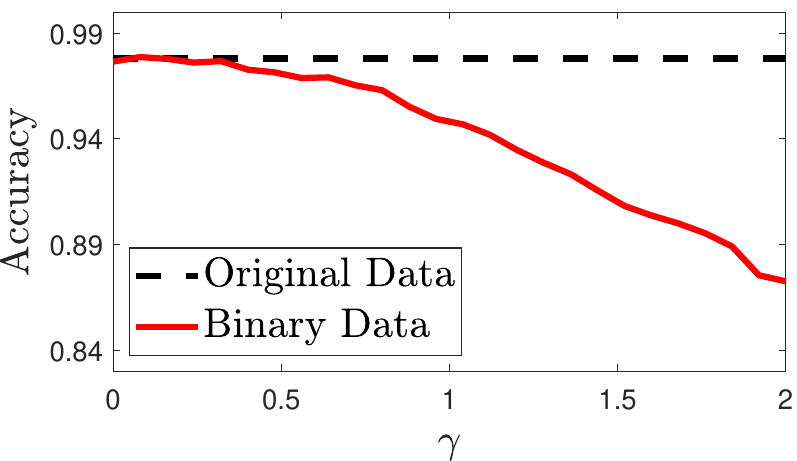}
			\caption{\centering Binary quantization on YaleB}
		\end{subfigure}
		\begin{subfigure}[b]{0.45\textwidth}
			\includegraphics[width=\textwidth]{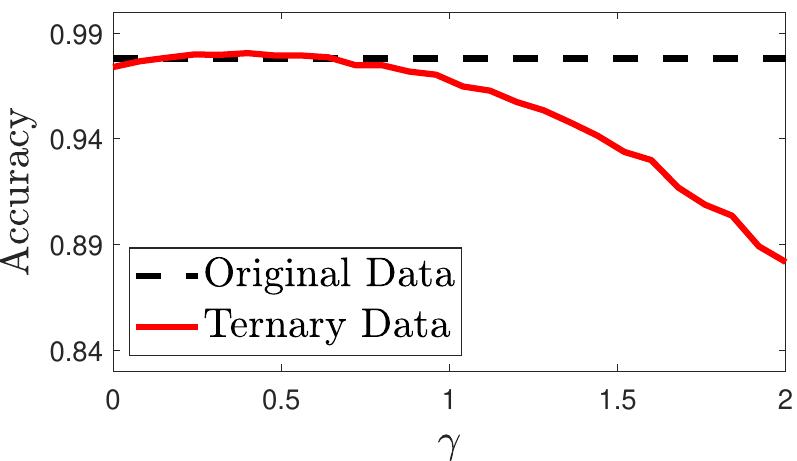}
			\caption{\centering Ternary quantization on YaleB}
		\end{subfigure}
	\end{minipage}
	
	\begin{minipage}{0.9\textwidth}
	\centering
	\begin{subfigure}[b]{0.45\textwidth}
		\includegraphics[width=\textwidth]{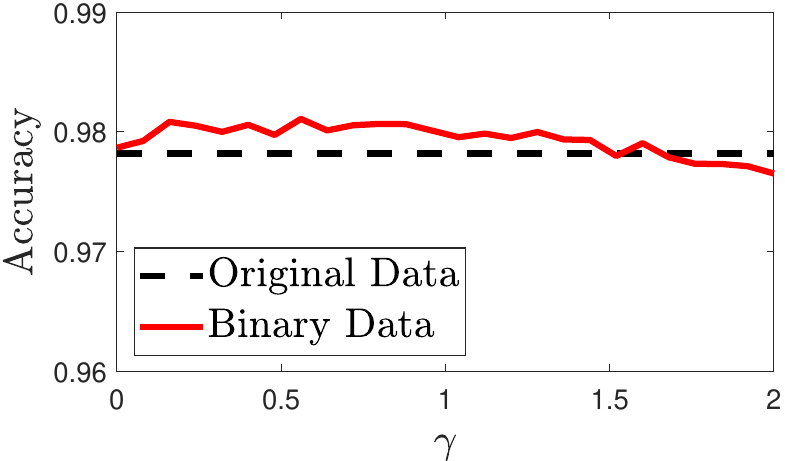}
		\caption{\centering Binary quantization on CIFAR10}
	\end{subfigure}
	\begin{subfigure}[b]{0.45\textwidth}
		\includegraphics[width=\textwidth]{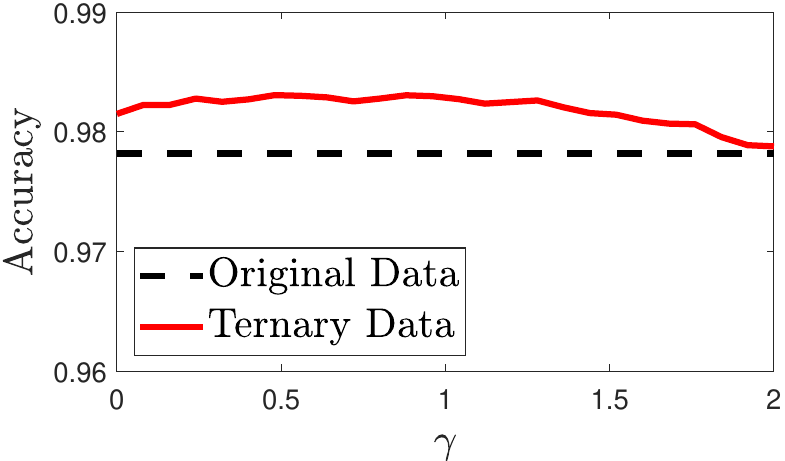}
		\caption{\centering Ternary quantization on CIFAR10}
	\end{subfigure}
	\end{minipage}
	
	\begin{minipage}{0.9\textwidth}
	\centering
	\begin{subfigure}[b]{0.45\textwidth}
		\includegraphics[width=\textwidth]{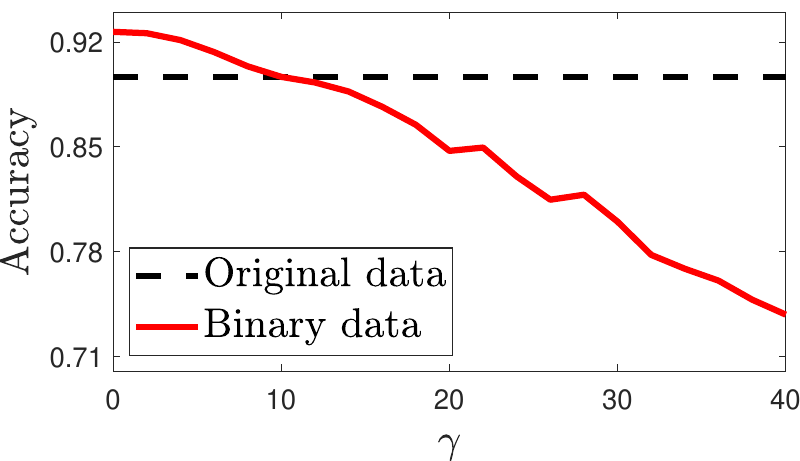}
		\caption{\centering Binary quantization on Newsgroup}
	\end{subfigure}
	\begin{subfigure}[b]{0.45\textwidth}
		\includegraphics[width=\textwidth]{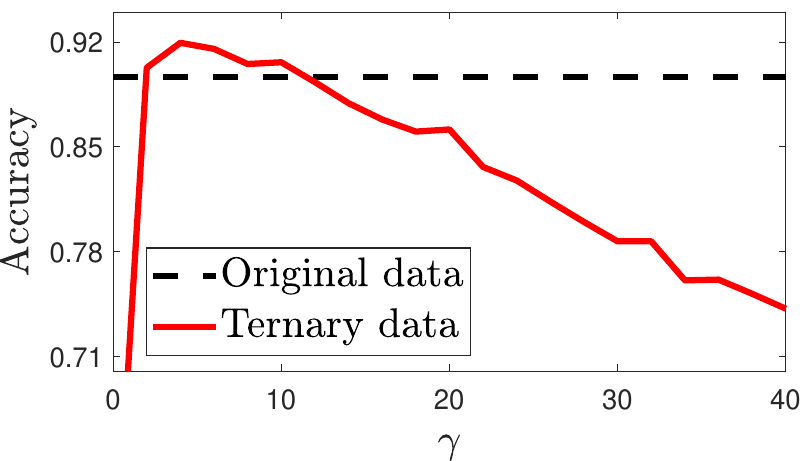}
		\caption{\centering Ternary quantization on Newsgroup}
	\end{subfigure}
	\end{minipage}
	
	\caption{\textbf{MLP}-based binary classification accuracy for the binary, ternary, and original data  on three different datasets. The parameter $\gamma$ corresponds to a quantization threshold  $\tau=\gamma\cdot\eta$,  where  $\eta$ denotes  the average magnitude of the feature elements  in all  feature vectors. \textbf{Comment:} It can be seen that for both binary and ternary quantization, there exist quantization threshold $\tau$ values that can  achieve improved or at least comparable classification performance compared to the original data. }
	\captionsetup{font=normalsize}
	\label{fig:MLP-binary}
	
\end{figure}

\newpage
\begin{figure}[H]
    \centering
	\begin{minipage}{\textwidth}
		\centering
		\begin{subfigure}[b]{0.45\textwidth}
			\includegraphics[width=\textwidth]{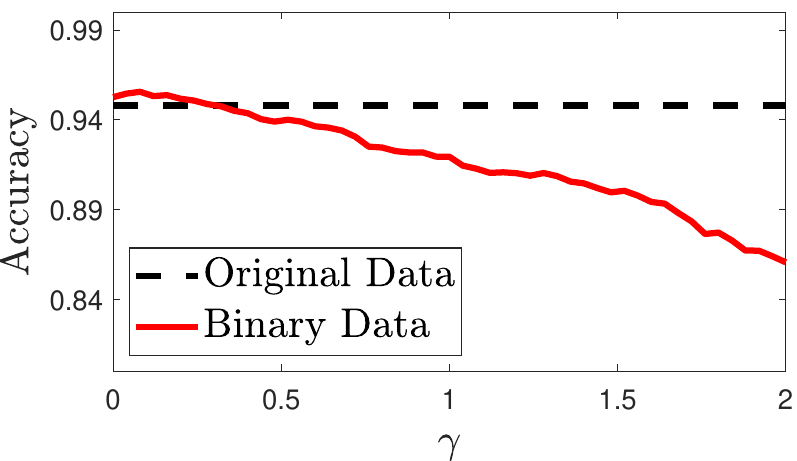}
			\caption{\centering Binary quantization on YaleB}
		\end{subfigure}
		\begin{subfigure}[b]{0.45\textwidth}
			\includegraphics[width=\textwidth]{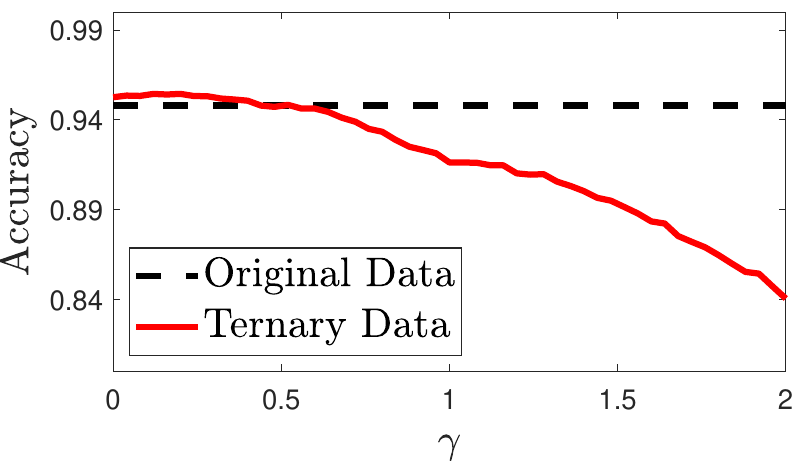}
			\caption{\centering Ternary quantization on YaleB}
		\end{subfigure}
	\end{minipage}
	
	\begin{minipage}{\textwidth}
	\centering
	\begin{subfigure}[b]{0.45\textwidth}
		\includegraphics[width=\textwidth]{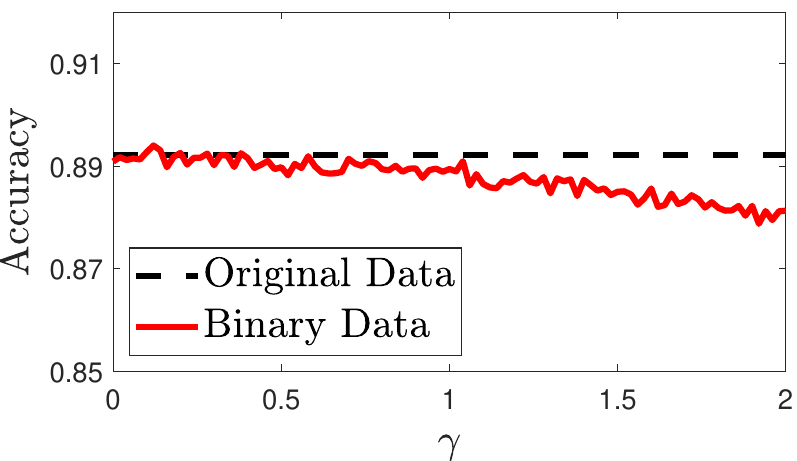}
		\caption{\centering Binary quantization on CIFAR10}
	\end{subfigure}
	\begin{subfigure}[b]{0.45\textwidth}
		\includegraphics[width=\textwidth]{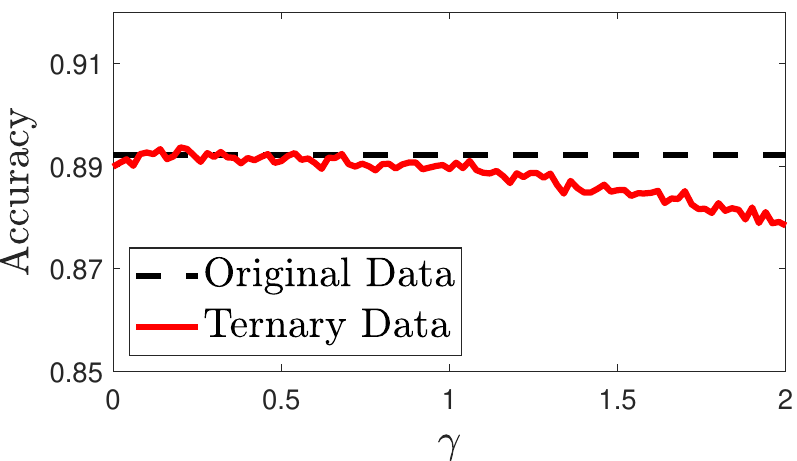}
		\caption{\centering Ternary quantization on CIFAR10}
	\end{subfigure}
	\end{minipage}
	
	\begin{minipage}{\textwidth}
	\centering
	\begin{subfigure}[b]{0.45\textwidth}
		\includegraphics[width=\textwidth]{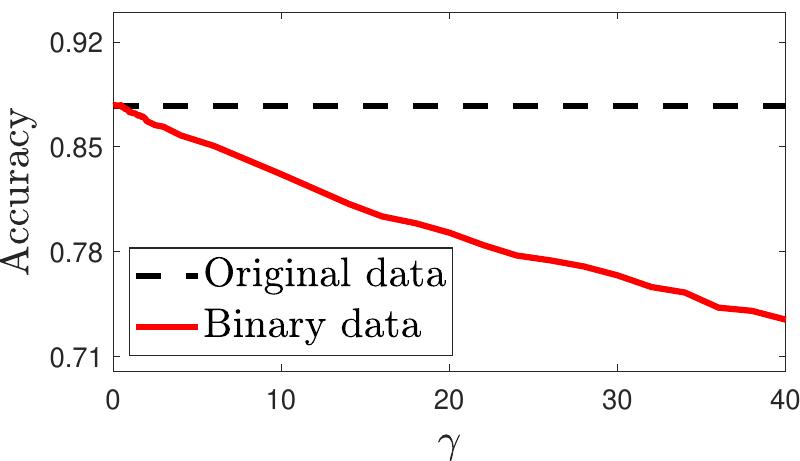}
		\caption{\centering Binary quantization on Newsgroup}
	\end{subfigure}
	\begin{subfigure}[b]{0.45\textwidth}
		\includegraphics[width=\textwidth]{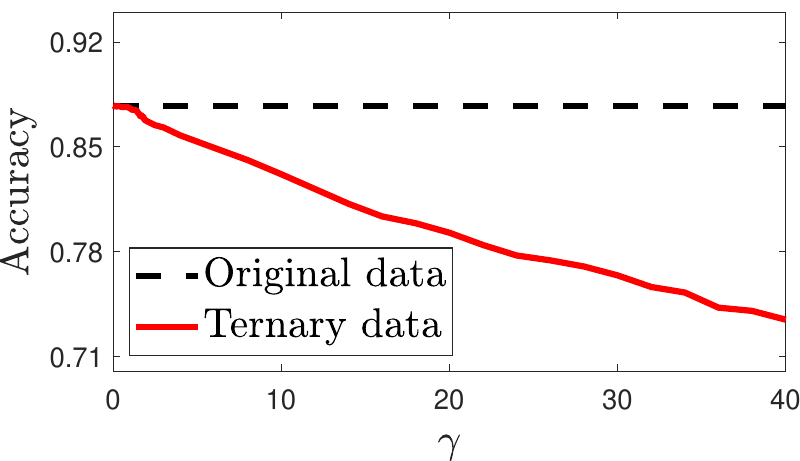}
		\caption{\centering Ternary quantization on Newsgroup}
	\end{subfigure}
	\end{minipage}
	
	\caption{\textbf{Decision trees}-based binary classification accuracy for the binary, ternary, and original data  on three different datasets. The parameter $\gamma$ corresponds to a quantization threshold  $\tau=\gamma\cdot\eta$,  where  $\eta$ denotes  the average magnitude of the feature elements  in all  feature vectors.  \textbf{Comment:} It can be seen that for both binary and ternary quantization, there exist quantization threshold $\tau$ values that can  achieve improved or at least comparable classification performance compared to the original data.}
	\captionsetup{font=normalsize}
	\label{fig:Decision tree-binary}
	
\end{figure}

\newpage
\subsection{Binary and Multiclass Classifications on ImageNet1000 (with 1000 classes of samples)}

\begin{figure}[H]
	\centering
	\begin{minipage}{\textwidth}
		\centering
		\begin{subfigure}[b]{0.45\textwidth}
			\includegraphics[width=\textwidth]{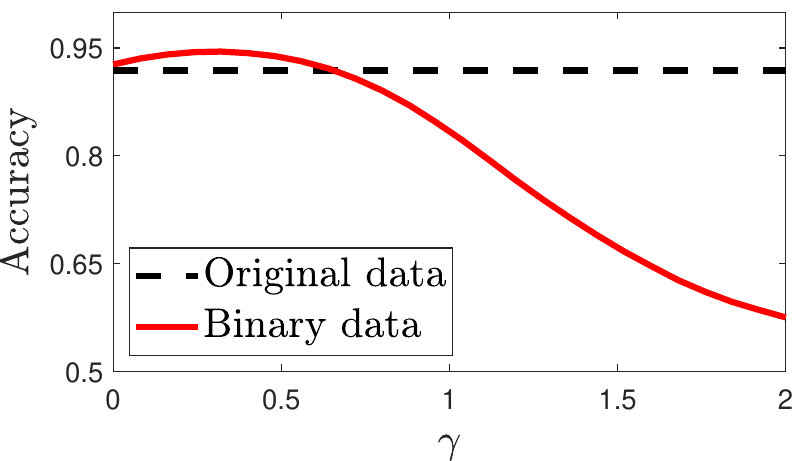}
		\caption{\centering Binary quantization}
		\end{subfigure}
		\begin{subfigure}[b]{0.45\textwidth}
			\includegraphics[width=\textwidth]{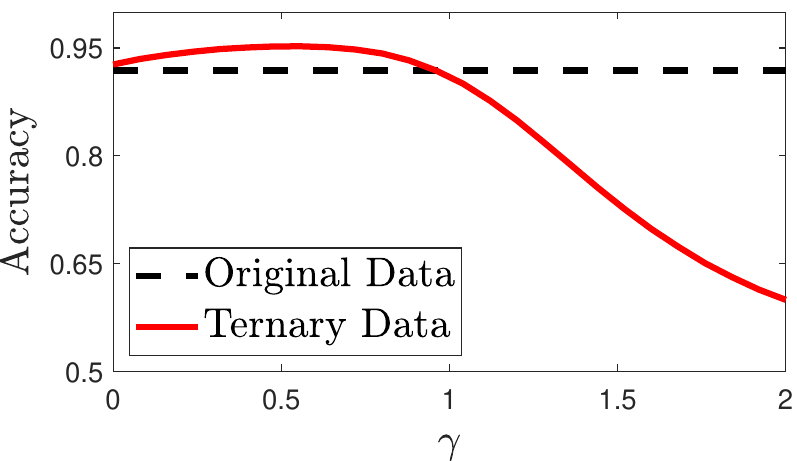}
		\caption{\centering Ternary quantization}
		\end{subfigure}
	\end{minipage}

	\captionsetup{font=normalsize}
	\caption{\textbf{Binary} classification accuracy for the binary, ternary, and original data in ImageNet1000, using the classifier KNN (Euclidean distance). The parameter $\gamma$ corresponds to a quantization threshold  $\tau=\gamma\cdot\eta$,  where  $\eta$ denotes  the average magnitude of the feature elements  in all  feature vectors. \textbf{Comment:} It can be seen that for both binary and ternary quantization, there exist quantization threshold  $\tau$  values that can enhance the binary classification accuracy compared to the original data. }
 %% Numerical
	\label{fig:ImageNet-KNN-BC}
\end{figure}

\begin{figure}[H]
	\centering
	\begin{minipage}{\textwidth}
		\centering
		\begin{subfigure}[b]{0.45\textwidth}
			\includegraphics[width=\textwidth]{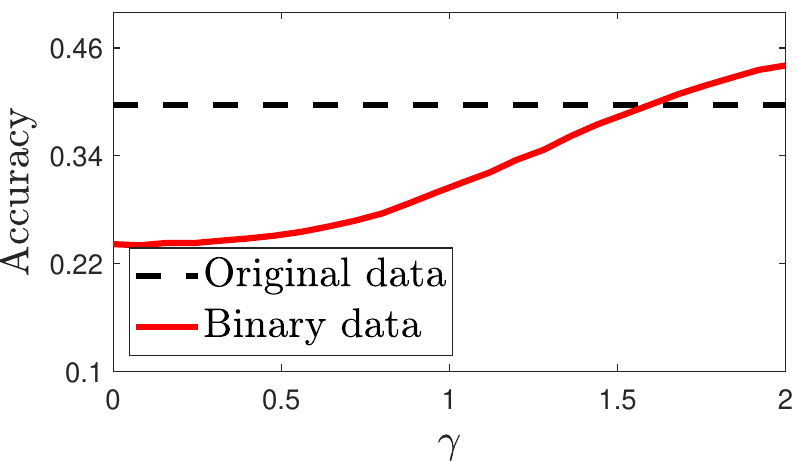}
		\caption{\centering Binary quantization}
		\end{subfigure}
		\begin{subfigure}[b]{0.45\textwidth}
			\includegraphics[width=\textwidth]{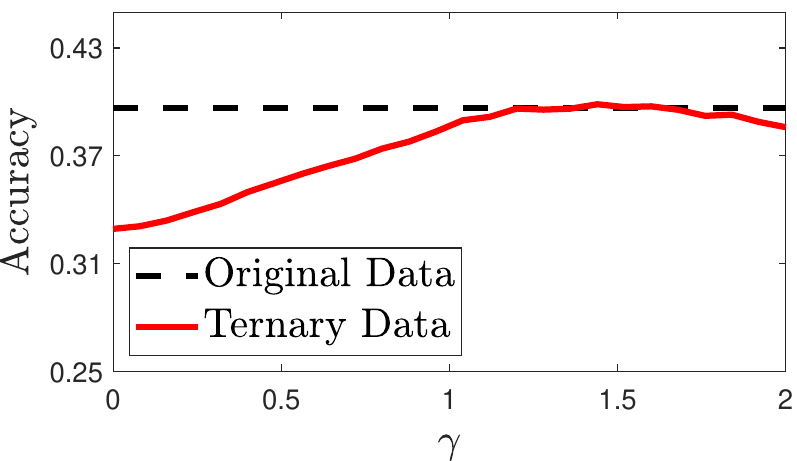}
		\caption{\centering Ternary quantization}
		\end{subfigure}
	\end{minipage}

	\captionsetup{font=normalsize}
	\caption{\textbf{Multiclass} (1000-class) classification accuracy for the binary, ternary, and original data in ImageNet1000, using the classifier KNN (Euclidean distance). The parameter $\gamma$ corresponds to a quantization threshold  $\tau=\gamma\cdot\eta$,  where  $\eta$ denotes  the average magnitude of the feature elements  in all  feature vectors. \textbf{Comment:} It can be seen that for both binary and ternary quantization, there are quantization threshold $\tau$ values that can achieve improved or at least comparable  classification accuracy compared to the original data.}
 %% Numerical
	\label{fig:ImageNet-KNN-MC}
\end{figure}
\end{document}